%% file: main.tex
\documentclass[10pt,twocolumn,letterpaper]{article}

\usepackage[pagenumbers]{cvpr}

\usepackage{comment}
\usepackage[percent]{overpic}
\usepackage{multirow}

\include{packages}
\include{macros}

\usepackage{array}
\newcolumntype{P}[1]{>{\centering\arraybackslash}p{#1}}

\begin{document}

\title{TEXTure: Text-Guided Texturing of 3D Shapes}

\author{Elad Richardson\textsuperscript{*}
\qquad
Gal Metzer\textsuperscript{*}
\qquad
Yuval Alaluf
\qquad
Raja Giryes
\qquad
Daniel Cohen-Or\\ \\
Tel Aviv University 
}

\twocolumn[{%
\renewcommand\twocolumn[1][]{#1}%
\vspace{-1em}
\maketitle
\vspace{-3em}
\input{figures/teaser/fig}

}]

\def\thefootnote{*}\footnotetext{Denotes equal contribution}

\begin{abstract}
    \input{sections/abstract.tex}

\end{abstract}
\input{sections/intro}
\input{sections/relatedwork}

\input{sections/method}

\input{sections/results}

\input{sections/conclusion}

\section*{Acknowledgements}
We thank Dana Cohen and Or Patashnik for their early feedback and insightful comments. We would also like to thank Harel Richardson for generously allowing us to use his toys throughout this paper. The beautiful meshes throughout this paper are taken from~\cite{wu20153d,keenan3D,turtle_mesh, teddy_bear_mesh, elephant_mesh, klein_mesh, threedscan, orangutan_mesh,  pika_mesh, michel2022text2mesh}.

{\small
\bibliographystyle{ieee_fullname}
\bibliography{egbib}
}

\clearpage

\input{figures/paint_additional_results/fig}
\input{figures/experiments/joint_fig.tex}

\input{figures/paint_more_results/fig}
\input{figures/images2mesh/fig_extra}
\input{figures/edits/fig}

\end{document}

%% file: packages.tex
\usepackage{graphicx}
\usepackage{amsmath}
\usepackage{amssymb}
\usepackage{booktabs}

\usepackage[pagebackref,breaklinks,colorlinks]{hyperref}

\usepackage[capitalize]{cleveref}

\usepackage{xcolor}

\usepackage{algorithm, algorithmic}

\usepackage{lipsum}  
\usepackage{pifont}
\usepackage{enumitem,amssymb}

\usepackage{physics}

%% file: macros.tex
\crefname{section}{Sec.}{Secs.}
\Crefname{section}{Section}{Sections}
\Crefname{table}{Table}{Tables}
\crefname{table}{Tab.}{Tabs.}

\definecolor{mydarkblue}{rgb}{0,0.08,1}
\definecolor{mydarkgreen}{rgb}{0.02,0.6,0.02}
\definecolor{mydarkorange}{rgb}{0.40,0.2,0.02}

\makeatletter
\DeclareRobustCommand\onedot{\futurelet\@let@token\@onedot}
\def\@onedot{\ifx\@let@token.\else.\null\fi\xspace}

\def\naive{na\"{\i}ve\xspace}
\def\naively{na\"{\i}vely\xspace}

\newif\ifdraft
\draftfalse

\definecolor{mygreen}{RGB}{112,173,71}
\definecolor{myblue}{RGB}{91,155,213}
\definecolor{myorange}{RGB}{237,125,49}
\definecolor{myred}{RGB}{255,22,67}
\definecolor{honey}{RGB}{201,89,95}
\definecolor{amber}{RGB}{180,148,41}
\definecolor{darkblue}{RGB}{84,96,126}

\newcommand{\keep}{{\color{mygreen} \textit{``keep''}}\xspace}
\newcommand{\refine}{{\color{myblue} \textit{``refine''}}\xspace}
\newcommand{\refined}
{{\color{myblue} \textit{``refined''}}\xspace}
\newcommand{\generate}
{{\color{myred} \textit{``generate''}}\xspace}

\newcommand{\shortcite}[1]{\cite{#1}}

%% file: figures/teaser/fig.tex
\begin{center}
\centering
    \centering
    \includegraphics[width=\linewidth]{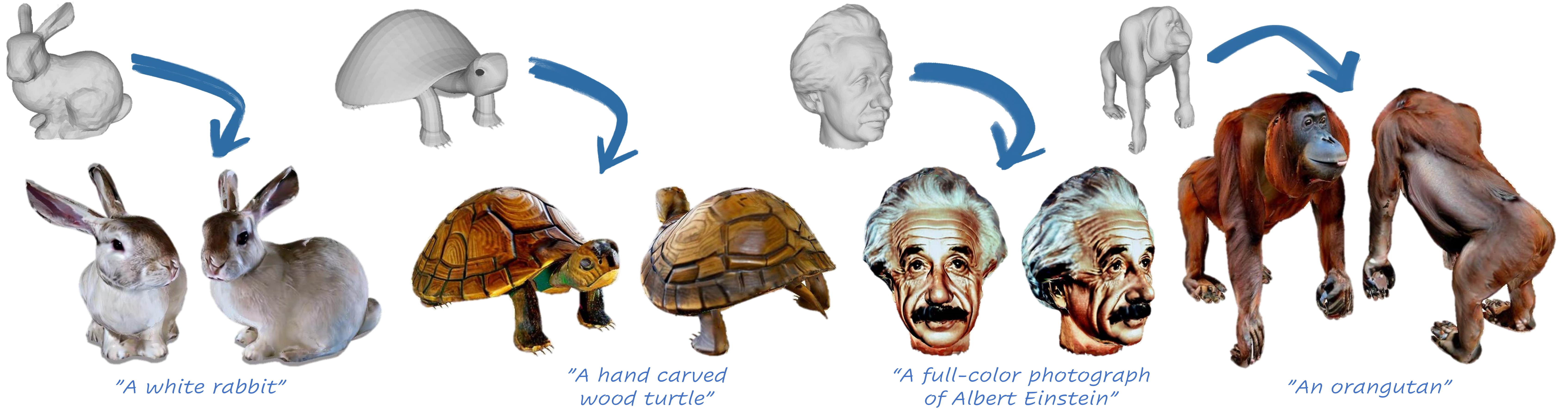}
    \vspace{-0.4cm}
    \captionof{figure}{
        Texturing results. TEXTure takes an input mesh and a conditioning text prompt and paints the mesh with high-quality textures. 
    }
    \label{fig:teaser}
\end{center}

%% file: sections/abstract.tex
In this paper, we present TEXTure, a novel method for text-guided generation, editing, and transfer of textures for 3D shapes.
Leveraging a pretrained depth-to-image diffusion model, TEXTure applies an iterative scheme that paints a 3D model from different viewpoints. Yet, while depth-to-image models can create plausible textures from a single viewpoint, the stochastic nature of the generation process can cause many inconsistencies when texturing an entire 3D object.
To tackle these problems, we dynamically define a trimap
partitioning of the rendered image into three progression states, and present a novel elaborated diffusion sampling process that uses this trimap representation to generate seamless textures from different views.
We then show that one can transfer the generated texture maps to new 3D geometries without requiring explicit surface-to-surface mapping, as well as extract semantic textures from a set of images without requiring any explicit reconstruction.
Finally, we show that TEXTure can be used to not only generate new textures but also edit and refine existing textures using either a text prompt or user-provided scribbles.
We demonstrate that our TEXTuring method excels at generating, transferring, and editing textures through extensive evaluation, and further close the gap between 2D image generation and 3D texturing.  Code is available on: \url{https://texturepaper.github.io/TEXTurePaper/}

%% file: sections/intro.tex
\section{Introduction}
\label{sec:intro}
The ability to paint pictures with words has long been a sign of a master storyteller, and with recent advancements in text-to-image models, this has become a reality for us all. Given a textual description, these new models are able to generate highly detailed imagery that captures the essence and intent of the input text. Despite the rapid progress in text-to-image generation, painting 3D objects remains a significant challenge as it requires to consider the specific shape of the surface being painted.
Recent works have begun making significant progress in painting and texturing 3D objects by using language-image models as guidance ~\cite{michel2022text2mesh,chen2022tango,xu2022dream3d,metzer2022latent}.
Yet, these methods still fall short in terms of quality compared to their 2D counterparts.

In this paper, we focus on texturing 3D objects, and present TEXTure, a technique that leverages diffusion models~\cite{rombach2021highresolution} to seamlessly paint a given 3D input mesh. 
Unlike previous texturing approaches~\cite{metzer2022latent,lin2022magic3d} that apply score distillation~\cite{poole2022dreamfusion} to indirectly utilize Stable Diffusion ~\cite{rombach2021highresolution} as a texturing prior, we opt to directly apply a full denoising process on rendered images using a depth-conditioned diffusion model~\cite{rombach2021highresolution}. 

At its core, our method iteratively renders the object from different viewpoints, applies a depth-based painting scheme, and projects it back to the mesh vertices or atlas. We show that our approach can result in a significant boost in both running time and generation quality. However, applying this process \naively would result in highly inconsistent texturing with noticeable seams due to the stochastic nature of the generation process (see~\Cref{fig:ablation_teaser} (A)).

To alleviate these inconsistencies, we introduce a dynamic partitioning of the rendered view to a trimap of \keep, \refine, and \generate regions, which is estimated  before each diffusion process.
The \generate regions are areas in the rendered viewpoint that are viewed for the first time and need to be painted. A \refine region is an area that was already painted in previous iterations, but is now seen from a better angle 
and should be repainted. Finally, \keep regions are painted regions that should not be repainted from the current view.
We then propose a modified diffusion process that takes into account our trimap partitioning. By freezing \keep regions during the diffusion process we attain more consistent outputs, but the newly generated regions still lack global consistency (see~\Cref{fig:ablation_teaser} (B)).
 To encourage better global consistency in the \generate regions, we further propose to  incorporate both depth-guided and mask-guided diffusion models into the sampling process (see~\Cref{fig:ablation_teaser} (C)). Finally, for \refine regions, we design a novel process that repaints these regions but takes into account their existing texture. 
Together these techniques allow the generation of highly-realistic results in mere minutes (see~\Cref{fig:ablation_teaser} (D) and~\Cref{fig:teaser} for results).

\input{figures/ablation_teaser/fig}

Next, we show that our method can be used not only to texture meshes guided by a text prompt, but also based on an existing textures from some other colored mesh or even from a small set of images.
Our method requires no surface-to-surface mapping or any intermediate reconstruction step. Instead, we propose to learn semantic tokens that represent a specific texture by building on Textual Inversion~\cite{gal2022textual_inversion} and DreamBooth~\cite{ruiz2022dreambooth}, while extending them to depth-conditioned models and introducing learned viewpoint tokens. We show that we can successfully capture the essence of a texture even from a few unaligned images and use it to paint a 3D mesh based on its semantic texture.

Finally, in the spirit of diffusion-based image editing~\cite{hertz2022prompt,huk2022shape,kawar2022imagic,tumanyan2022plug},  we show that one can further refine and edit textures. 
We propose two editing techniques. First, we present a text-only refinement where an existing texture map is modified using a guiding prompt to better match the semantics of the new text. Second, we illustrate how users can directly apply an edit on a texture map, where we refine the texture to fuse the user-applied edits into the 3D shape. 

We evaluate TEXTure and show its effectiveness for texture generation, transfer, and editing. We demonstrate that TEXTure offers a significant speedup compared to previous approaches, and more importantly, offers significantly higher-quality generated textures. 

%% file: figures/ablation_teaser/fig.tex
\begin{figure}[t]
    \centering
    \setlength{\tabcolsep}{0pt}
    {\small
    \begin{tabular}{c c c c}
        \includegraphics[width=0.20\linewidth,trim={10cm 8cm 10cm 6cm},clip]{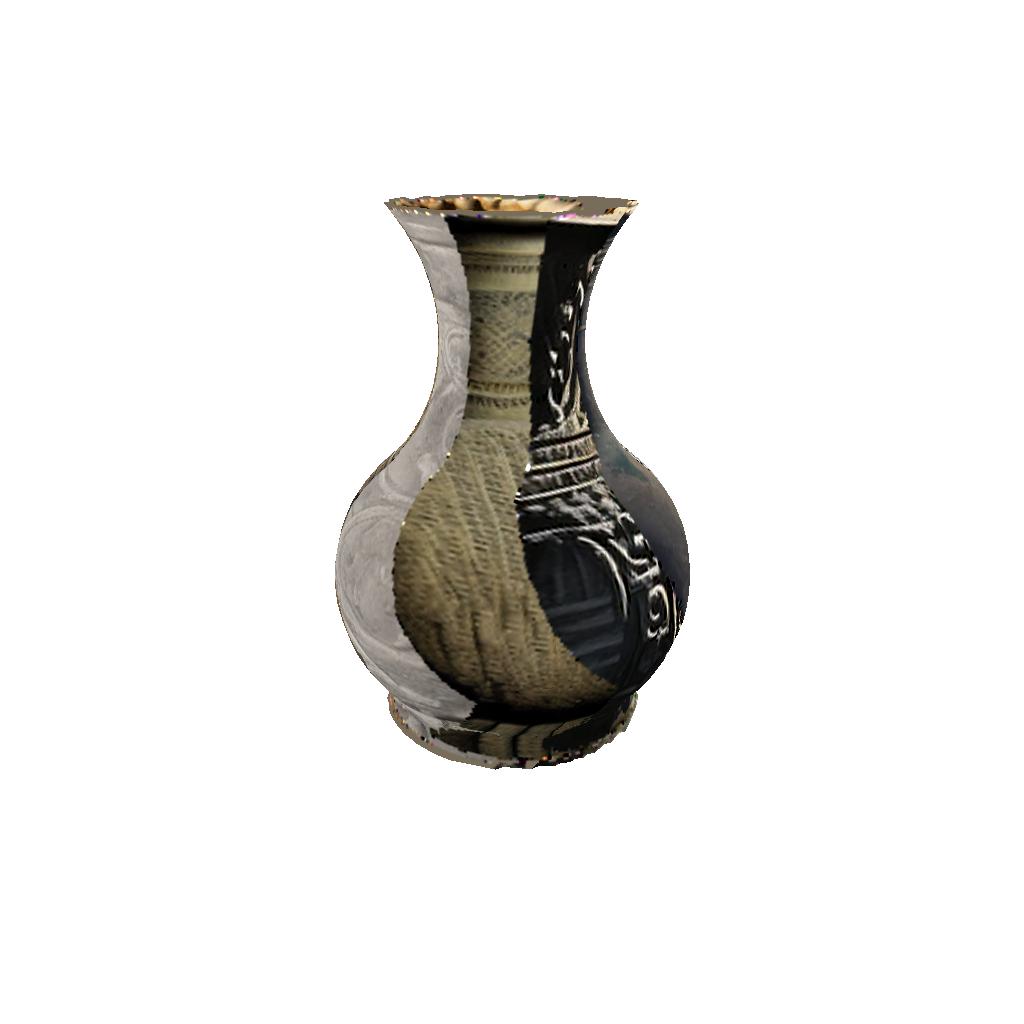} &
        \includegraphics[width=0.20\linewidth,trim={10cm 8cm 10cm 6cm},clip]
        {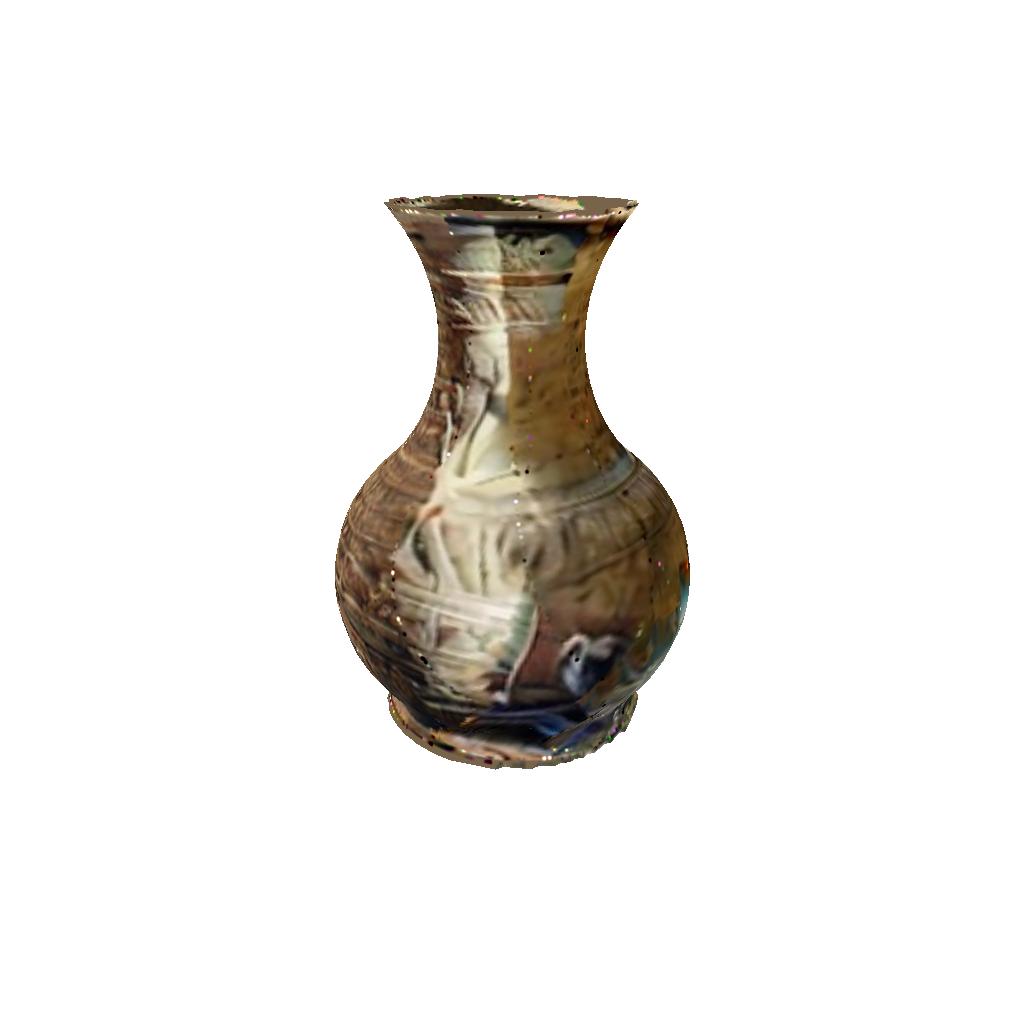} &
        \includegraphics[width=0.20\linewidth,trim={10cm 8cm 10cm 6cm},clip]
        {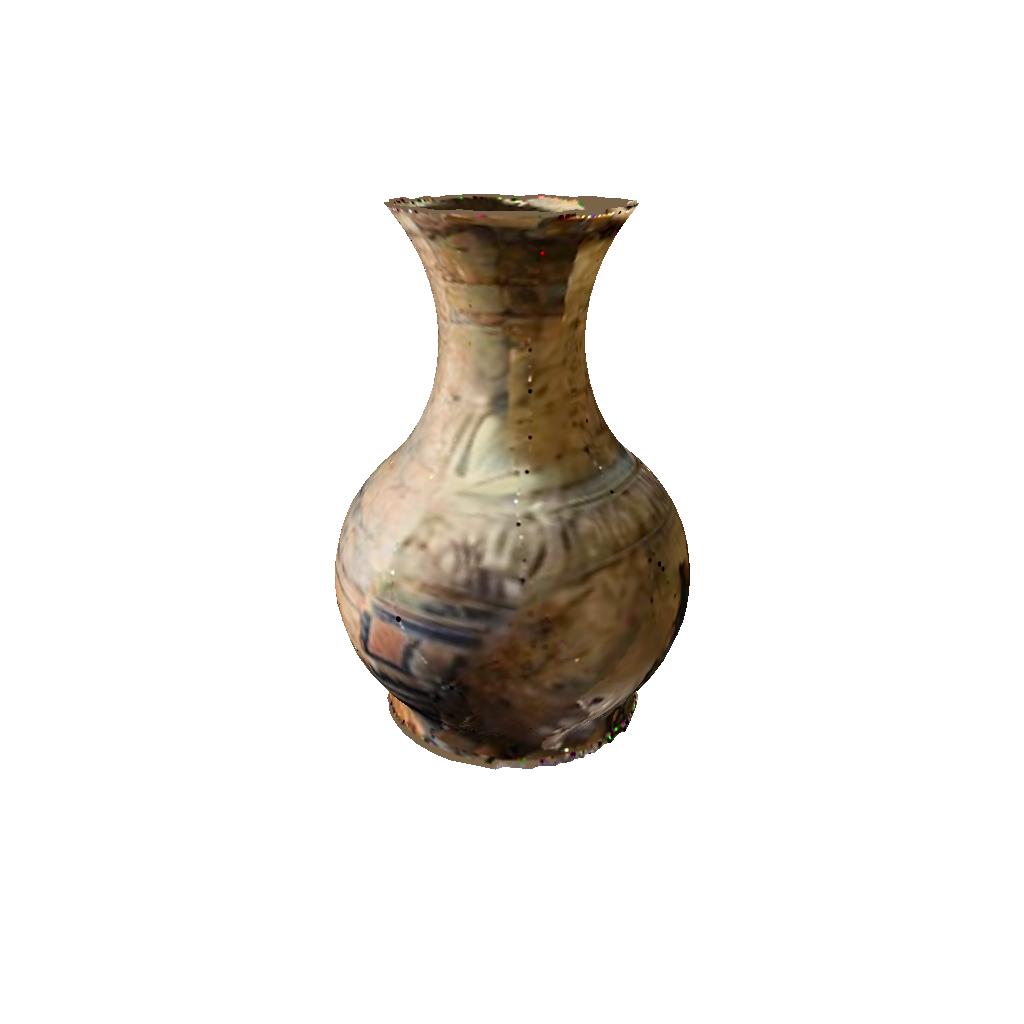} &
        \includegraphics[width=0.20\linewidth,trim={10cm 8cm 10cm 6cm},clip]{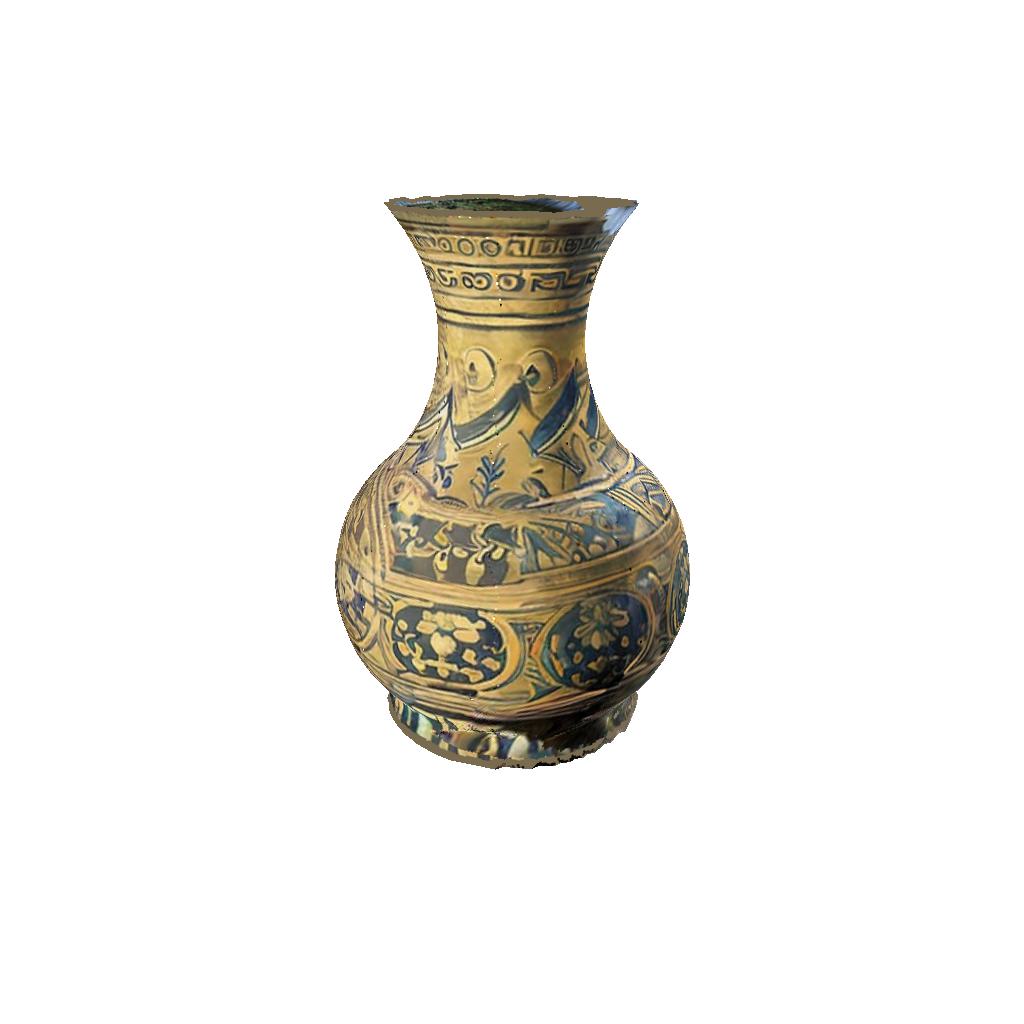} \\
        A & B & C & D
    \end{tabular}
    }
    \vspace{-0.35cm}
    \caption{Ablation of our different components: (A) is a \naive painting scheme. In (B)  we introduce \keep region. (C) is our improved scheme for \generate regions, and (D) is our complete scheme with \refine regions.}
    \label{fig:ablation_teaser}
    \vspace{-0.4cm}
\end{figure}

%% file: sections/relatedwork.tex
\vspace{-0.1cm}

\section{Related Work}
\label{sec:related}

\paragraph{\textbf{Text-to-Image Diffusion Models.}}
The past year has seen the development of multiple large diffusion models~\cite{imagen,rombach2021highresolution,ramesh2022hierarchical,nichol2021glide} capable of producing impressive images with pristine details guided by an input text prompt.
The widely popular Stable Diffusion~\cite{rombach2021highresolution}, is trained on a rich text-image dataset~\cite{schuhmann2022laion} and is conditioned on CLIP's~\cite{radford2021learning} frozen text encoder. 
Beyond simple text-conditioning, Stable Diffusion has multiple extensions that allow conditioning its denoising network on additional input modalities such as a depth map or an inpainting mask. Given a guiding prompt and an estimated depth image~\cite{Ranftl2021}, the depth-conditioning model is tasked with generating images that follow the same depth values while being semantically faithful with respect to the text.
Similarly, the inpainting model completes missing image regions, given a masked image. 

Although current text-to-image models generate high-quality results when conditioned on a text prompt or depth map, editing an existing image or injecting objects specified by a few exemplar images remains challenging ~\cite{avrahami2022blended_latent,song2022objectstitch}.
To introduce a user-specific concept to a pre-trained text-to-image model, \cite{gal2022textual_inversion} introduce Textual Inversion to map a few exemplar images into a learned pseudo-tokens in the embedding space of the frozen text-to-image model. DreamBooth~\cite{ruiz2022dreambooth} further fine-tune the entire diffusion model on the set of input images to achieve more faithful compositions. 
The learned token or fine-tuned model can then be used to generate novel images using the custom token in new user-specified text prompts.

\paragraph{\textbf{Texture and Content Transfer.}}
Early works~\cite{efros1999texture, zhu1998filters, heeger1995pyramid, de1997multiresolution} focus on 2D texture synthesis through probabilistic models, while more recent works~\cite{zhou2018non, Xian2017TextureGANCD, Frhstck2019TileGANSO, Sendik2017DeepCF} take a data-driven approach to generate textures using deep neural networks.
Generating textures over 3D surfaces is a more challenging problem, as it requires attention to both colors and geometry. 
For geometric texture synthesis, \cite{golovinskiy2006statistical} applies a similar statistical method to \cite{heeger1995pyramid} while \cite{breckon2012hierarchical} extended non-parametric sampling proposed by \cite{efros1999texture} to 3D meshes.
\cite{berkiten2017learning} introduces a metric learning approach to transfer details from a source to a target shape, while \cite{Hertz2020deep} use an internal learning technique to transfer geometric texture.
For 3D color texture synthesis, given an exemplar colored mesh, \cite{mertens2006texture, lu2007context, chen2012non} use the relation between geometric features and color values to synthesize new textures on target shapes.

\paragraph{\textbf{3D Shape and Texture Generation.}}
Generating shapes and textures in 3D has recently gained significant interest.
Text2Mesh~\cite{michel2022text2mesh}, Tango~\cite{chen2022tango}, and CLIP-Mesh~\cite{khalid2022clipmesh} use CLIP-space similarities as an optimization objective to generate novel shapes and textures. CLIP-Mesh deforms an initial sphere with UV texture parameterization. Tango optimizes per-vertex colors and focuses on generating novel textures. Text2Mesh optimizes per-vertex color attributes, while allowing small geometric displacements.
Get3D~\cite{gao2022get3d} is trained to generate shape and texture through a DMTet~\cite{shen2021dmtet} mesh extractor and 2D adversarial losses.

Recently, DreamFusion~\cite{poole2022dreamfusion} introduced the use of pre-trained image diffusion models for generating 3D NeRF models conditioned on a text prompt. The key component in DreamFusion is the \textit{Score-Distillation} loss which enables the use of a pretrained 2D diffusion model as a critique for optimizing the 3D NeRF scene.
Recently, Latent-NeRF~\cite{metzer2022latent} showed how the same Score-Distillation loss can be used in Stable Diffusion's latent space to generate latent 3D NeRF models. In the context for texture generation,  \cite{metzer2022latent} present \textit{Latent-Paint}, a texture generation technique, where latent texture maps are painted using Score-Distillation and are then decoded to RGB for the final colorization output. Similarly, \cite{lin2022magic3d} uses score-distillation to texture and refine a coarse initial shape. Both methods suffer from relatively slow convergence and less defined textures compared to our proposed approach.

%% file: sections/method.tex
\section{Method}
\label{sec:method}
We first lay the foundation for our text-guided mesh texturing scheme, illustrated in~\Cref{fig:texturing_pipeline}.
Our TEXTure scheme performs an incremental texturing of a given 3D mesh, where at each iteration we paint the currently visible regions of the mesh as seen from a single viewpoint. To encourage both local and global consistency, we segment the mesh into a trimap of \keep, \refine, \generate regions. A  modified depth-to-image diffusion process is presented to incorporate this information into the denoising steps.

We then propose two extensions of TEXTure.
First, we present a texture transfer scheme (\Cref{sec:texture_transfer}) that transfers the texture of a given mesh to a new mesh, by learning a custom concept that represents the given texture.
Finally, we present a texture editing technique that allows users to edit a given texture map, either through a guiding text prompt or a user-provided scribble (\Cref{sec:texture_editing}). 

\input{figures/texturing_pipeline/fig.tex}

\subsection{Text-Guided Texture Synthesis}
\label{subsec:pure_text_guided}
Our texture generation method relies on a pretrained depth-to-image diffusion model $\mathcal{M}_{depth}$ and a pretrained inpainting diffusion model $\mathcal{M}_{paint}$, both based on Stable Diffusion~\cite{rombach2021highresolution} and with a shared latent space. 
During the generation process, the texture is represented as an atlas through a UV mapping that is calculated using XAtlas~\cite{xatlas}.

We start from an arbitrary initial viewpoint $v_0=(r=1.25, \phi_0=0, \theta=60)$ where $r$ is the radius of the camera, $\phi$ is the azimuth camera angle, and $\theta$ is the camera elevation. We then use $\mathcal{M}_{depth}$ to generate an initial colored image $I_0$ of the mesh as viewed from $v_0$, conditioned on the rendered depth map $\mathcal{D}_0$.
The generated image $I_0$ is then projected back to the texture atlas $\mathcal{T}_0$ to color the shape's visible parts from $v_0$.
Following this initialization step, we begin a process of incremental colorization, illustrated in~\Cref{fig:texturing_pipeline}, where we iterate through a fixed set of viewpoints.
For each viewpoint, we then render the mesh using a renderer $\mathcal{R}$~\cite{KaolinLibrary} to obtain $\mathcal{D}_t$ and $Q_t$, where $Q_t$ is the rendering of the mesh as seen from the viewpoint $v_t$ that considers all previous colorization steps. Finally, we generate the next image $I_t$ and project $I_t$ back to the updated texture atlas $\mathcal{T}_t$ while taking into account $Q_t$.

Once a single view has been painted, the generation task becomes more challenging due to the need for local and global consistency along the generated texture. Below we consider a single iteration $t$ of our incremental painting process and elaborate on our proposed techniques to handle these challenges.

\vspace{-0.1cm}
\paragraph{\textbf{Trimap Creation.}} 
Given a viewpoint $v_t$, we first apply a partitioning of the rendered image into three regions: \keep, \refine, and \generate. The \generate regions are rendered areas that are viewed for the first time and need to be painted to match the previously painted regions. The distinction between \keep and \refine regions is slightly more nuanced and is based on the fact that coloring a mesh from an oblique angle can result in high distortion. This is because the cross-section of a triangle with the screen is low, resulting in a low-resolution update to the mesh texture image $\mathcal{T}_t$. Specifically, we measure the triangle's cross-section as the $z$ component of the face normal $n_z$ in the camera's coordinate system.

Ideally, if the current view provides a better colorization angle for some of the previously-painted regions, we would like to \refine their existing texture.
Otherwise, we should \keep the original texture and avoid modifying it to ensure consistency with previous views.
To keep track of seen regions and the cross-section at which they were previously colored from, we use an additional meta-texture map $\mathcal{N}$ that is updated at every iteration. This additional map can be efficiently rendered together with the texture map at each iteration and is used to define the current trimap partitioning.

\vspace{-0.1cm}
\paragraph{\textbf{Masked Generation.}} 
As the depth-to-image diffusion process was trained to generate an entire image, we must modify the sampling process to \keep part of the image fixed. Following Blended Diffusion~\cite{avrahami2022blended,avrahami2022blended_latent}, in each denoising step we explicitly inject a noised versions of $Q_t$, \textit{i.e.} $z_{Q_t}$, at the \keep regions into the diffusion sampling process, such that these areas are seamlessly blended into the final generated result. Specifically, the latent at the current sampling timestep $i$ is computed as
\begin{equation}~\label{eq:blended}
    z_i \leftarrow z_i \odot m_{blended} + z_{Q_t} \odot (1 - m_{blended})
\end{equation}
where the mask $m_{blended}$ is defined in Equation~\ref{eq:m_blended}. That is, for \keep regions, we simply set $z_i$ fixed according to their original values.

\vspace{-0.1cm}
\paragraph{\textbf{Consistent Texture Generation.}}
Injecting \keep regions into the diffusion process results in better blending with \generate regions. Still, when moving away from the \keep boundary and deeper into the \generate regions, the generated output is mostly governed by the sampled noise and is not consistent with previously painted regions. We first opt to use the same sampled noise from each viewpoint, this sometimes improves consistency, but is still very sensitive to the change in viewpoint.
We observe that applying an inpainting diffusion model $\mathcal{M}_{paint}$ that was directly trained to complete masked regions, results in more consistent generations. 
However, this in turn deviates from the conditioning depth $\mathcal{D}_t$ and may generate new geometries. To benefit from the advantages of both models we introduce an interleaved process where we alternate between the two models during the initial sampling steps. Specifically, during  sampling, the next noised latent $z_{i-1}$ is computed as:
\begin{equation*}
    z_{i-1} = \begin{cases}
        \mathcal{M}_{depth}(z_i, \mathcal{D}_t) & 0 \le i < 10 \\
        \mathcal{M}_{paint}(z_i, \text{\generate}) & 10 \le i < 20 \\
        \mathcal{M}_{depth}(z_i, \mathcal{D}_t) & 20 \le i < 50
        \end{cases}
\end{equation*}
When applying $\mathcal{M}_{depth}$, the noised latent is guided by the current depth $\mathcal{D}_t$ while when applying $\mathcal{M}_{paint}$, the sampling process is tasked with completing the \generate regions in a globally-consistent manner. 

\vspace{-0.1cm}
\paragraph{\textbf{Refining Regions.}} To handle \refine regions we use another novel modification to the diffusion process that generates new textures while taking into account their previous values. Our key observation is that by using an alternating checkerboard-like mask in the first steps of the sampling process, we can guide the noise towards values that locally align with previous completions. 

The granularity of this process can be controlled by changing the resolution of the checkerboard mask and the number of constrained steps. In practice, we apply the mask for the first $25$ sampling steps. Namely, the masked $m_{blended}$ applied in~\Cref{eq:blended} is set as,
\begin{equation}
    m_{blended} = \begin{cases}
        0 &\text{\keep} \\
        \text{checkerboard} &\text{\refine} \wedge i\le 25 \\
        1 &\text{\refine} \wedge i > 25 \\
        1 &\text{\generate}
        \end{cases}
    \label{eq:m_blended}
\end{equation}
where a value of $1$ indicates that this region should be painted and kept otherwise. Our blending mask is visualized in~\Cref{fig:texturing_pipeline}.

\vspace{-0.1cm}
\paragraph{\textbf{Texture Projection.}}
To project $I_t$ back to the texture atlas $\mathcal{T}_t$, we apply gradient-based optimization for $\mathcal{L}_t$ over the values of $\mathcal{T}_t$ when rendered through the differential renderer $\mathcal{R}$. That is,
\begin{equation*}
    \nabla_{\mathcal{T}_t} \mathcal{L}_t = [(\mathcal{R}(mesh, \mathcal{T}_t, v_t) - I_t) \odot m_s] \frac{\partial \mathcal{R}\odot m_s}{\partial \mathcal{T}_t}
    \label{eq:loss_at_t}
\end{equation*}

To achieve smoother texture seams of the projections from different views, a soft mask $m_s$ is applied at the boundaries of the \refine and \generate region:
\begin{equation*}
    \;\;\;\; m_s = m_h * g \;\;\;\;\;\;\;\;  m_h = \begin{cases}
        0 &\text{\keep} \\
        1 &\text{\refine $\cup$ \generate}
        \end{cases}
    \label{eq:hard_mask}
\end{equation*}
where $g$ is a 2D Gaussian blur kernel.

\vspace{-0.1cm}
\paragraph{\textbf{Additional Details.}}
Our texture is represented as a $1024\times1024$ atlas, where the rendering resolution is $1200\times1200$. For the diffusion process, we segment the inner region, resize it to $512\times512$ and mat it onto a realistic background. All shapes are rendered with $8$ viewpoints around the object, and two additional top/bottom views. We show that viewpoint order can also affect the end results.
\input{figures/transfer_pipeline/fig.tex}
\subsection{Texture Transfer}~\label{sec:texture_transfer}
Having successfully generated a new texture on a given 3D mesh, we now turn to describe how to \textit{transfer} a given texture to a new, untextured target mesh. We show how to capture textures from either a painted mesh, or from a small set of input images.
Our texture transfer approach builds on previous work on concept learning over diffusion models~\cite{ruiz2022dreambooth,gal2022textual_inversion}, by fine-tuning a pretrained diffusion model and learning a pseudo-token representing the generated texture. The fine-tuned model is then used for texturing a new geometry.
To improve the generalization of the fine-tuned model to new geometries, we further propose a novel \textit{spectral augmentation} technique, described next. We then discuss our concept learning scheme from meshes or images.

\vspace{-0.2cm}
\paragraph{\textbf{Spectral Augmentations.}}
Since we are interested in learning a token representing the input \textbf{texture} and not the original input geometry itself, we should ideally learn a common token over a range of geometries containing the input texture.
Doing so disentangles the texture from its specific geometry and improves the generalization of the fine-tuned diffusion model.
Inspired by the concept of surface caricaturization~\cite{sela2015computational}, we propose a novel \textit{spectral augmentation} technique. In our case, we propose random low-frequency geometric deformations to the textured source mesh, regularized by the mesh Laplacian's spectrum~\cite{meyer2003discrete}.

Modulating random deformations over the spectral eigenbasis results in smooth deformations that keep the integrity of the input shape.
Empirically, we choose to apply random inflations or deflations to the mesh, with a magnitude proportional to a randomly selected eigenfunction.
We provide examples of such augmentations in~\Cref{fig:transfer_pipeline}, and additional details in the supplementary materials.

\input{figures/paint_results/fig_single.tex}
\vspace{-0.1cm}
\paragraph{\textbf{Texture Learning.}}
Applying our spectral augmentation technique, we generate a large set of images with corresponding depth maps of the input shape. We render the images from several viewpoints (left, right, overhead, bottom, front, and back) and paste the rendered object onto a randomly colored background (see~\Cref{fig:transfer_pipeline}).

Given the set of rendered images, we follow~\cite{gal2022textual_inversion} and optimize an embedding vector representing our texture using prompts of the form ``a $\langle D_v \rangle$ photo of a $\langle \mathcal{S}_{texture} \rangle$'' where $\langle D_v \rangle$ is a learned token representing the view direction of the rendered image and $\langle \mathcal{S}_{texture} \rangle$ is the token representing our texture.
Observe, that we have six learned directional tokens $D_v$, shared within images from the same view, and a single token $\mathcal{S}_{texture}$ representing the texture, shared across all images.
Additionally, to better capture the input texture we fine-tune the diffusion model itself as well, as done in~\cite{ruiz2022dreambooth}.
Our texture learning scheme is illustrated  in~\Cref{fig:transfer_pipeline}.
After training, we use TEXTure (Section~\ref{subsec:pure_text_guided}), to color the target shape, swapping the original Stable Diffusion model with our fine-tuned model.

\vspace{-0.1cm}
\paragraph{\textbf{Texture from Images.}}
Next, we explore the more challenging task of texture generation based on a small set of sample images. While we cannot expect the same quality given only a few images, we can still potentially learn concepts that represent different textures. Unlike standard textual inversion techniques~\cite{gal2022textual_inversion,ruiz2022dreambooth}
our learned concepts represent mostly texture and not structure as they are trained on a depth-conditioned model. This potentially makes them more suitable for texturing other 3D shapes. 

For this task, we segment the prominent object from the image using a pretrained saliency network~\cite{qin2020u2}, apply standard scale and crop augmentations, and paste the result onto a randomly-colored background.
Our results show that one can successfully learn semantic concepts from images and apply them to 3D shapes without any explicit reconstruction stage in between. We believe this creates new opportunities for creating captivating textures inspired by real objects.

\subsection{Texture-Editing}~\label{sec:texture_editing}
We show that our trimap-based TEXTuring can be used to easily extend 2D editing techniques to a full mesh. For text-based editing, we wish to alter an existing texture map guided by a textual prompt. To this end, we define the entire texture map as a \refine region and apply our TEXTuring process to modify the texture to align with the new text prompt. We additionally provide scribble-based editing where a user can directly edit a given texture map (\textit{e.g.} to define a new color scheme over a desired region). To allow this, we simply define the altered regions as \refine regions during the TEXTuring process and \keep the remaining texture fixed.

%% file: figures/texturing_pipeline/fig.tex
\begin{figure}
    \centering
    \includegraphics[width=0.99\columnwidth]{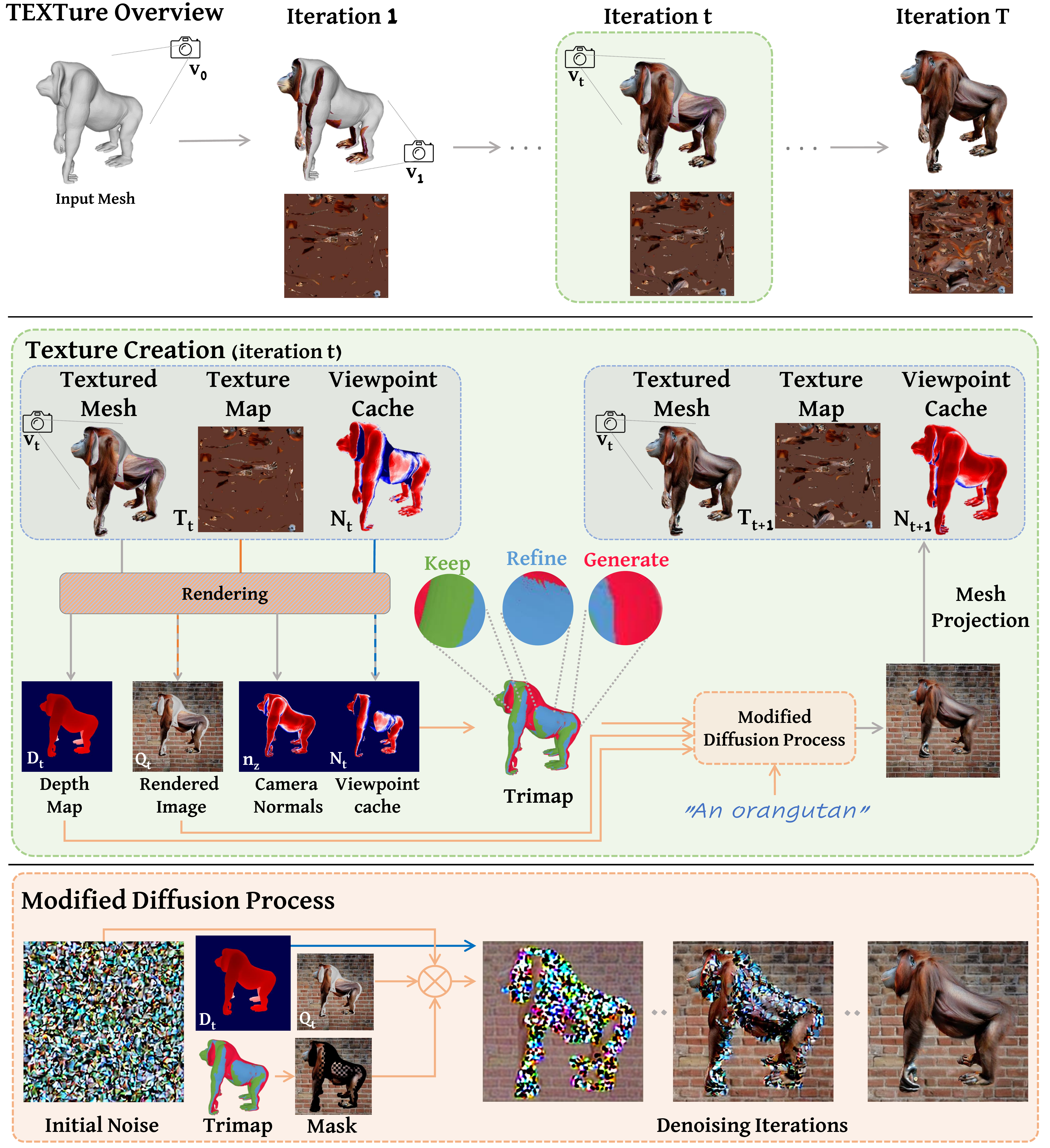}
    \vspace{-0.25cm}
    \caption{Our schematic texturing process. A mesh is iteratively painted from different viewpoints. In each painting iteration, we render the mesh alongside its depth and normal maps. We then calculate a trimap partitioning of the image to three distinct areas based on the camera normals and a viewpoint cache representing previously viewed angles. These inputs are then fed into a modified diffusion process alongside a given text prompt which generates an updated image. This image is then projected back to the texture map for the next iteration.}
    \vspace{-0.5cm}
    \label{fig:texturing_pipeline}
\end{figure} 

%% file: figures/transfer_pipeline/fig.tex
\begin{figure}
    \centering
    \includegraphics[width=0.985\columnwidth]{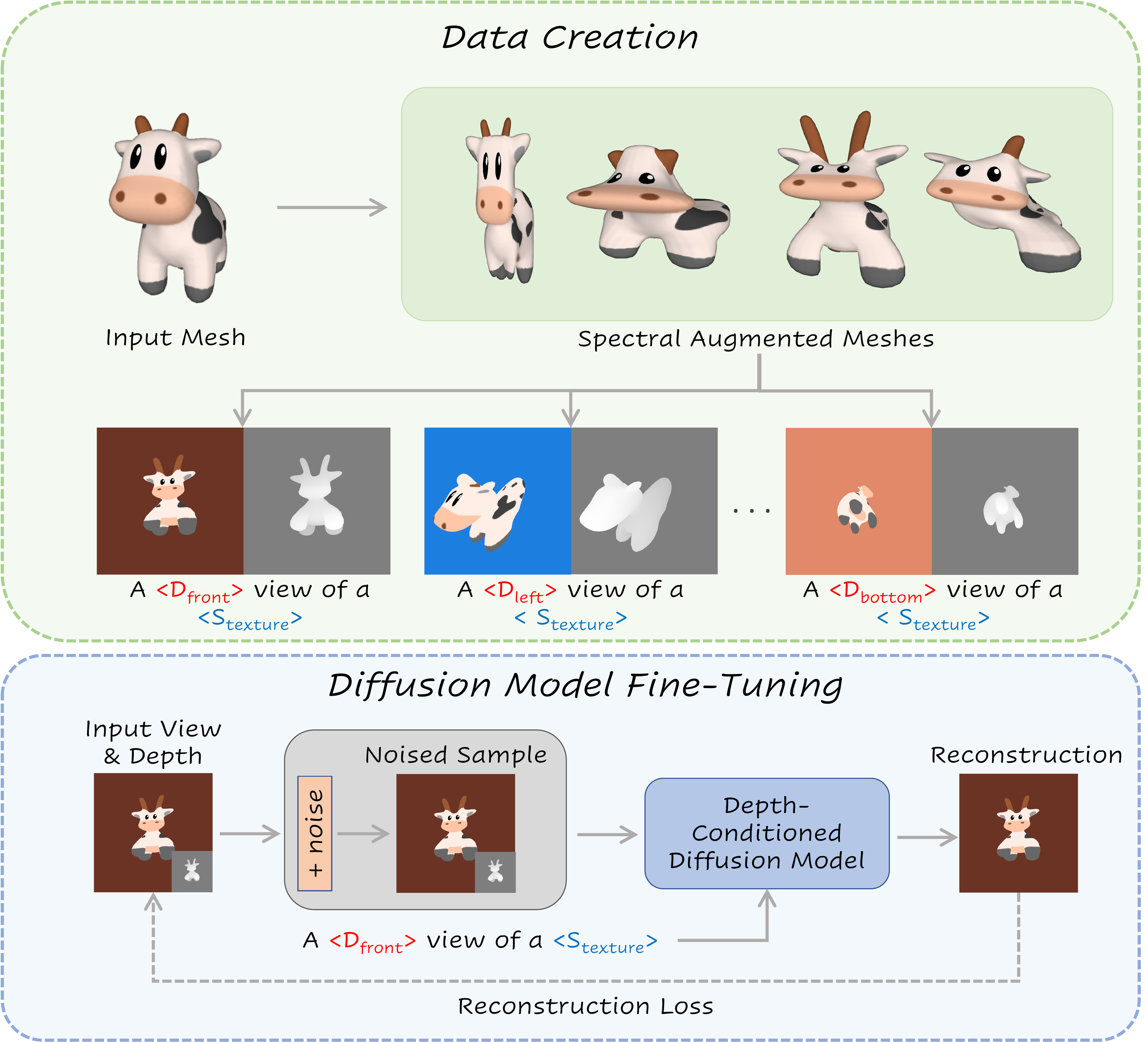}
    \vspace{-0.3cm}
    \caption{Fine-Tuning for texture transfer. (top) Given an input mesh, spectral augmentations are applied to generate a variety of textured geometries. The geometries are then rendered from random viewpoints, where each image is coupled with a sentence, based on the viewpoint.
    (bottom) a depth-conditioned diffusion model is then fine-tuned with a set of learnable tokens, a fixed $\langle \mathcal{S}_{texture} \rangle$ token representing the object, and an additional viewpoint token,  $\langle \mathcal{D}_{v} \rangle$. The tuned model is used to paint new objects.  }
    \vspace{-0.3cm}
    \label{fig:transfer_pipeline}
\end{figure} 

%% file: figures/paint_results/fig_single.tex
\begin{figure*}[t]
\centering
    \centering
    \setlength{\tabcolsep}{0pt}
    {\small
    \begin{tabular}{c c c c c c c c c}
        \includegraphics[height=0.11\linewidth,trim={11cm 10cm 6.5cm 11cm},clip]{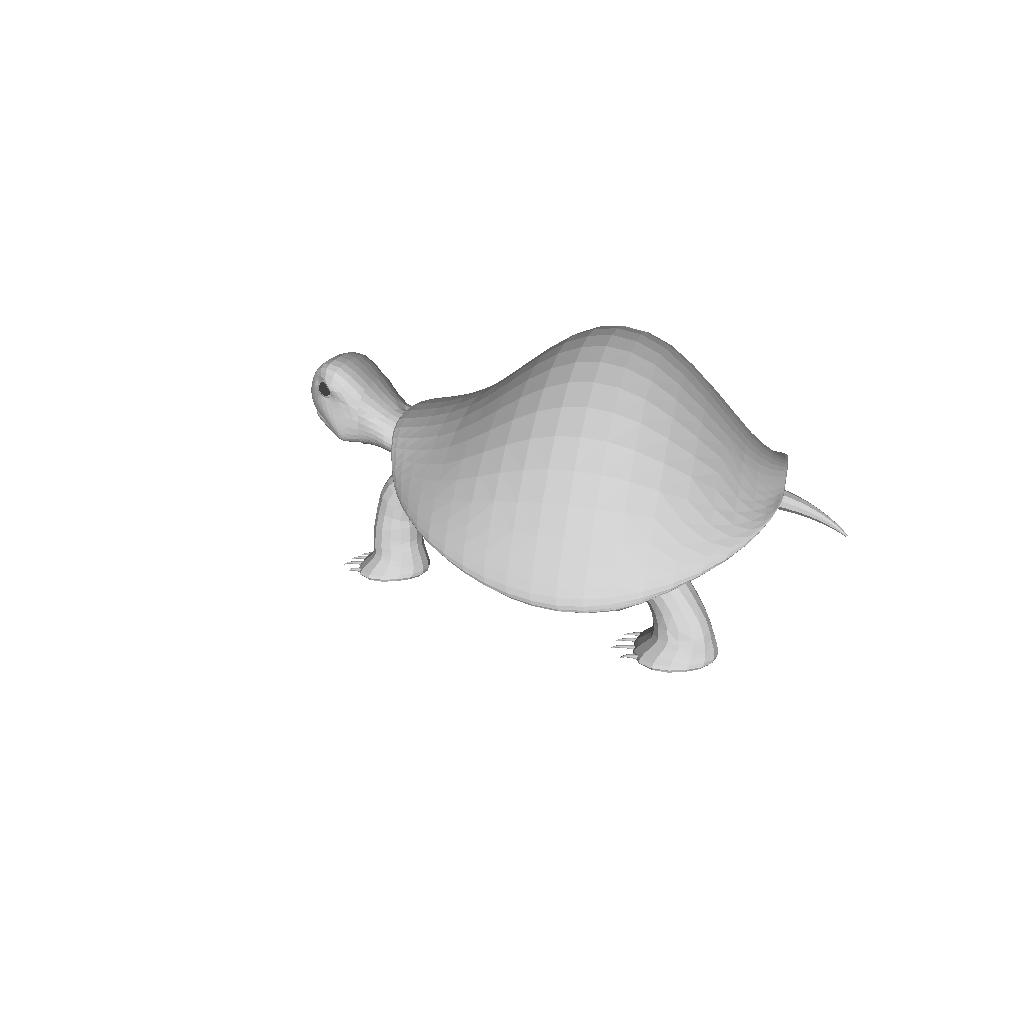} &
        \includegraphics[height=0.11\linewidth,trim={11cm 10cm 6.5cm 11cm},clip]{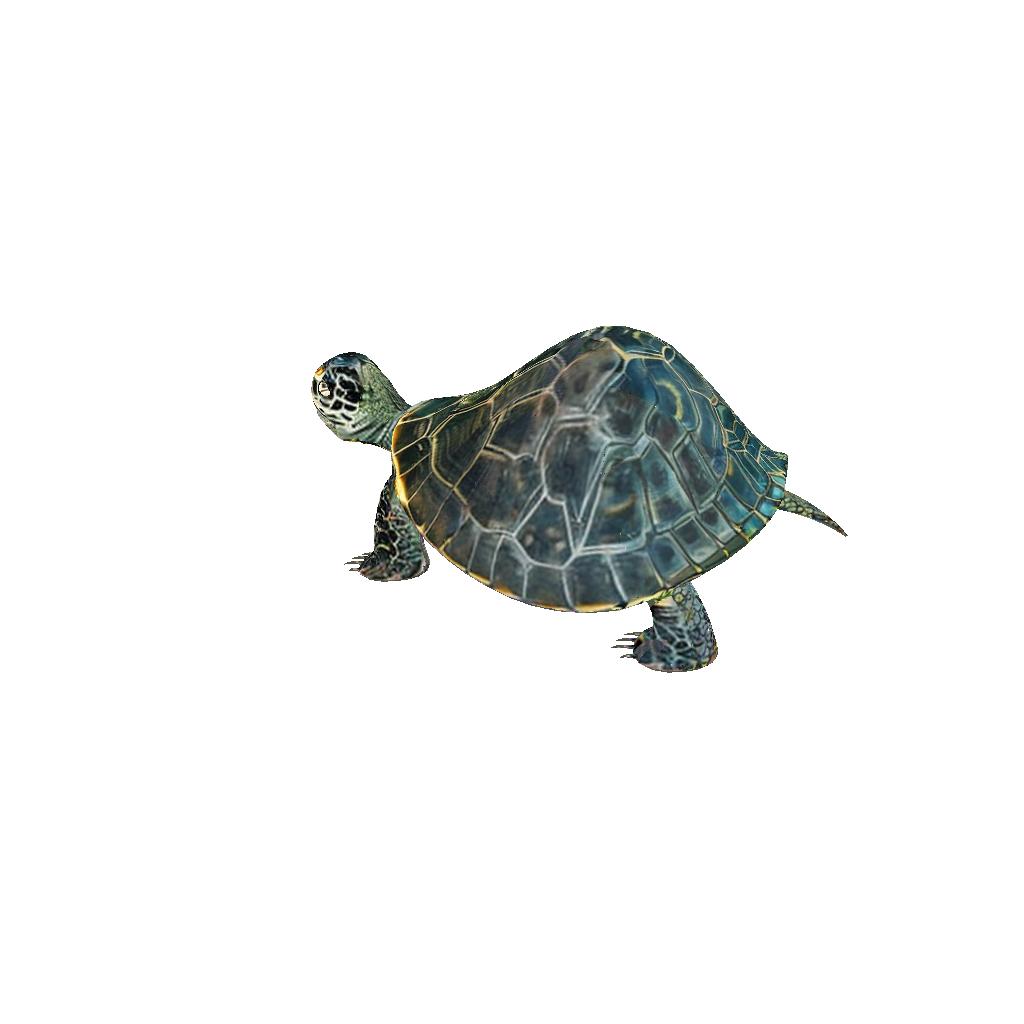}
        \includegraphics[height=0.11\linewidth,trim={9.5cm 12cm 10cm 10cm},clip]{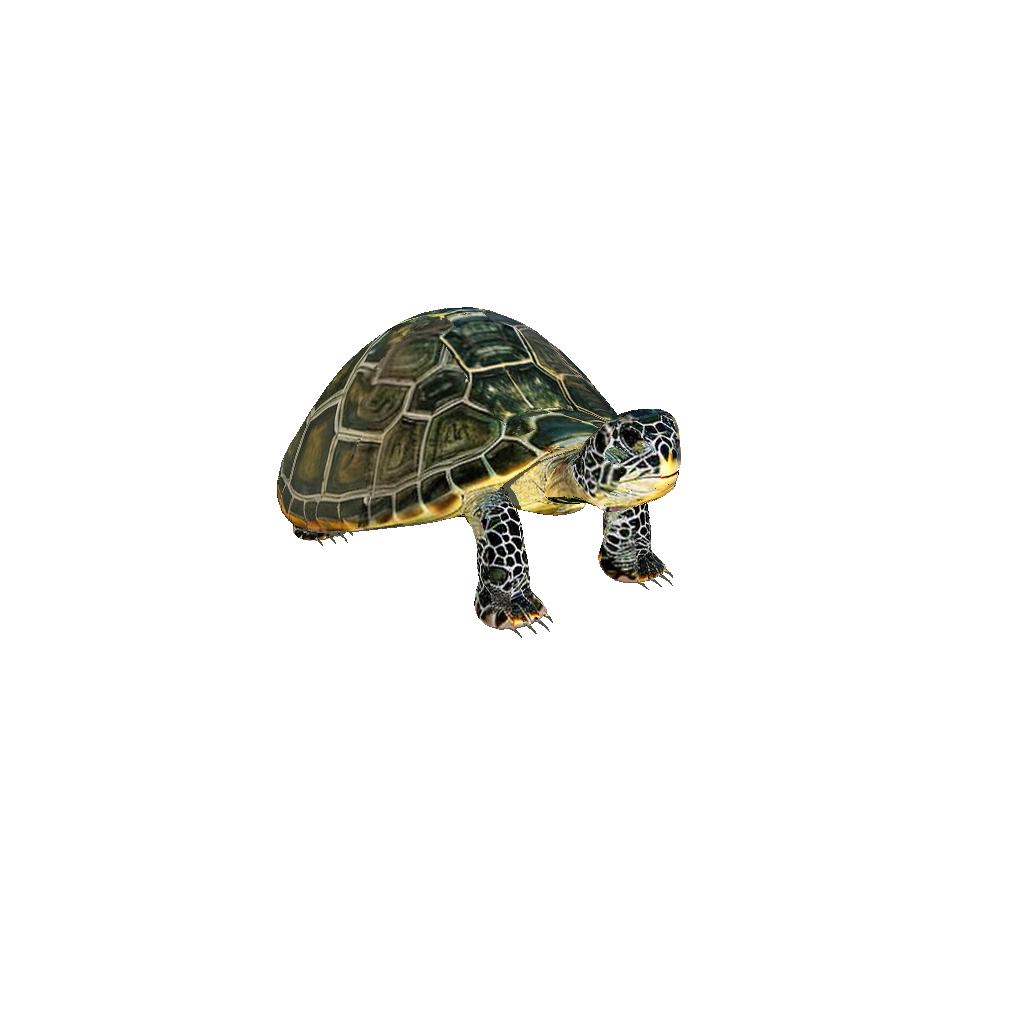} &
        \hspace{0.05cm}
        \includegraphics[height=0.12\linewidth,trim={9cm 7cm 11cm 8cm},clip]{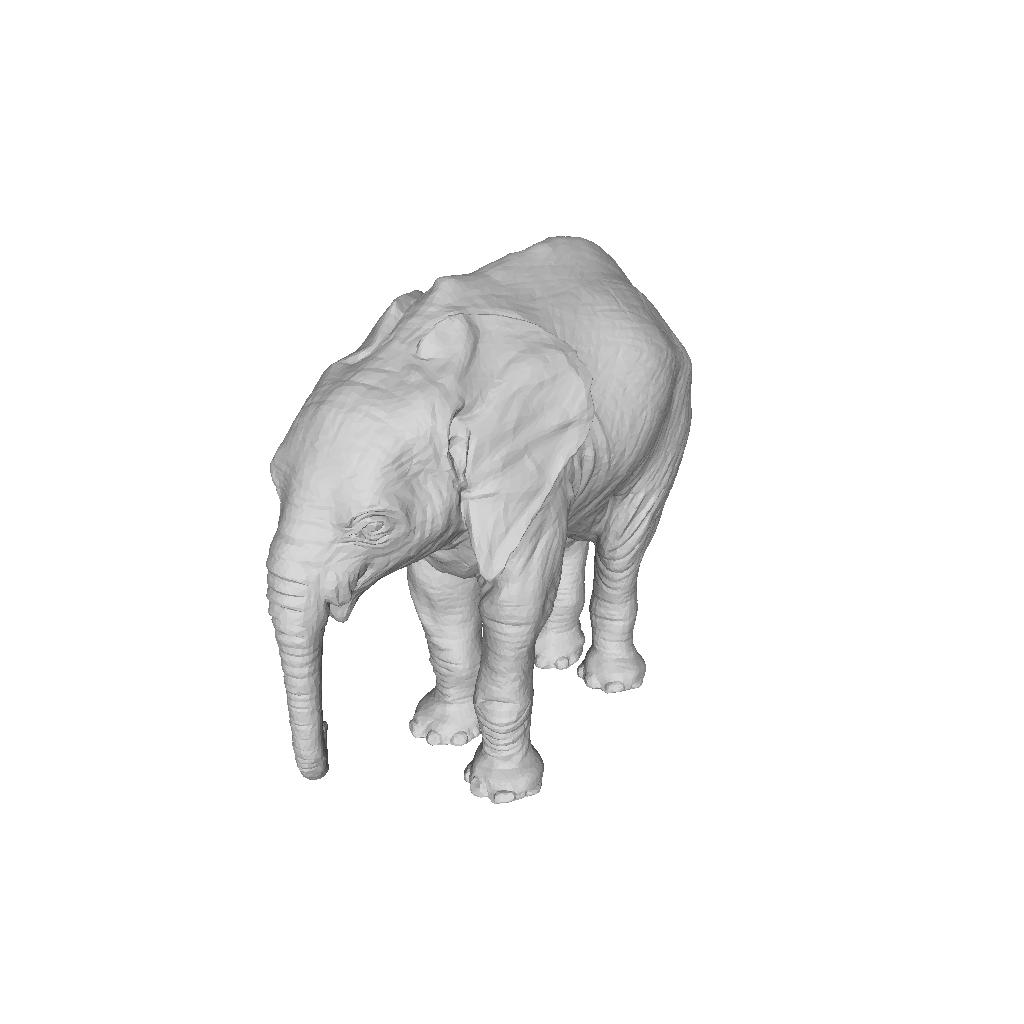} &
        \includegraphics[height=0.12\linewidth,trim={9cm 7cm 11cm 8cm},clip]{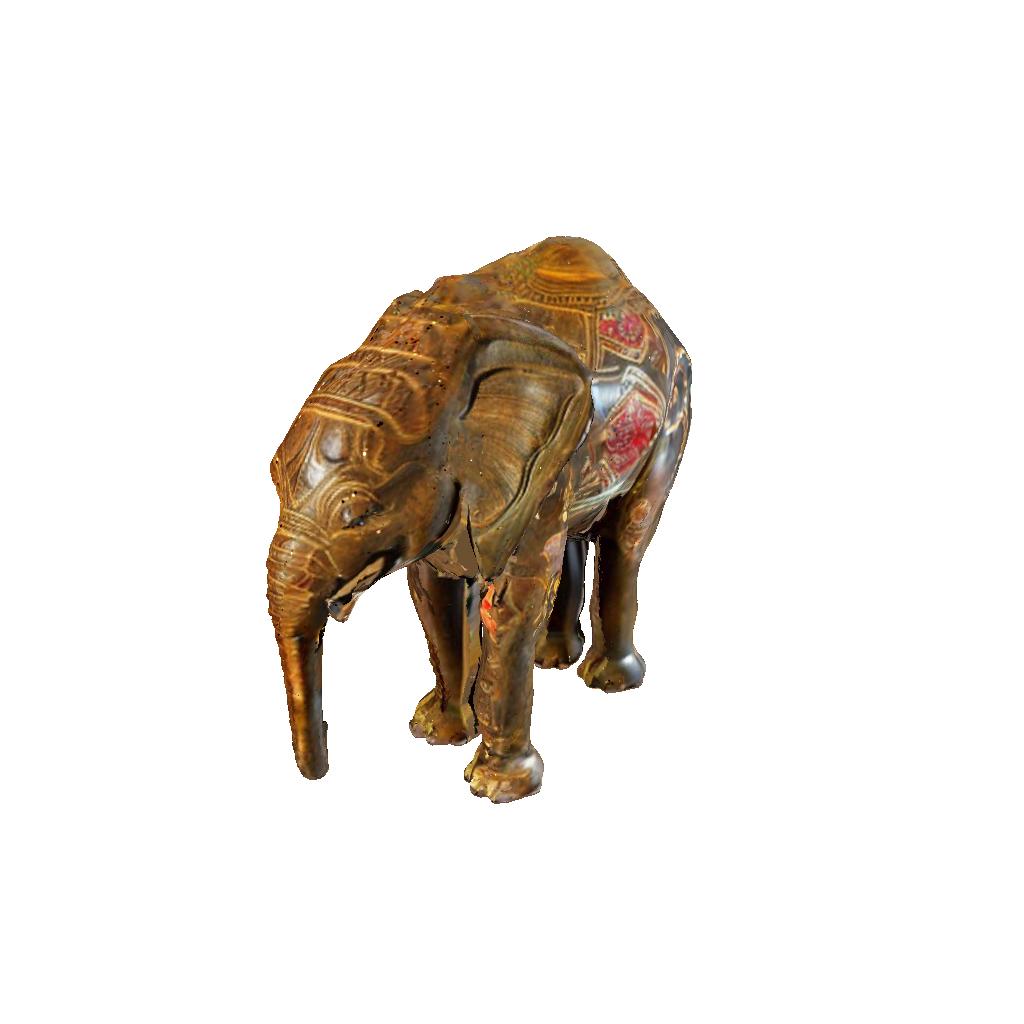}
        \includegraphics[height=0.12\linewidth,trim={7cm 7cm 9cm 9cm},clip]{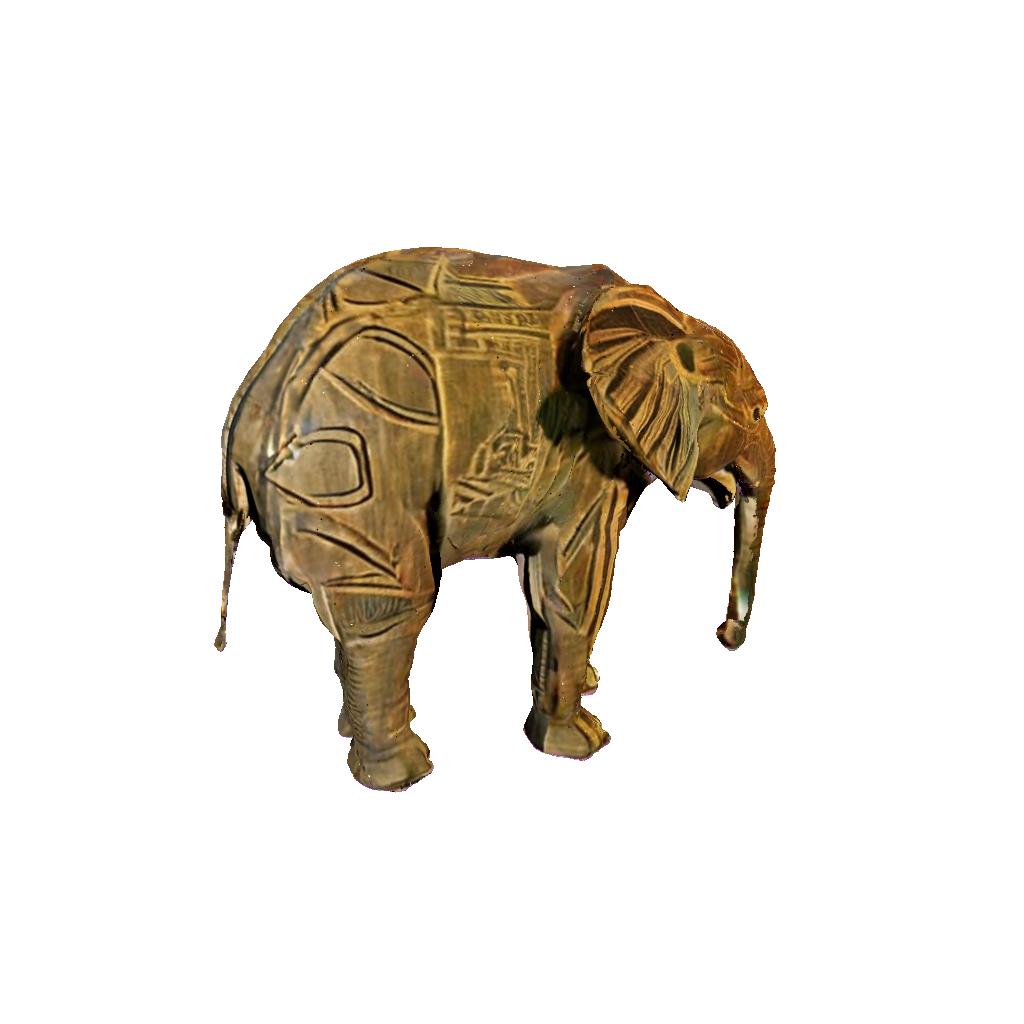} &
        \hspace{0.05cm}
        \includegraphics[height=0.12\linewidth,trim={12cm 8.5cm 11cm 6cm},clip]{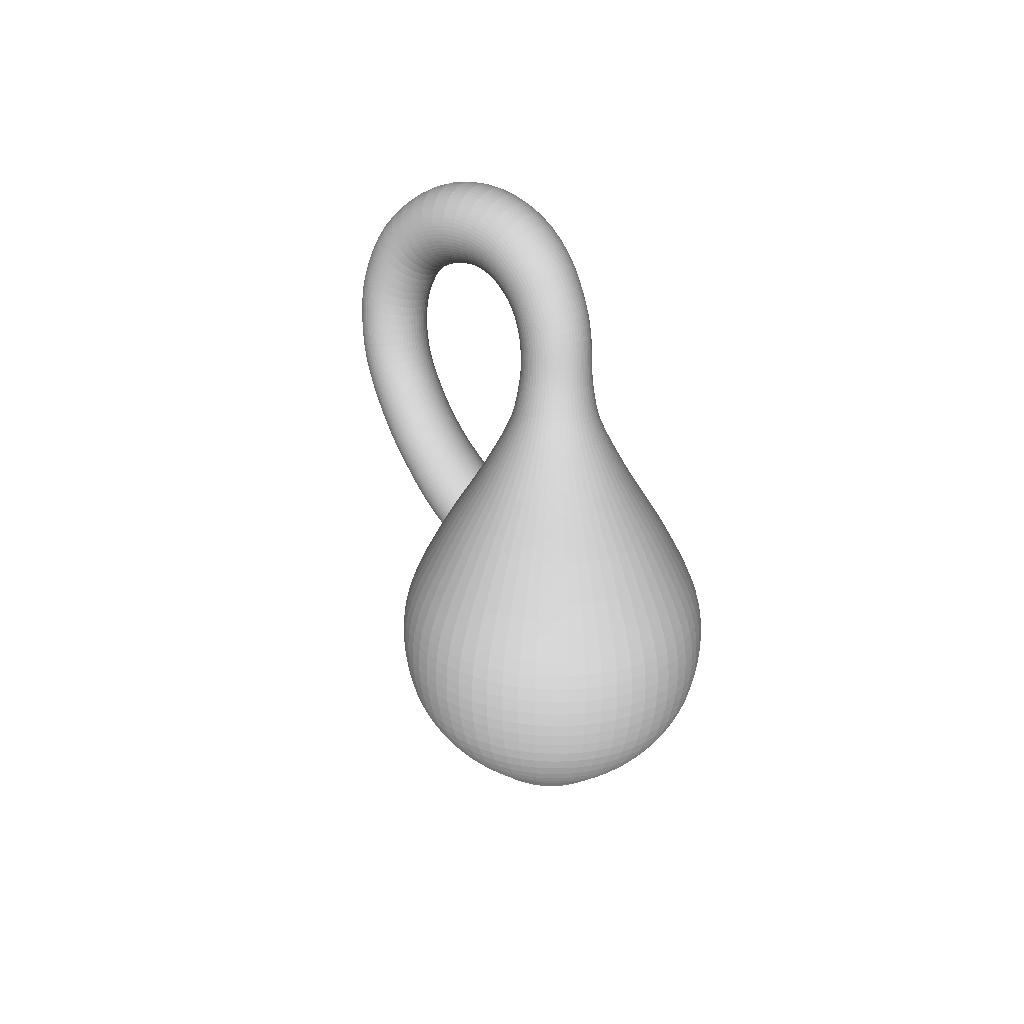} &
        \includegraphics[height=0.12\linewidth,trim={12cm 8.5cm 11cm 6cm},clip]{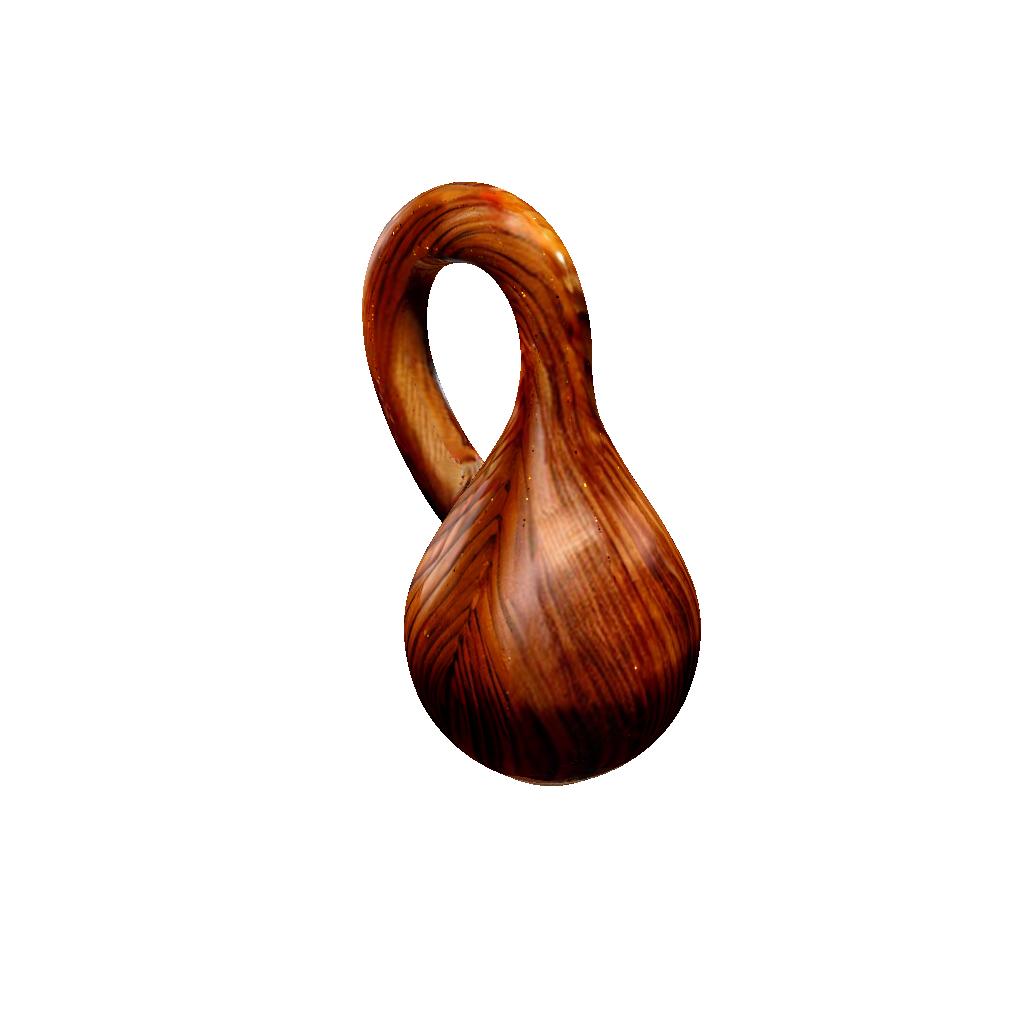}
        \includegraphics[height=0.12\linewidth,trim={12cm 10cm 12cm 6cm},clip]{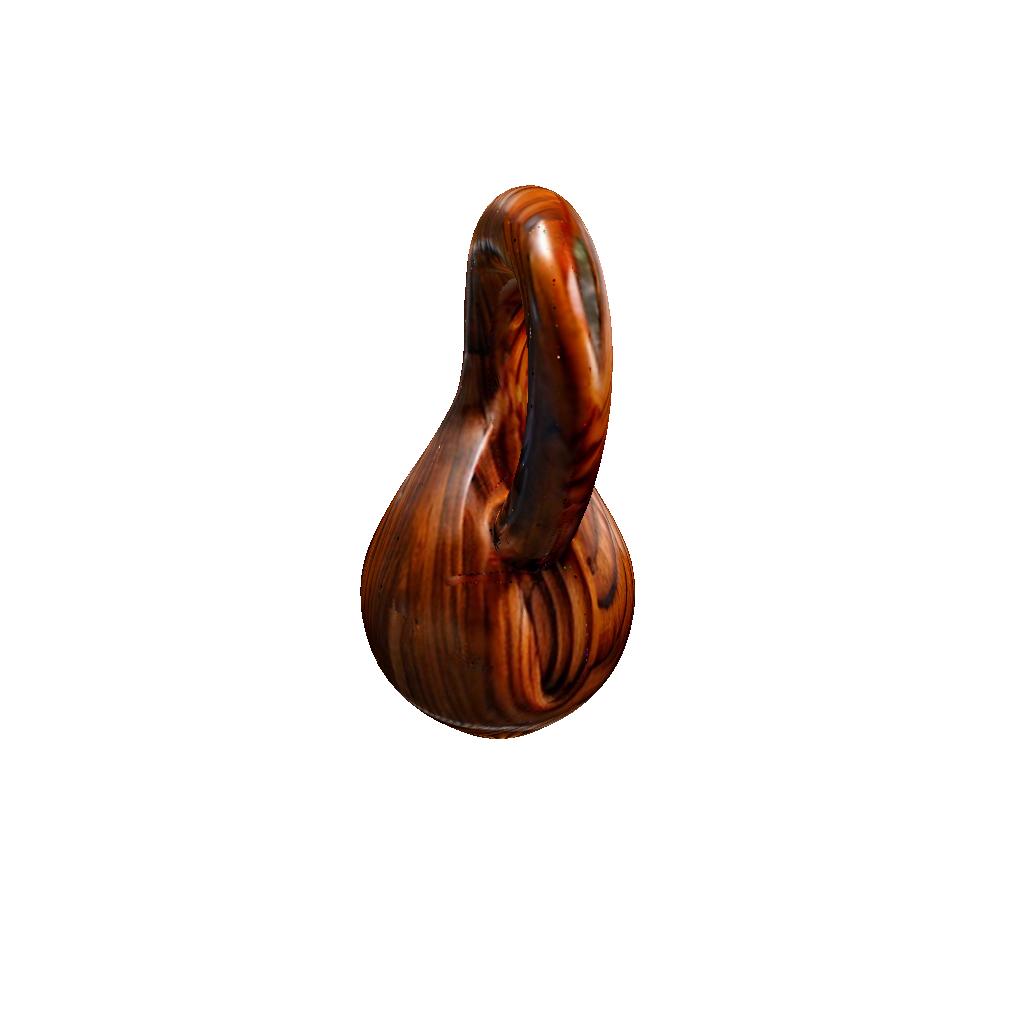}
        \\ 
         & \multicolumn{1}{c}{``A turtle''}  & & \multicolumn{1}{l}{``A hand carved wood elephant''} &  & \multicolumn{1}{l}{``A wooden klein bottle''}
    \end{tabular}
    }
    \vspace{-0.4cm}
    \caption{
        Texturing results. Our method generates a high-quality texture for a collection of prompts and geometries.
        \vspace{0.1cm}
    }
    \label{fig:paint_results}
\end{figure*}

%% file: sections/results.tex
\section{Experiments}

We now turn to validate the robustness and effectiveness on our proposed method through a set of experiments.

\subsection{Text-Guided Texturing}

\paragraph{\textbf{Qualitative Results.}}
We first demonstrate results achieved with TEXTure across several geometries and driving text prompts. \Cref{fig:paint_results},~\Cref{fig:teaser} and~\Cref{fig:more_paint_results} show highly-detailed realistic textures that are conditioned on a single text prompt. Observe, for example, how the generated textures nicely align with the geometry of the turtle and elephant shapes.
Moreover, the generated textures are consistent both on a local scale (e.g., along the shell of the turtle) and a global scale (e.g., the shell is consistent across different views). 
Furthermore, TEXTure can successfully generate textures of an individual (e.g., of Albert Einstein in~\Cref{fig:teaser}). Observe how the generated texture captures fine details of Albert Einstein's face.
Finally, TEXTure can successfully handle challenging geometric shapes, including the non-orientable Klein bottle, shown in~\Cref{fig:paint_results}.

\input{figures/experiments/comparisons1/fig.tex}

\vspace{-0.1cm}
\paragraph{\textbf{Qualitative Comparisons.}}
In~\Cref{fig:comparisons1} we compare our method to several state-of-the-art methods for 3D texturing.
First, we observe that Clip-Mesh~\cite{khalid2022clipmesh} and Text2Mesh~\cite{michel2022text2mesh} struggle in achieving globally-consistent results due to their heavy reliance CLIP-based guidance. See the "90's boombox" example at the top of~\Cref{fig:comparisons1} where speakers are placed sporadically in competing methods.
For Latent-Paint~\cite{metzer2022latent}, the results are more plausible but still lack in terms of quality. For example, Latent-Paint often struggles in achieving visibly sharp textures such as those shown on ``'a desktop iMac'.
We attribute this shortcoming due to its reliance on score distillation~\cite{poole2022dreamfusion}, which tends to omit high-frequency details.

\input{figures/user_study/user_study.tex}

The last row in~\Cref{fig:comparisons1}, depicting a statue of Napoleon Bonaparte, showcases our method's ability to produce fine details compared to alternative methods that struggle to produce matching quality. 
Notably, our results were achieved significantly faster than the alternative methods. Generating a single texture with TEXTure takes approximately $5$ minutes compared to $19$ through $45$ minutes for alternative methods (See~\Cref{tb:user_study}). 
We provide additional qualitative results in~\ref{fig:more_paint_results} and in the supplementary materials.

\paragraph{\textbf{User Study. }} Finally, we conduct a user study to analyze the fidelity and overall quality of the generated textures. We select $10$ text prompts and corresponding 3D meshes and texture the meshes using TEXTure and two baselines: Text2Mesh~\cite{michel2022text2mesh} and Latent-Paint~\cite{metzer2022latent}. 
For each prompt and method, we ask each respondent to evaluate the result with respect to two aspects: (1) its overall quality and (2) the level at which it reflects the text prompt, on a scale of $1$ to $5$.
Results are presented in~\Cref{tb:user_study} where we show the average results across all prompts for each method. As can be seen, TEXTure outperforms both baselines in terms of both overall quality and text fidelity by a significant margin.
Importantly, our method's improved quality and fidelity are attained with a significant decrease in runtime. Specifically, TEXTure achieves a decrease of $6.4\times$ in running time compared to Text2Mesh and a decrease of $9.2\times$ relative to Latent-Paint. 

In addition to the above evaluation setting, we ask respondents to rank the methods relative to each other. Specifically, for each of the $10$ prompts, we show the results of all methods side-by-side (in a random order) and ask respondents to rank the results. In~\Cref{tb:user_study_rank} we present the average rank of each method, averaged across the $10$ prompts and across all responses.
Note that a lower rank is preferable in this setting. As can be seen, TEXTure has a significantly better average rank relative to the two baselines, as desired. 
This further demonstrates the effectiveness of TEXTure in generating high-quality, semantically-accurate textures.

\paragraph{\textbf{Ablation Study. }}
An ablation validating the different components of our TEXTure scheme is shown in~\Cref{fig:ablation}. One can see that each component is needed for achieving high-quality generations and improving our method's robustness to the sensitivity of the generation process. 
Specifically, without differentiating between \keep and \generate regions the generated textures are unsatisfactory. For both the teddy bear and the sports car one can clearly see that the texture presented in \texttt{A} fails to achieve local and global consistency, with inconsistent patches visible across the texture. In contrast, thanks to our blending technique for \keep regions, \texttt{B} achieves local consistency between views. Still, observe that the texture of the teddy bear legs in the top example, do not match the texture of its back.
By incorporating our improved \generate method, we achieve more consistent results across the entire shape, as shown in \texttt{C}.
These results tend to have smeared regions as can be observed in the fur of the teddy bear, or the text written across the hood of the car. We attribute this to the fact that some regions are painted  from oblique angles at early viewpoints, which are not ideal for texturing, and are not refined.
By identifying \refine regions and applying our full TEXTure scheme, we are able to effectively address these problems and produce sharper textures, see \texttt{D}.

\input{figures/ablation/fig.tex}
\input{figures/user_study/user_study_average_rank.tex}
\subsection{Texture Capturing}
We next validate our proposed texture capture and transfer technique that can be applied over both 3D meshes and images.

\paragraph{\textbf{Texture From Mesh}}
\input{figures/mesh2mesh/fig.tex}
As mentioned in~\Cref{sec:texture_transfer} we are able to capture a texture of a given mesh by combining concept learning techniques~\cite{gal2022textual_inversion,ruiz2022dreambooth} with view-specific directional tokens. We can then use the learned token with TEXTure to color new meshes accordingly.
\Cref{fig:mesh2mesh} presents texture transfer results from two existing meshes onto various target geometries.
While the training of the new texture token is performed using prompts of the form ``A  $\langle D_v \rangle$ photo of $\langle \mathcal{S}_{texture} \rangle$'', we can generate new textures by forming different texture prompts when transferring the learned texture.
Specifically, the "Exact" results in~\Cref{fig:mesh2mesh} were generated using the specific text prompt used for fine-tuning, and the rest of the outputs were generated "in the style of $\langle \mathcal{S}_{texture} \rangle$".
Observe how a single eye is placed on Einstein's face when the exact prompt is used, while two eyes are generated when only using a prompt "a $\langle D_v \rangle$ photo of Einstein that looks like $\langle \mathcal{S}_{texture} \rangle$".
A key component of the texture-capturing technique is our novel spectral augmentations scheme. We refer the reader to the supplementary materials for an ablation study over this component.

\vspace{-0.1cm}
\paragraph{\textbf{Texture From Image}}
In practice, it is often more practical to learn a texture for a set of images rather than from a 3D mesh. As such, in~\Cref{fig:images2mesh}, we demonstrate the results of our transferring scheme where the texture is captured from a collection of images depicting the source object. One can see that even when given only several images, our method can successfully texture different shapes with a semantically similar texture. We find these results exciting, as it means one can easily paint different 3D shapes using textures derived from real-world data, even when the texture itself is only implicitly represented in the fine-tuned model and concept tokens.

\subsection{Editing}
Finally, in~\Cref{fig:edits} we show the editing results of existing textures. 
The top row of~\Cref{fig:edits} shows scribble-based results where a user manually edits the texture atlas image. Then we \refine the texture atlas to seamlessly blend the edited region into the final texture. Observe how the manually annotated  white spot on the bunny on the right turns into a realistic-looking patch of white fur.

The bottom of~\Cref{fig:edits} shows text-based editing results, where an existing texture is \refined according to a new  text prompt.
The bottom right example illustrates a texture generated on a similar geometry with the same prompt \textbf{from scratch}.
Observe that the texture generated from scratch significantly differs from the original texture. In contrast, when applying editing, the  textures are able to remain semantically close to the input texture while generating novel details to match the target text.

\input{figures/images2mesh/fig.tex}

%% file: figures/experiments/comparisons1/fig.tex
\begin{figure}[b]
    \centering
    \vspace{-0.5cm}
    \setlength{\tabcolsep}{6pt}
    {\small
    \begin{tabular}{c c c c}
        \includegraphics[width=0.2\linewidth]{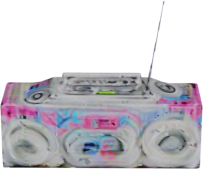} & 
        \includegraphics[width=0.2\linewidth]{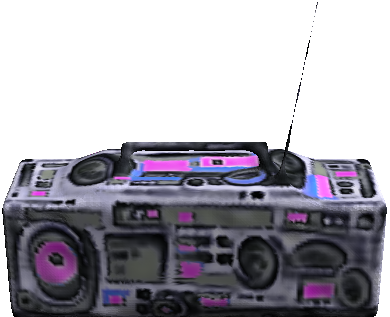} & 
        \includegraphics[width=0.2\linewidth]{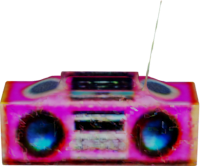} & 
        \includegraphics[width=0.2\linewidth]{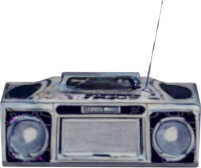} \\
        \multicolumn{4}{c}{``A 90's boombox''} \\

        \includegraphics[width=0.2\linewidth]{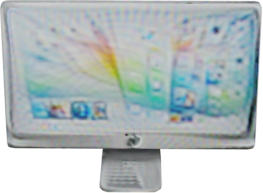} & 
        \includegraphics[width=0.2\linewidth]{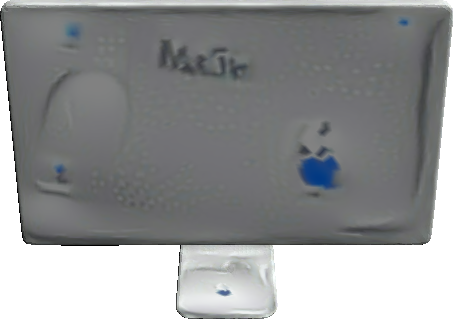} & 
        \includegraphics[width=0.2\linewidth]{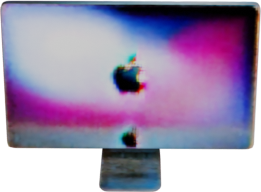} & 
        \includegraphics[width=0.2\linewidth]{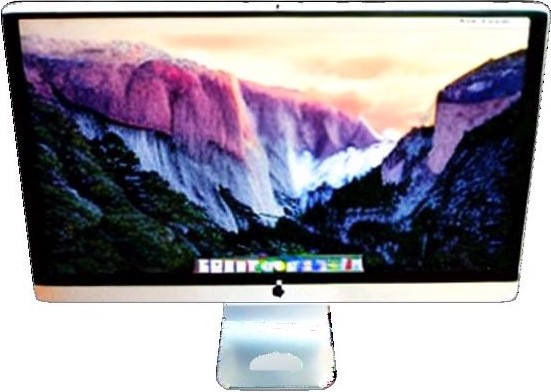} \\
        \multicolumn{4}{c}{``A desktop iMac''} \\ 
        \includegraphics[width=0.2\linewidth]{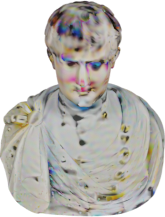} & 
        \includegraphics[width=0.2\linewidth]{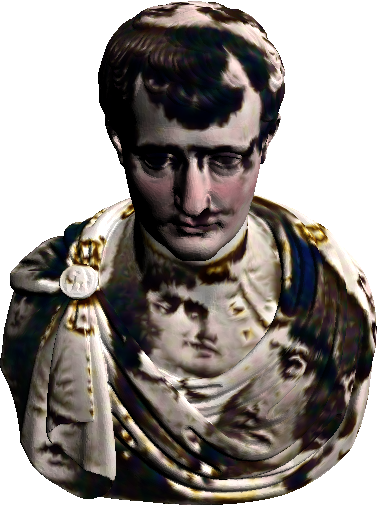} & 
        \includegraphics[width=0.2\linewidth]{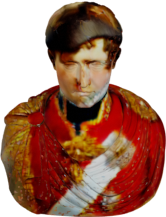} & 
        \includegraphics[width=0.2\linewidth]{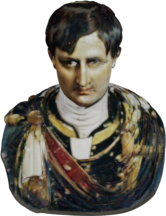} \\
        \multicolumn{4}{c}{``A photo of Napoleon Bonaparte''} \\[0.1pt]
        \vspace{-0.3cm}
         Clip-Mesh & Text2Mesh & Latent-Paint & Ours

    \end{tabular}}
    \caption{Visual comparison of text-guided texture generation. For each input prompt, we show results for a single viewpoint. Best viewed zoomed in. Meshes obtained from ModelNet40~\cite{wu20153d}.}
    \label{fig:comparisons1}
\end{figure} 

%% file: figures/user_study/user_study.tex
\begin{table}
\small
\centering
\setlength{\tabcolsep}{3pt}
\begin{tabular}{l c c c} 
\toprule
Method & \begin{tabular}{c} Overall  \end{tabular} & \begin{tabular}{c} Text  \end{tabular} & \begin{tabular}{c} Runtime\end{tabular} \\
 & \begin{tabular}{c}  Quality ($\uparrow$) \end{tabular} & \begin{tabular}{c}  Fidelity ($\uparrow$) \end{tabular} & \begin{tabular}{c}  (minutes) ($\downarrow$)\end{tabular} \\
\midrule
Text2Mesh        & 2.57          & 3.62          & 32 ($6.4\times$) \\
Latent-Paint     & 2.95          & 4.01          & 46 ($9.2\times$) \\
\textbf{TEXTure} & \textbf{3.93} & \textbf{4.44} & \textbf{5} \\
\bottomrule
\end{tabular}
\caption{User study results conducted with $30$ respondents. 
We ask respondents to rate the results on a scale of $1$ to $5$ with respect to the overall quality of the results and the level at which the result reflects the text prompt. 
Results are averaged across all responses and text prompts. 
}
\vspace{-0.3cm}
\label{tb:user_study}
\end{table}

%% file: figures/ablation/fig.tex
\begin{figure}[t]
    \centering
    \setlength{\belowcaptionskip}{-4pt}
    \setlength{\tabcolsep}{4pt}
    \renewcommand{\arraystretch}{1}
    \newcommand{\pl}{0.2}
    
    {\small
    \begin{tabular}{c c c c}
        \includegraphics[width=\pl\linewidth,trim={9cm 10cm 10cm 7cm},clip]{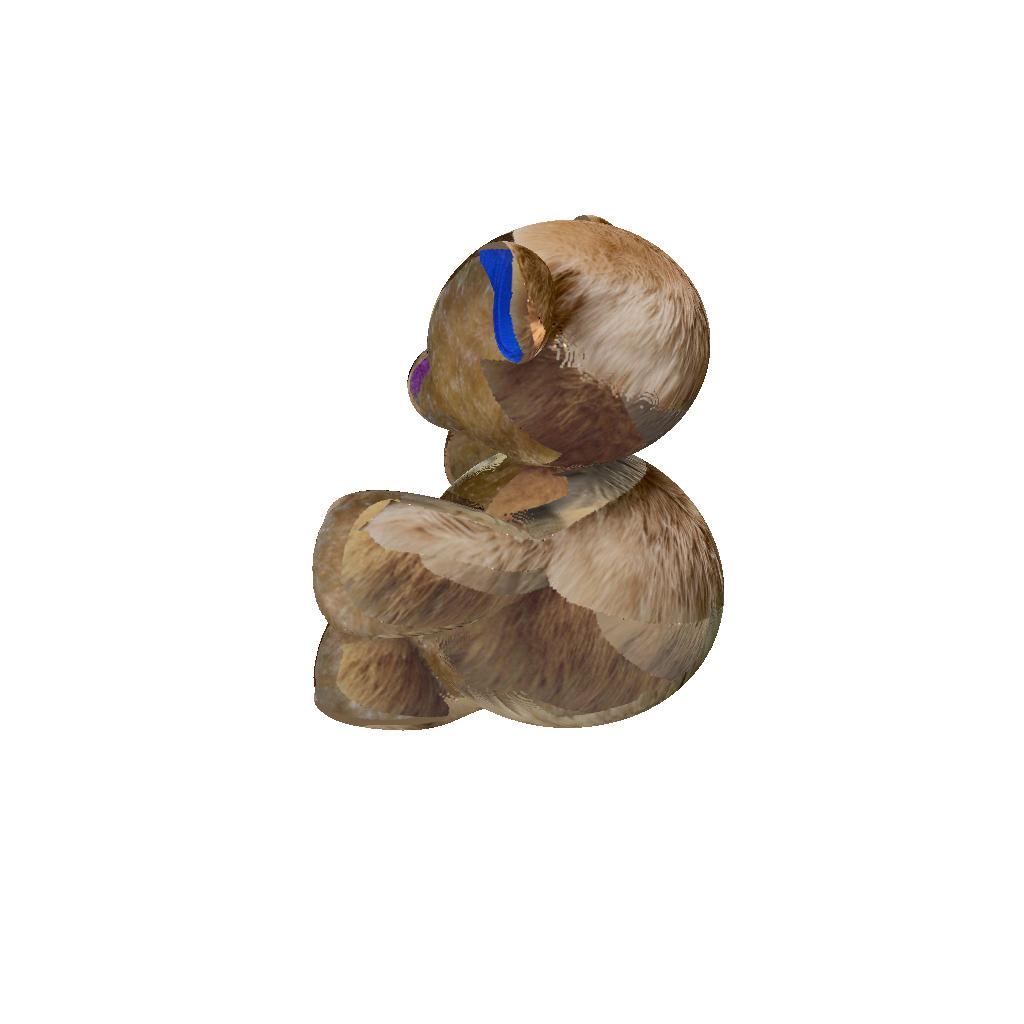} &
        \includegraphics[width=\pl\linewidth,trim={9cm 10cm 10cm 7cm},clip]{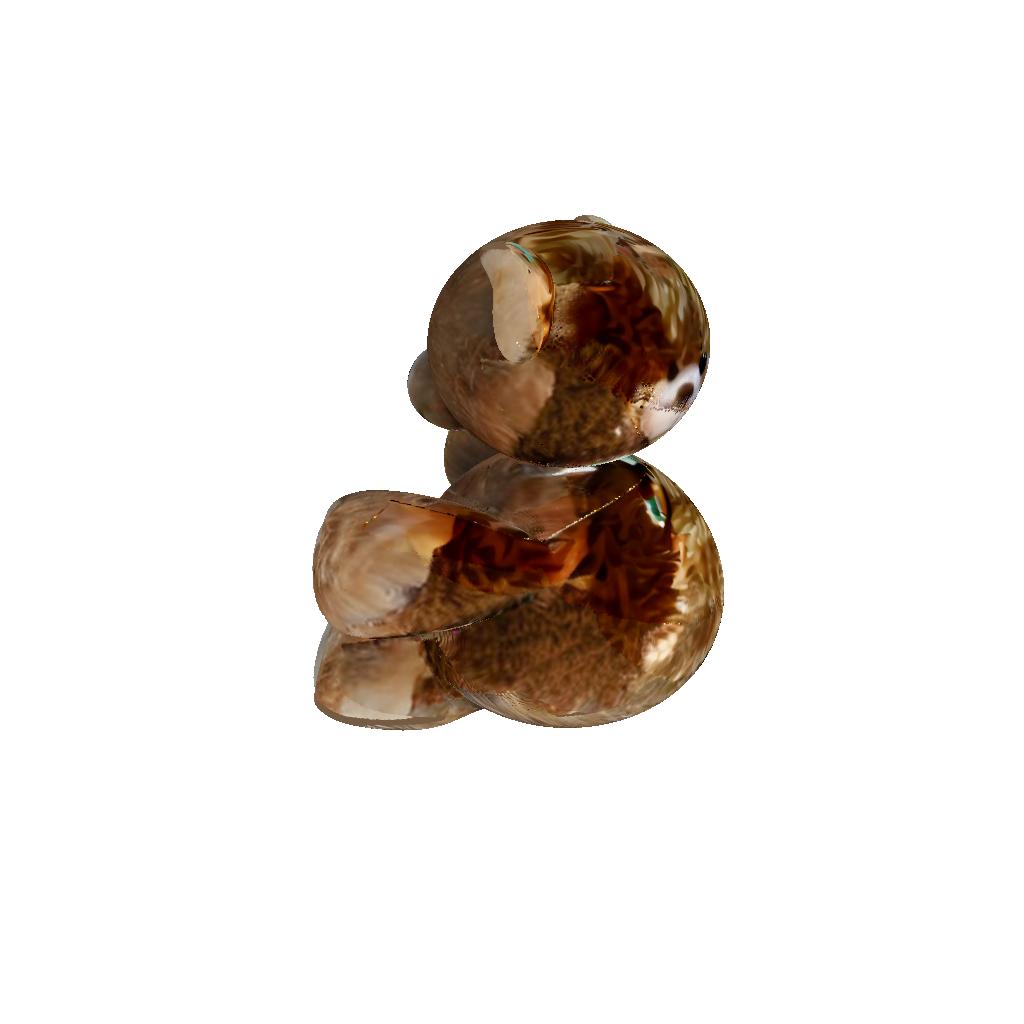} &
        \includegraphics[width=\pl\linewidth,trim={9cm 10cm 10cm 7cm},clip]{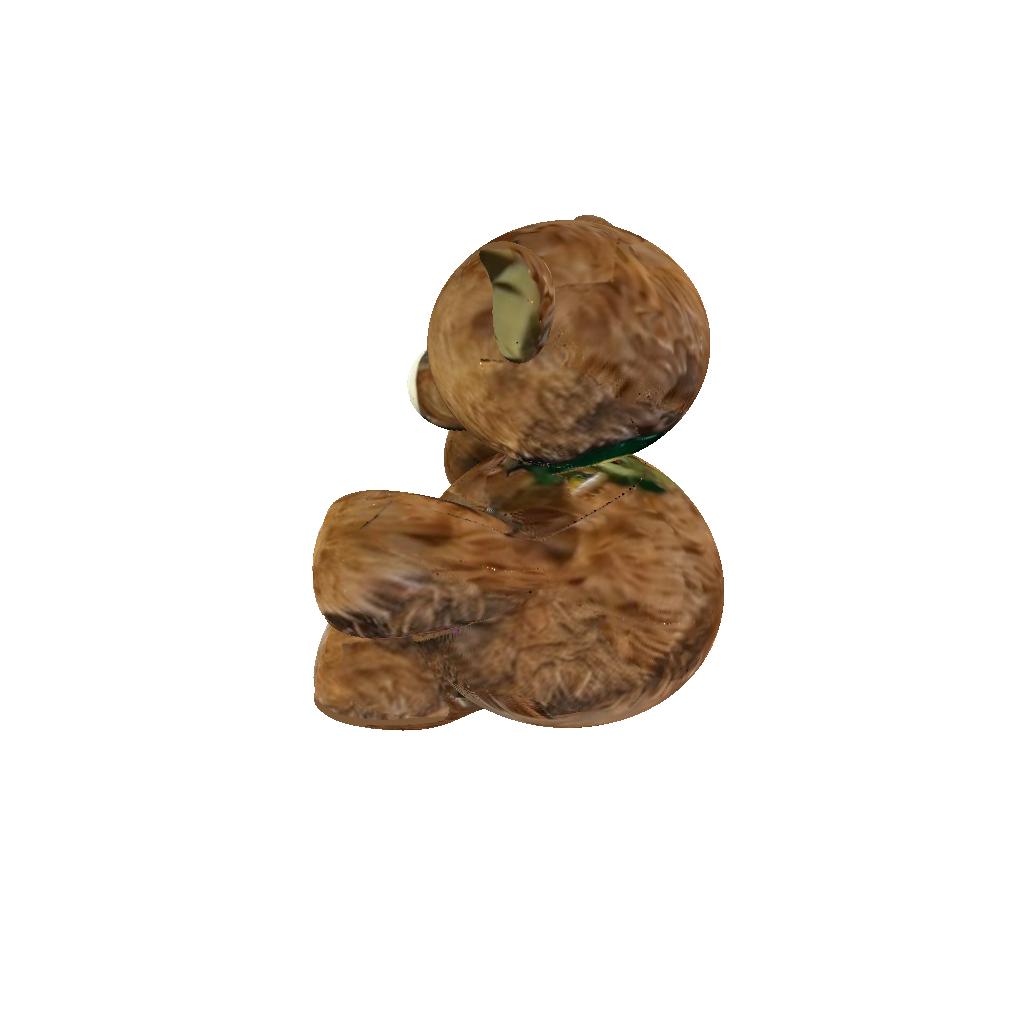} &
        \includegraphics[width=\pl\linewidth,trim={9cm 10cm 10cm 7cm},clip]{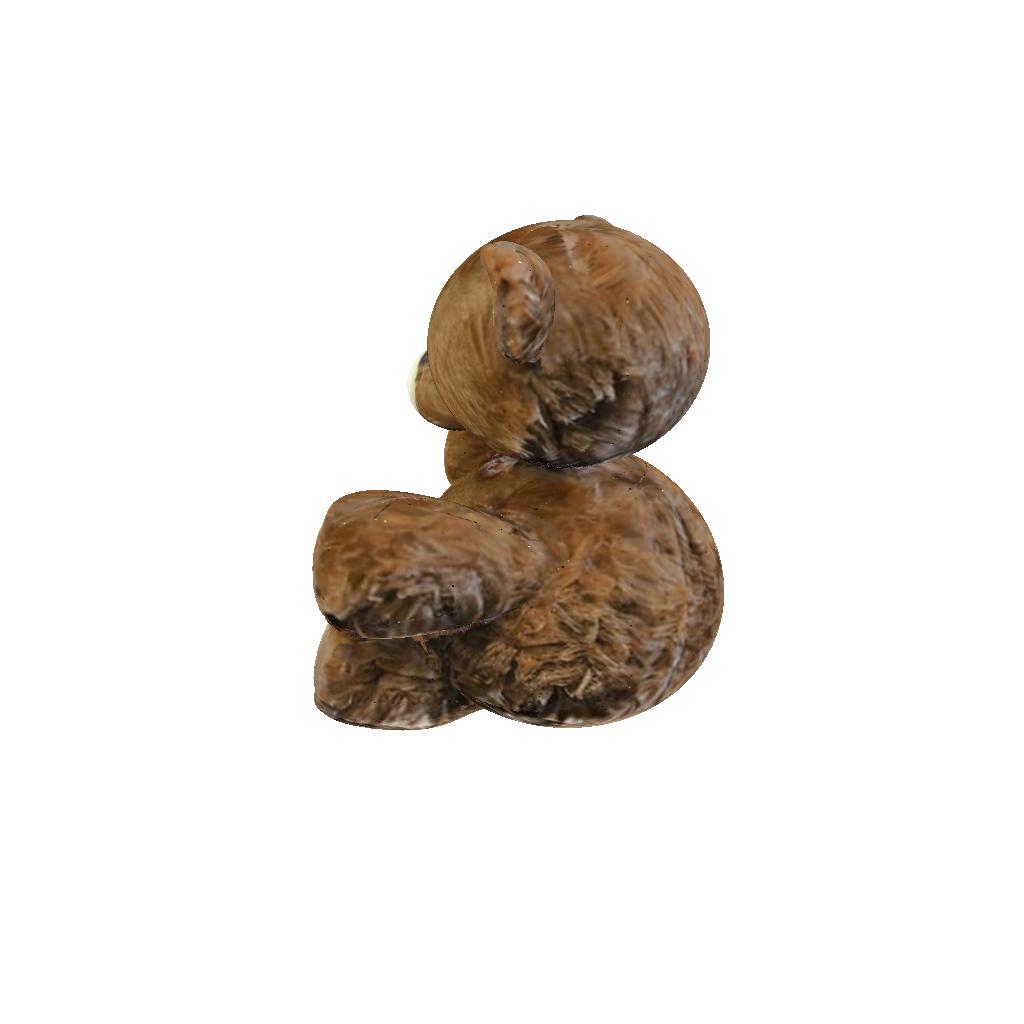} 
        \\
        \includegraphics[width=\pl\linewidth,trim={18cm 14cm 12cm 18cm},clip]{figures/ablation/teddy_naive.jpg} &
        \includegraphics[width=\pl\linewidth,trim={18cm 14cm 12cm 18cm},clip]{figures/ablation/teddy_blended.jpg} &
        \includegraphics[width=\pl\linewidth,trim={18cm 14cm 12cm 18cm},clip]{figures/ablation/teddy_paint.jpg} &
        \includegraphics[width=\pl\linewidth,trim={18cm 14cm 12cm 18cm},clip]{figures/ablation/teddy_full.jpg}         \\
        \includegraphics[width=\pl\linewidth,trim={18cm 13.5cm 12cm 18.5cm},clip]{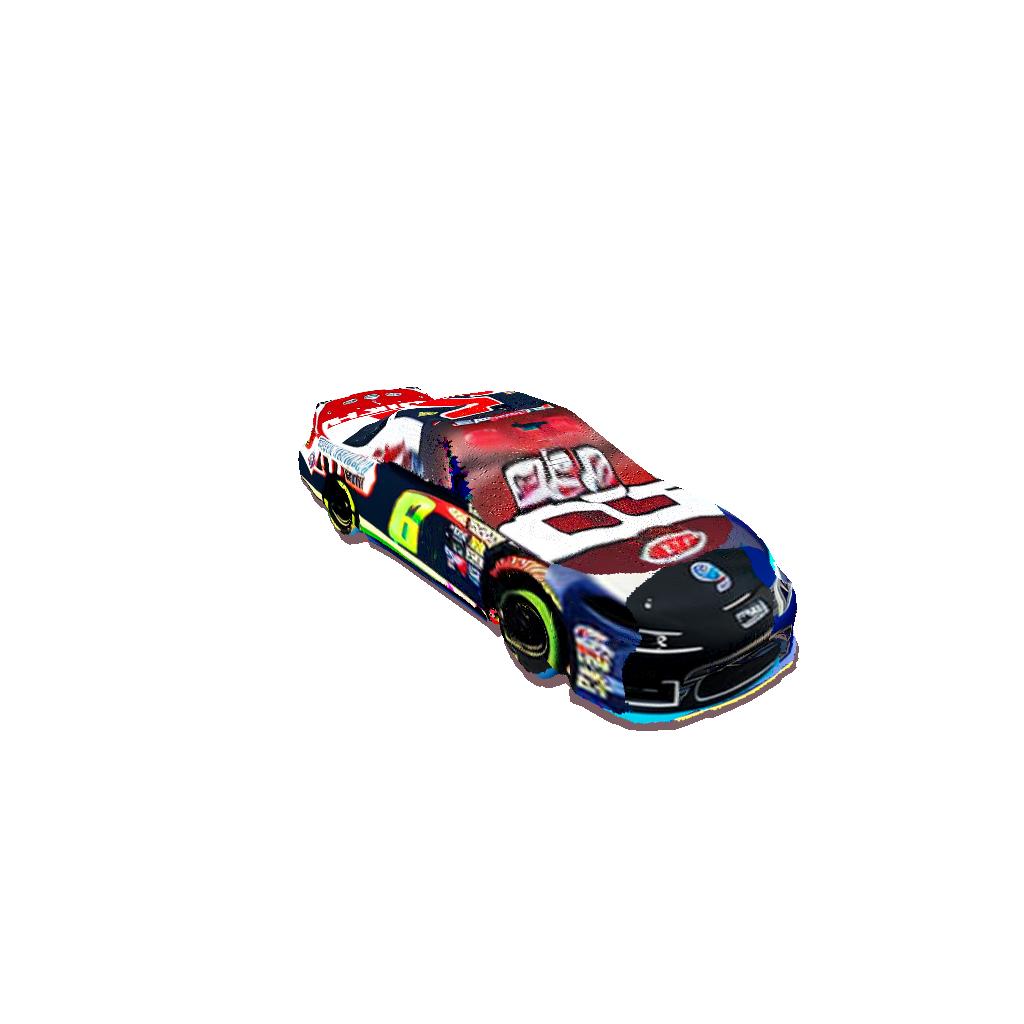} &
        \includegraphics[width=\pl\linewidth,trim={18cm 13.5cm 12cm 18.5cm},clip]{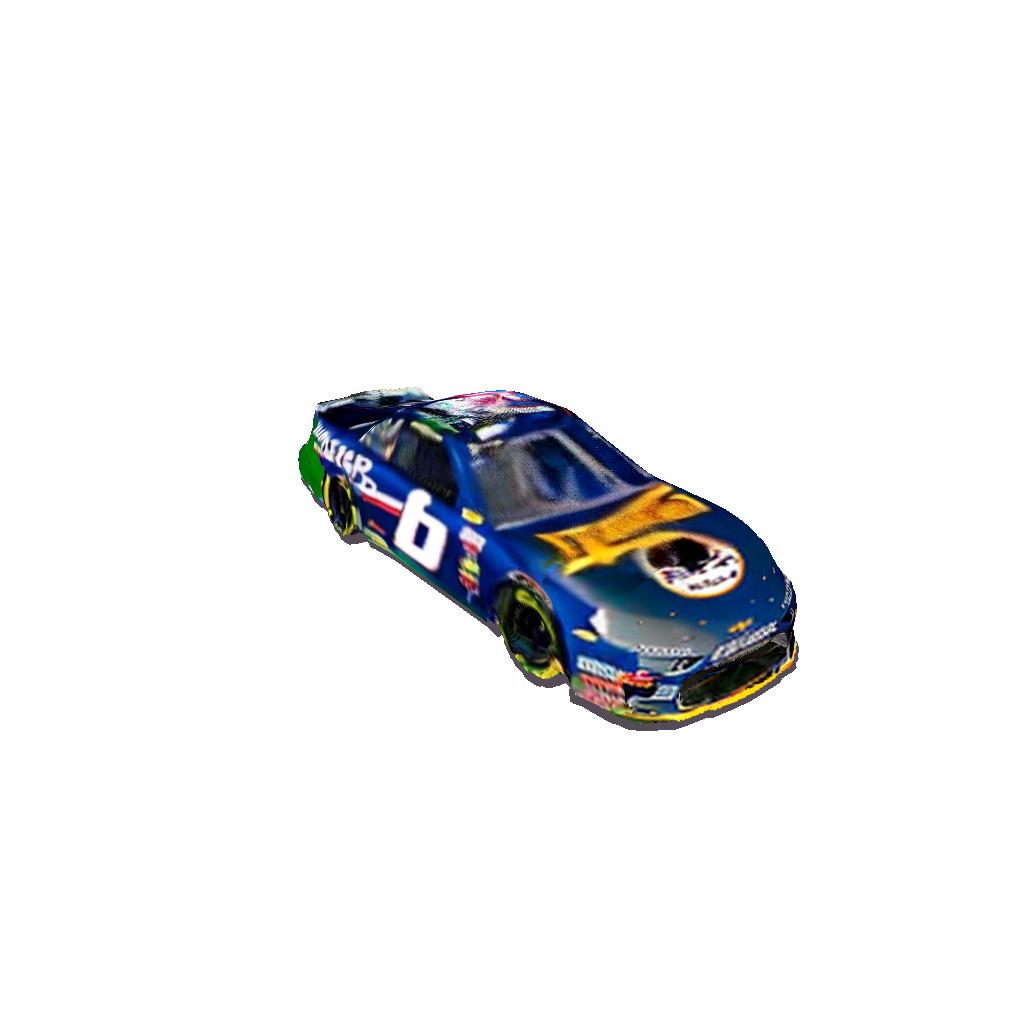} &
        \includegraphics[width=\pl\linewidth,trim={18cm 13.5cm 12cm 18.5cm},clip]{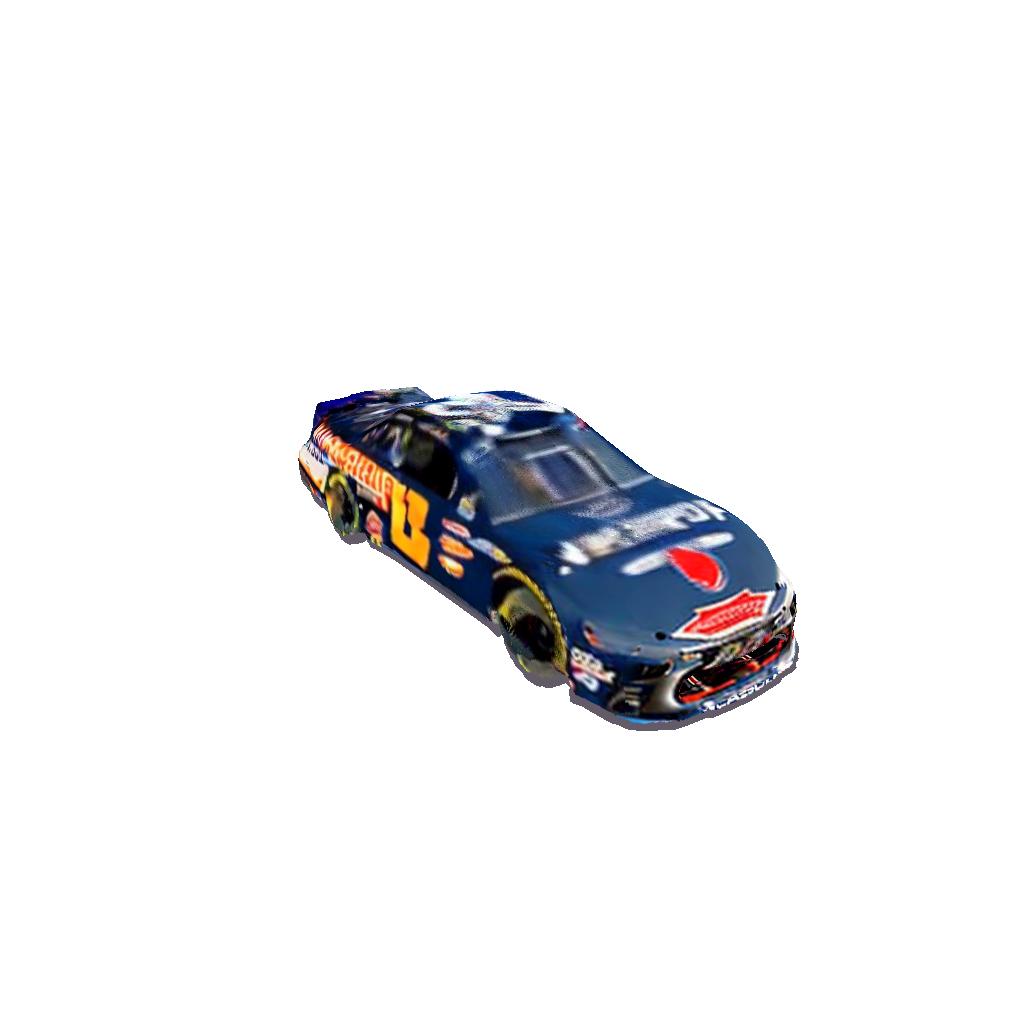} &
        \includegraphics[width=\pl\linewidth,trim={18cm 13.5cm 12cm 18.5cm},clip]{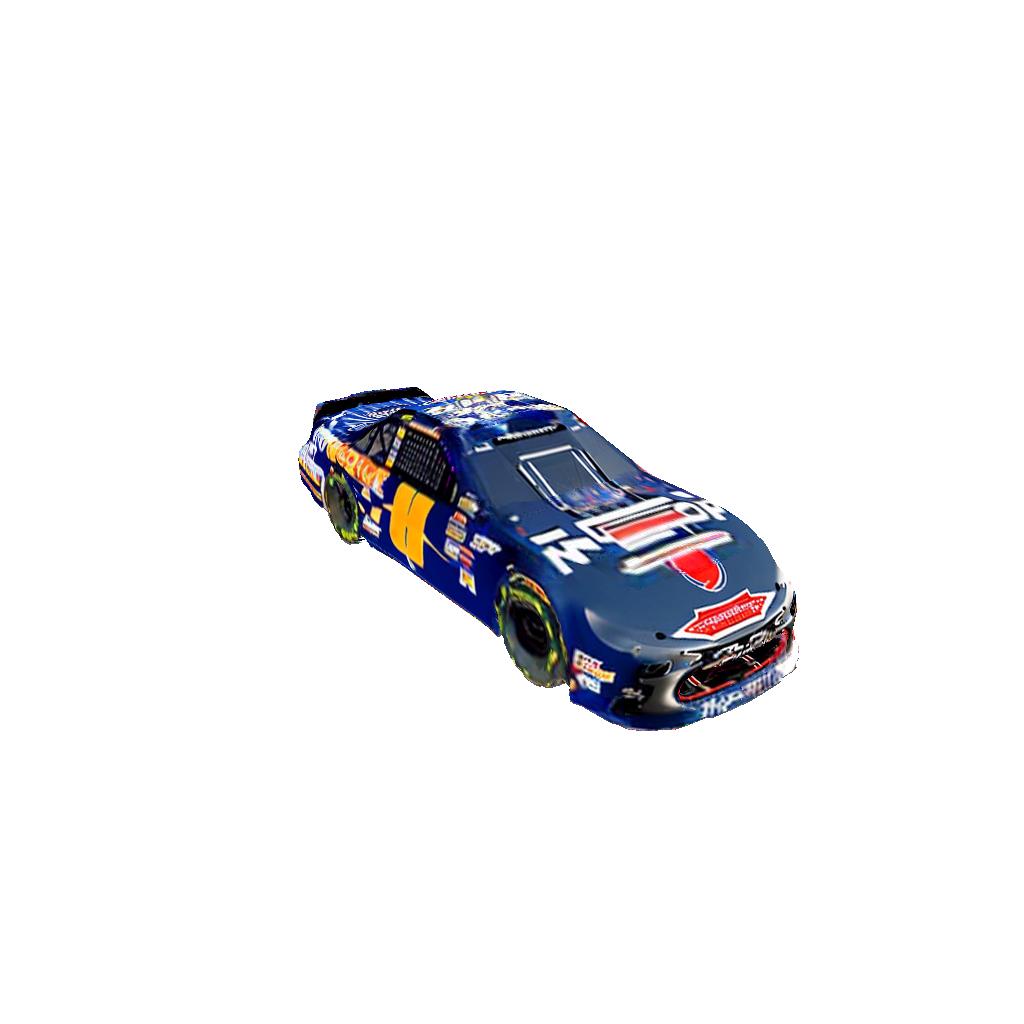}  \\
        \includegraphics[width=\pl\linewidth,trim={10cm 10cm 7.5cm 13cm},clip]{figures/ablation/car_2_naive.jpg} &
        \includegraphics[width=\pl\linewidth,trim={10cm 10cm 7.5cm 13cm},clip]{figures/ablation/car_2_blended.jpg} &
        \includegraphics[width=\pl\linewidth,trim={10cm 10cm 7.5cm 13cm},clip]{figures/ablation/car_2_inpaint.jpg} &
        \includegraphics[width=\pl\linewidth,trim={10cm 10cm 7.5cm 13cm},clip]{figures/ablation/car_2_full.jpg} 
        \\
        A & B & C & D
    \end{tabular}
    }
    \vspace{-0.3cm}
    \caption{Ablation of our different stages: (A) is a \naive painting scheme that paints the entire viewpoint. (B) takes into account \keep region. (C) is our inpainting-based scheme for \generate regions, and (D) is our complete scheme with \refine regions.
    Car model obtained from~\cite{wu20153d}, Teddy Bear model obtained from~\cite{teddy_bear_mesh}.}
    \label{fig:ablation}
\end{figure}

%% file: figures/user_study/user_study_average_rank.tex
\begin{table}
\small
\centering
\setlength{\tabcolsep}{2pt}
\begin{tabular}{l c c c} 
\toprule
& \begin{tabular}{c} Text2Mesh \end{tabular} & \begin{tabular}{c} Latent-Paint \end{tabular} & \begin{tabular}{c} \textbf{TEXTure} \end{tabular} \\
\midrule
Average Rank ($\downarrow$) & 2.44 & 2.24 & \textbf{1.32} \\
\bottomrule
\end{tabular}
\caption{
Additional user study results. Each respondent is asked to rank the results of the different methods with respect to overall quality.
} 
\vspace{-0.2cm}
\label{tb:user_study_rank}
\end{table}

%% file: figures/mesh2mesh/fig.tex
\begin{figure}
    \centering
    \setlength{\tabcolsep}{0pt}
    {\small
    \begin{tabular}{c c c}
        \includegraphics[height=0.22\linewidth,trim={8cm 9cm 11cm 9cm},clip]
        {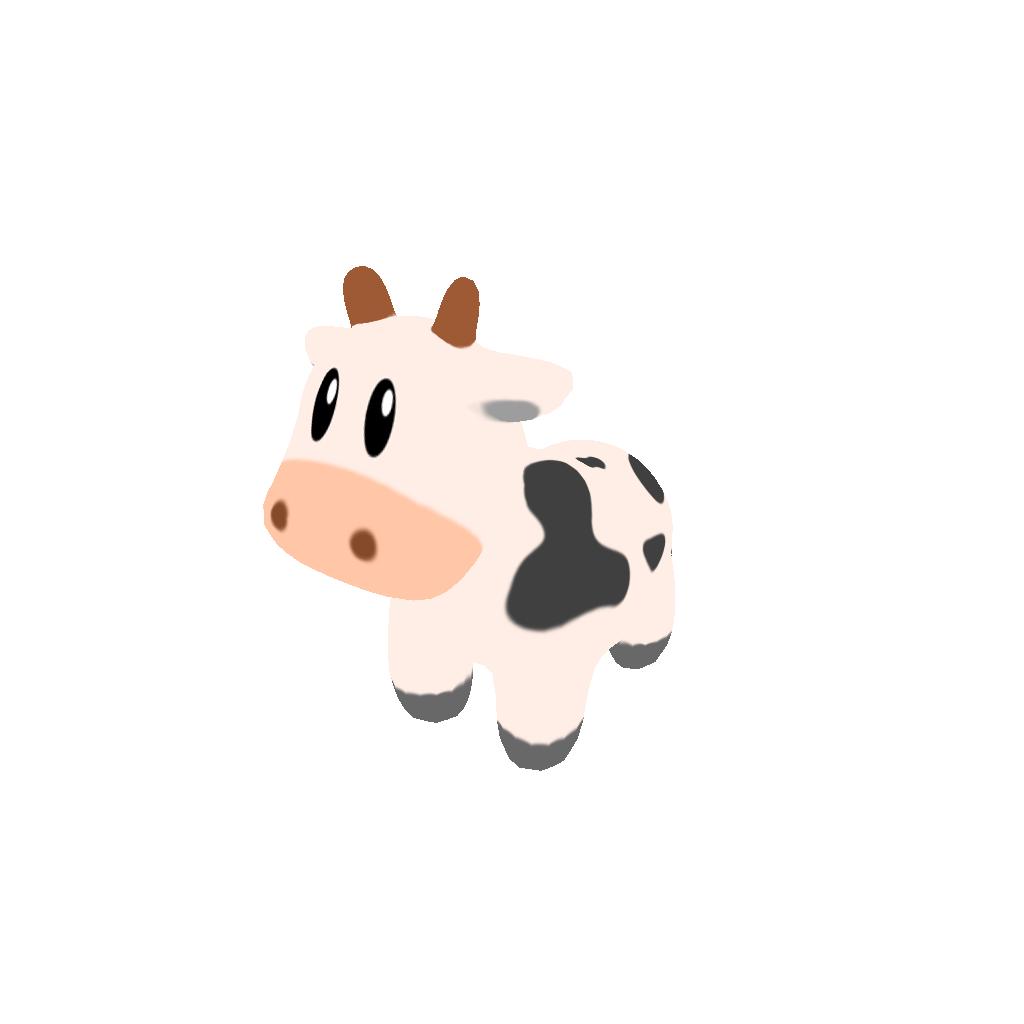} &
        \includegraphics[height=0.22\linewidth,trim={8cm 10cm 11cm 11cm},clip]
        {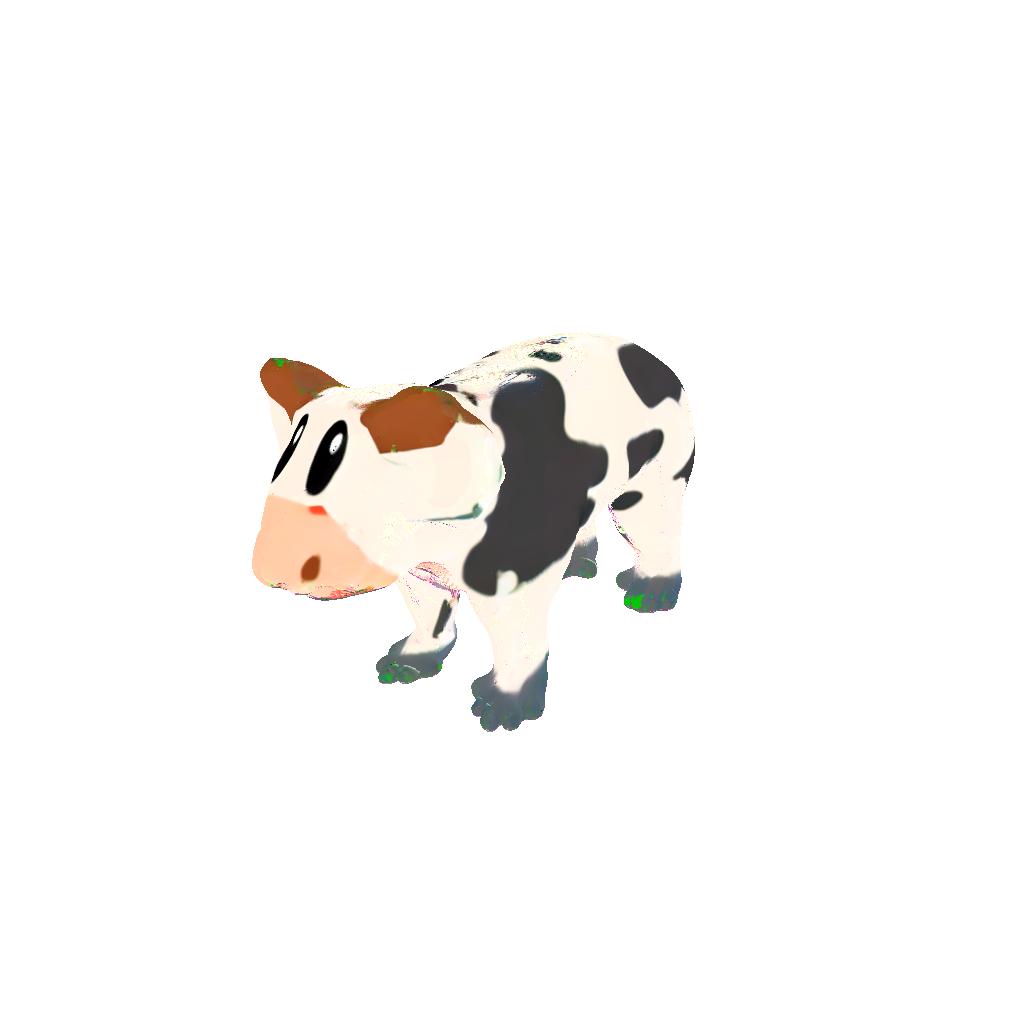} &
        \includegraphics[height=0.22\linewidth,trim={10cm 9cm 6cm 10cm},clip]{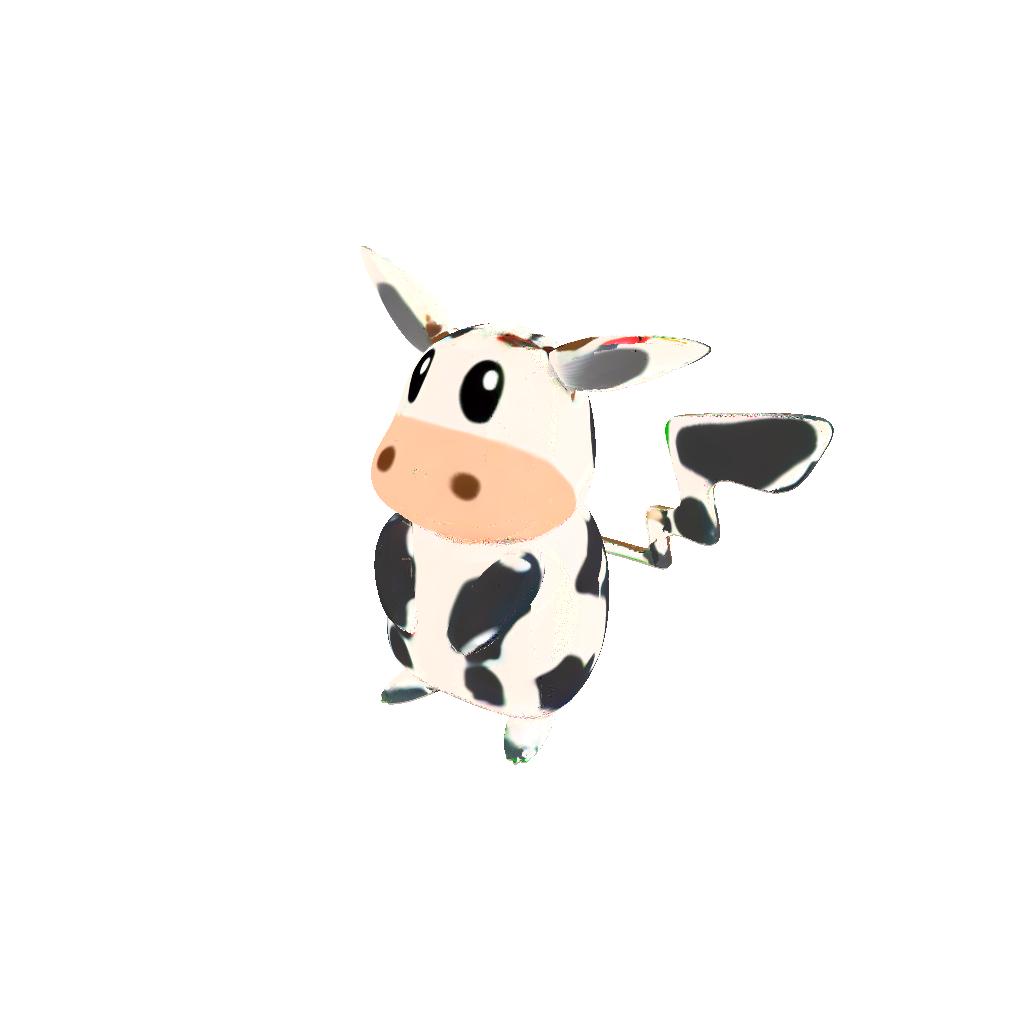} 
        \includegraphics[height=0.22\linewidth,trim={10cm 13cm 10cm 10cm},clip]
        {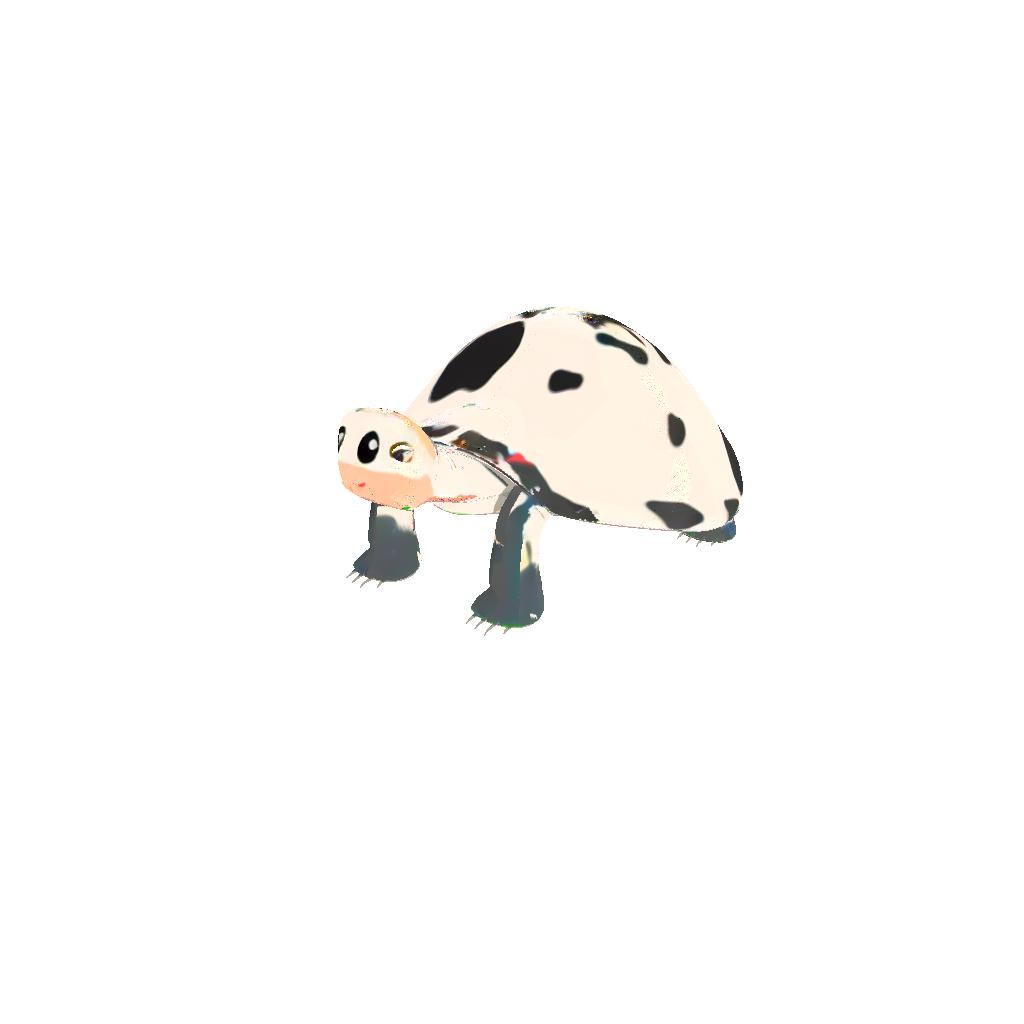} 
         \\
         \includegraphics[height=0.22\linewidth]
        {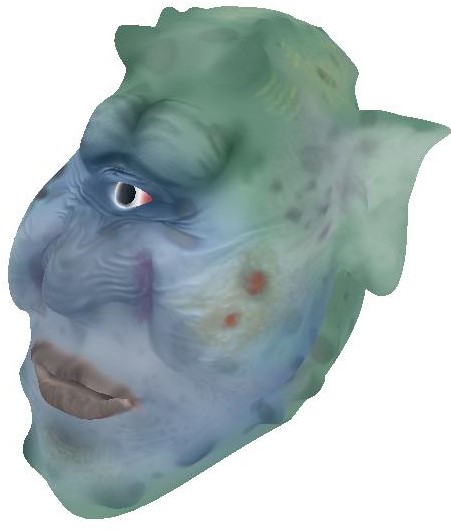} &
        \includegraphics[height=0.22\linewidth]
        {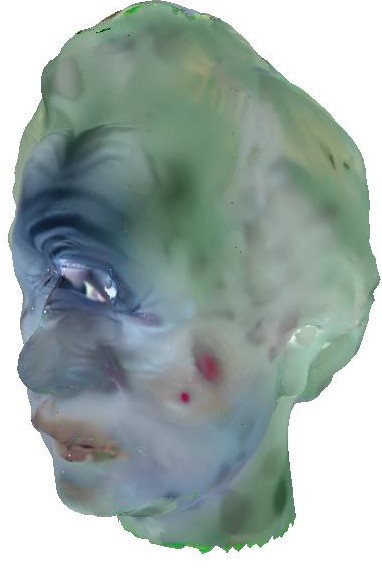} &
        \includegraphics[height=0.22\linewidth]
        {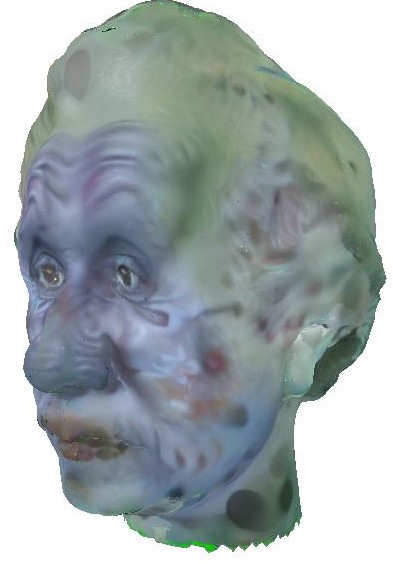} 
        \includegraphics[height=0.22\linewidth]{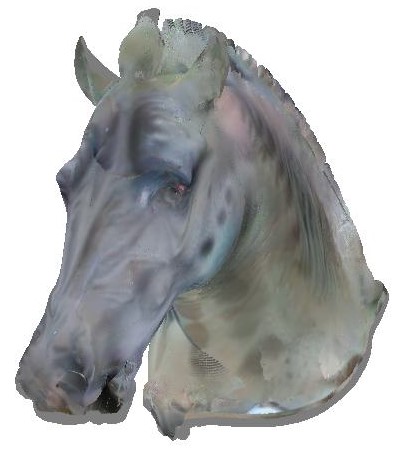} 
        \includegraphics[height=0.22\linewidth]
        {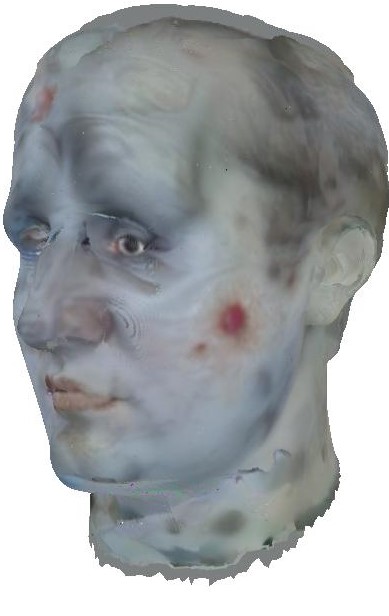} 
         \\
          Textured Input &  Exact & "...in the style of $\langle \mathcal{S}_{texture} \rangle$"
    \end{tabular}
    }
    \vspace{-0.3cm}
    \caption{Token-based Texture Transfer. The learned token $\langle \mathcal{S}_{texture} \rangle$ is applied to different geometries. 
    Einstein's head is shown with both exact and ``style'' prompts, demonstrating the effect on the transfer's semantics. Both Spot and Ogre are taken from Keenan's Repository~\shortcite{keenan3D}.
    }
    \label{fig:mesh2mesh}
    \vspace{-0.4cm}
\end{figure}

%% file: figures/images2mesh/fig.tex
\begin{figure}[t]
    \centering
    \setlength{\tabcolsep}{0pt}
    {\small
    \begin{tabular}{c c}
        \includegraphics[height=0.22\linewidth]
        {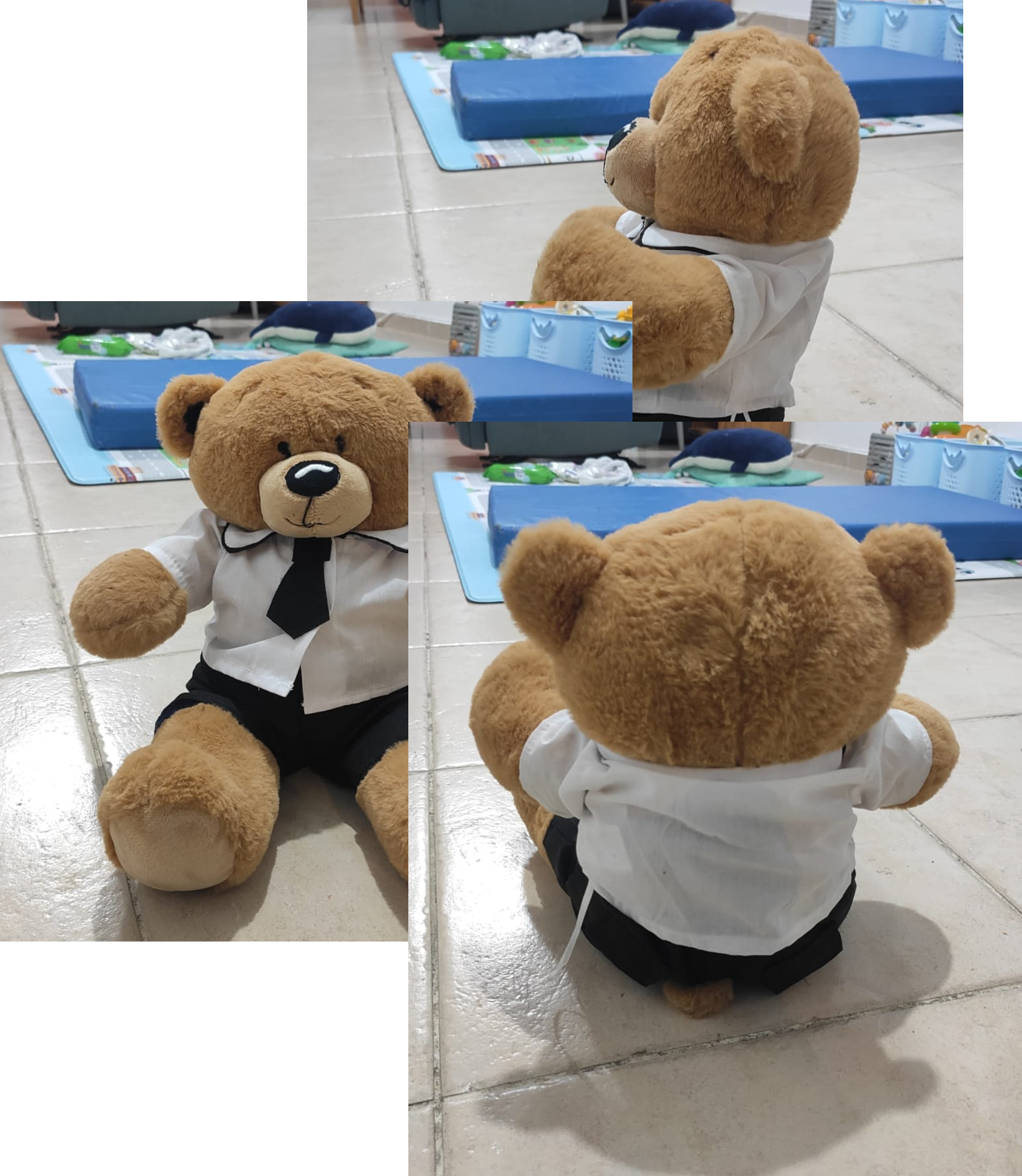} &
        \includegraphics[height=0.22\linewidth,trim={9cm 8cm 10cm 6cm},clip]
        {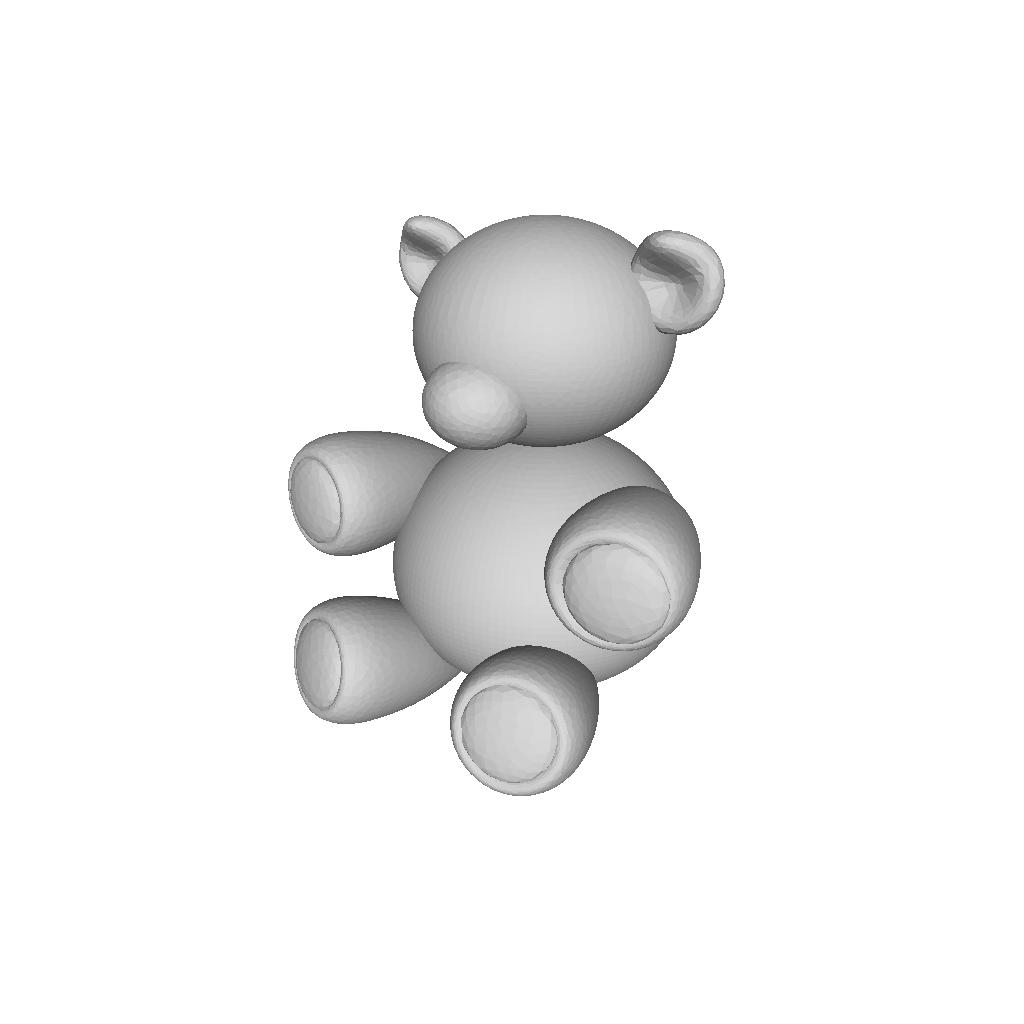} 
        \includegraphics[height=0.22\linewidth,trim={8cm 8cm 10cm 6cm},clip]{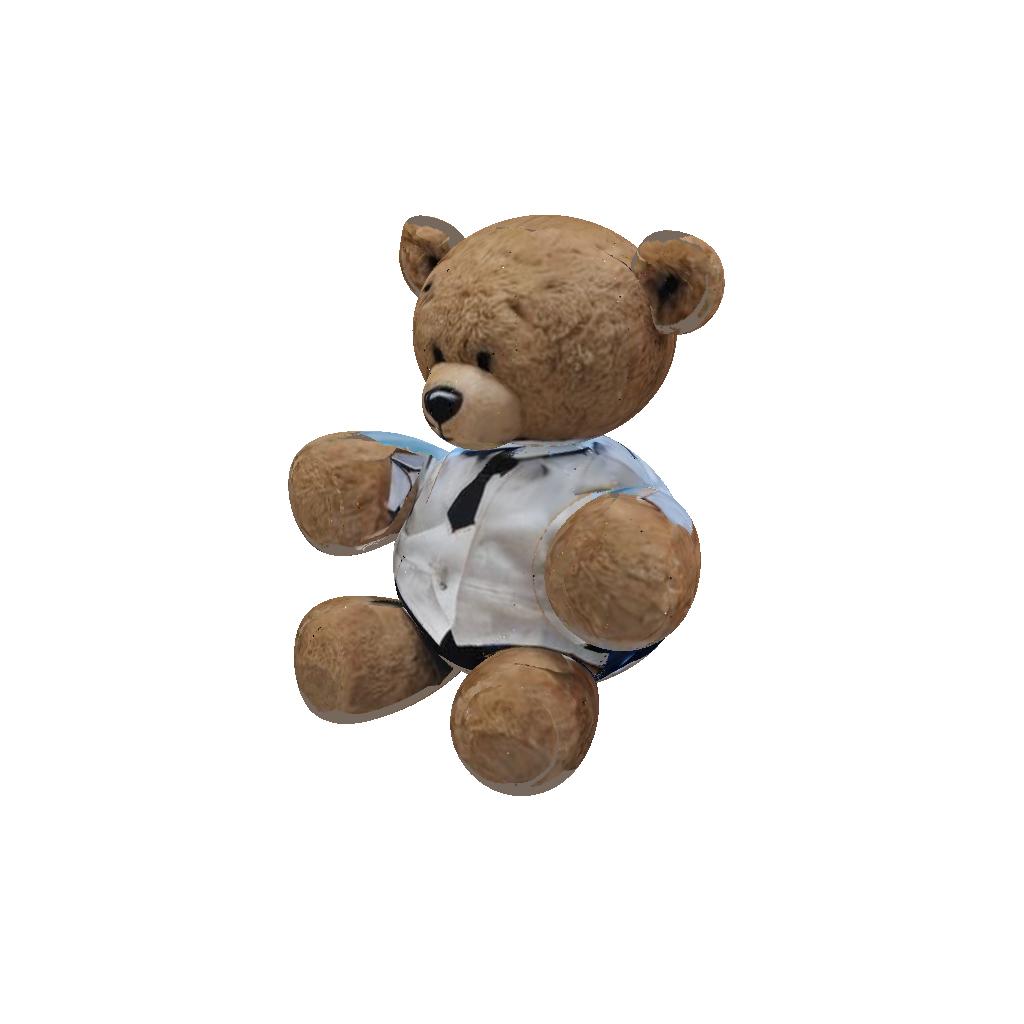} 
        \includegraphics[height=0.22\linewidth,trim={9cm 9cm 9cm 10cm},clip]
        {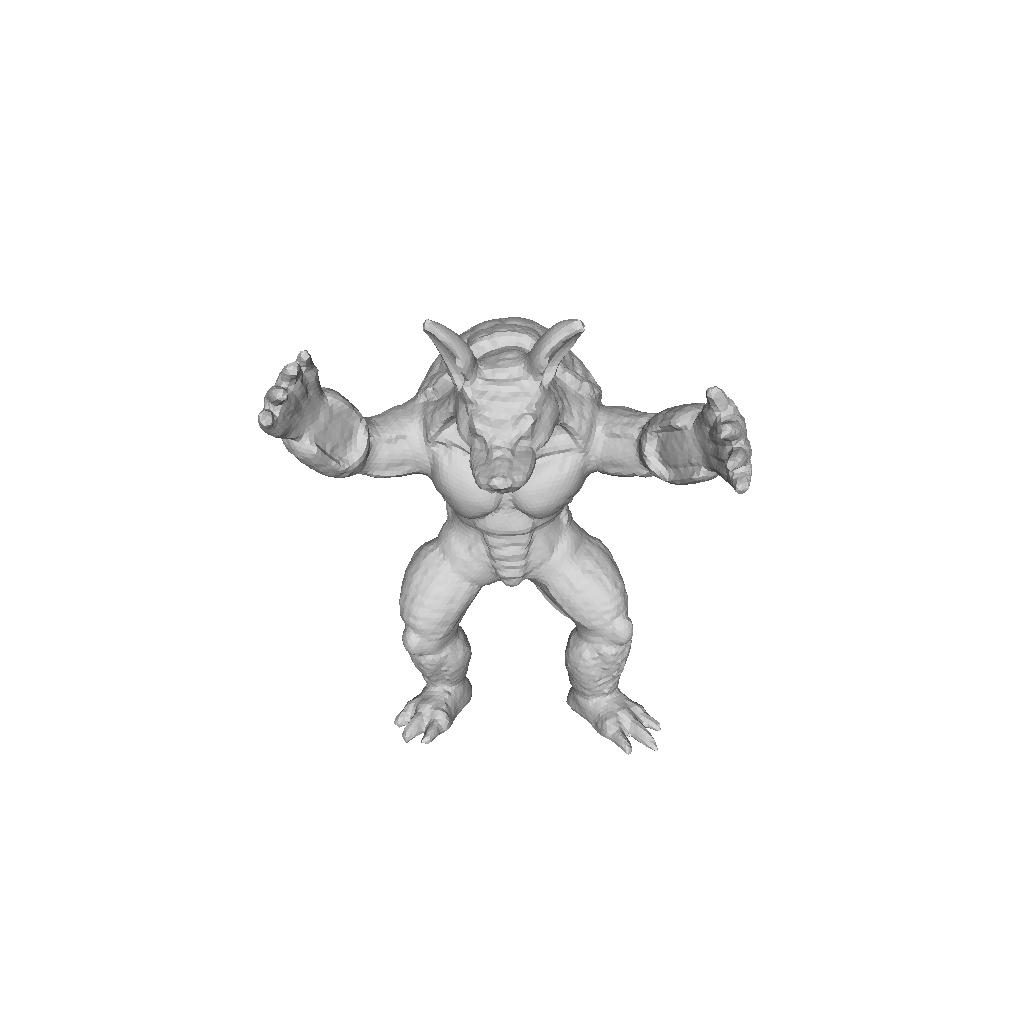} 
        \includegraphics[height=0.22\linewidth,trim={9cm 9cm 9cm 10cm},clip]
        {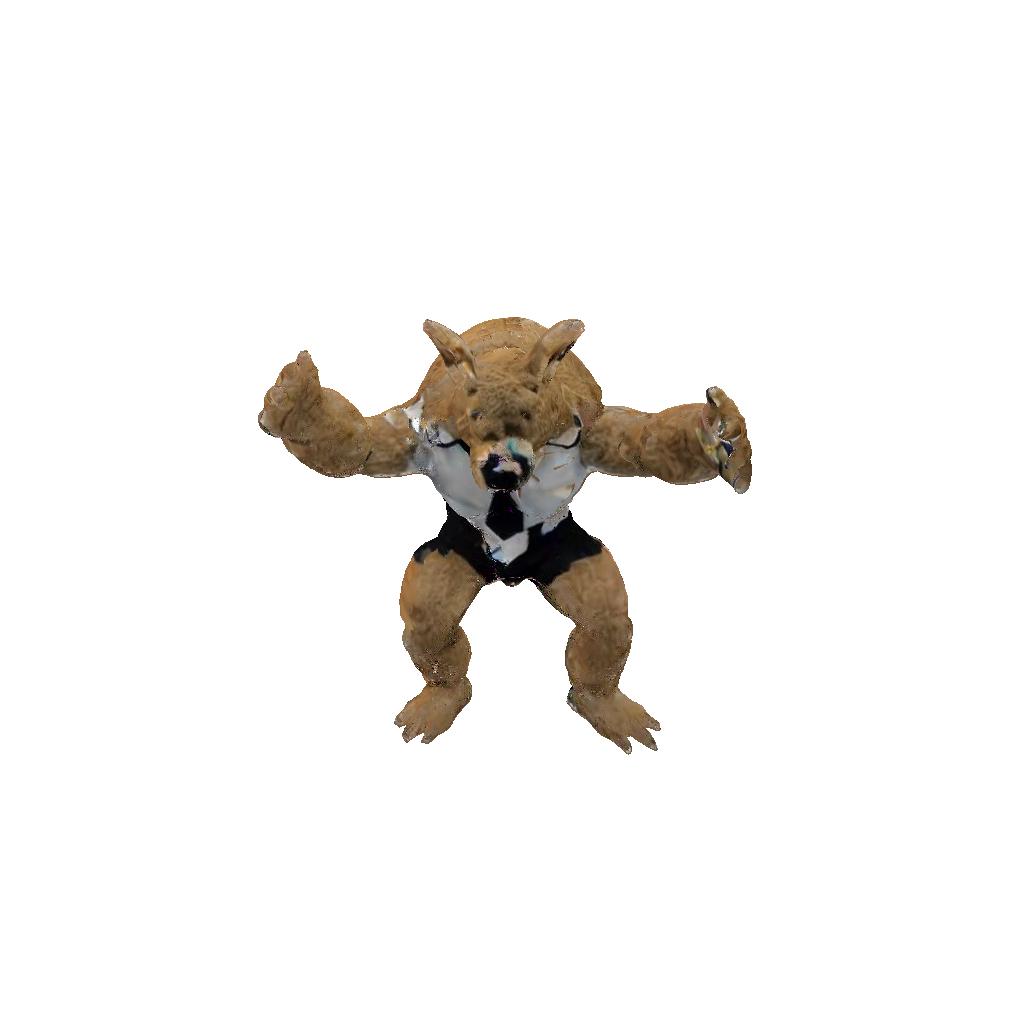} 
         \\
        \includegraphics[height=0.22\linewidth]
        {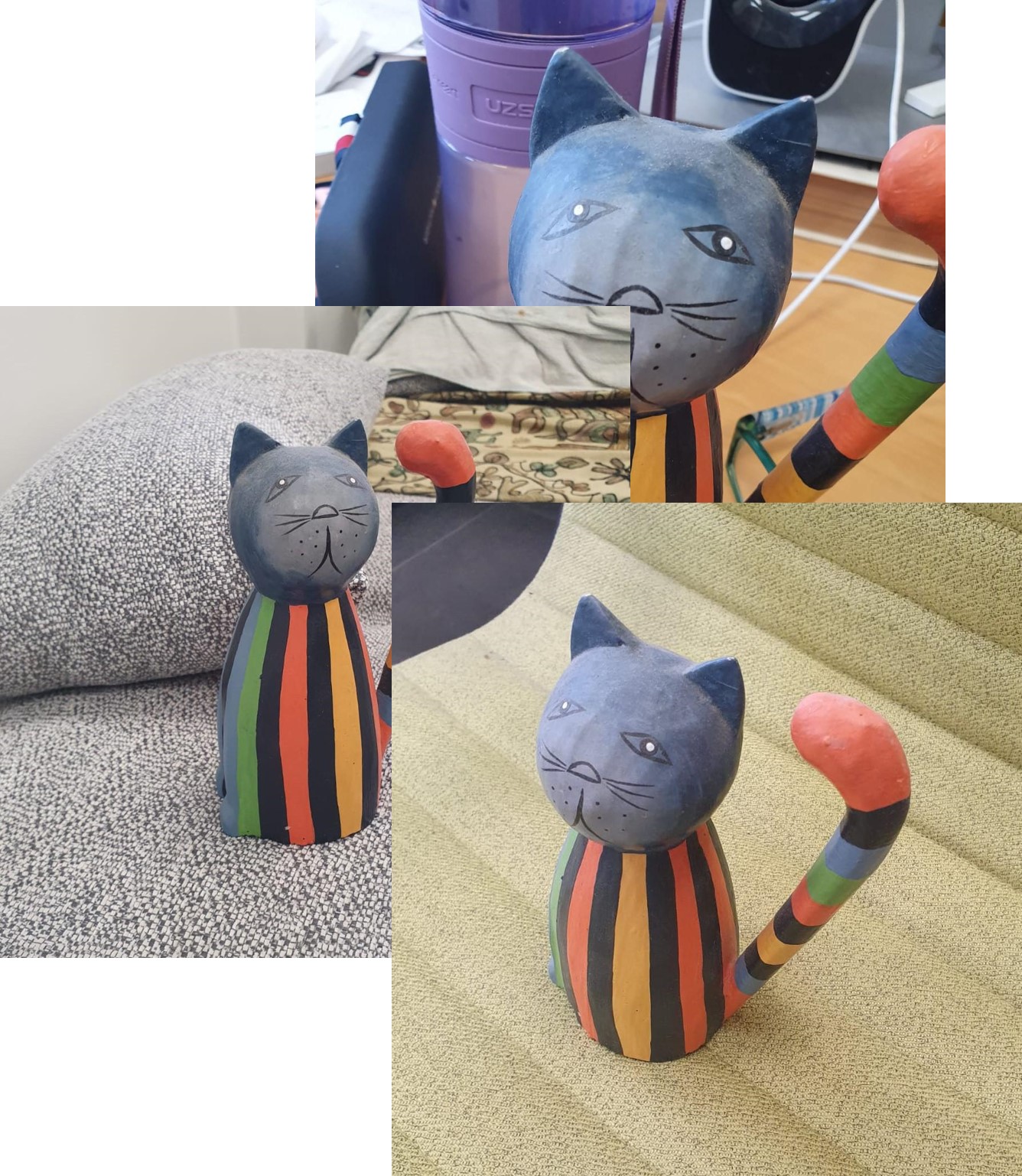} &
        \includegraphics[height=0.22\linewidth]
        {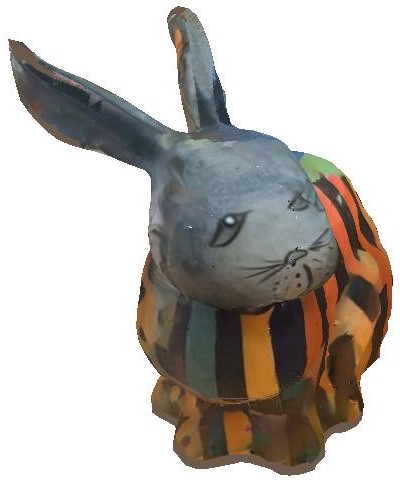} 
        \includegraphics[height=0.22\linewidth]
        {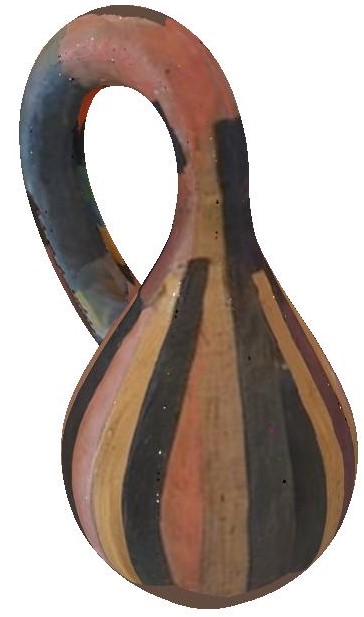} 
        \includegraphics[height=0.22\linewidth]
        {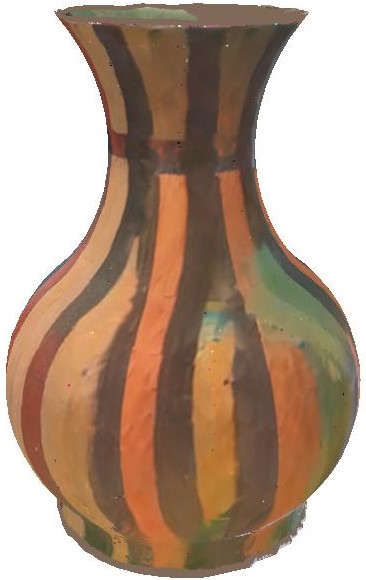} 
        \includegraphics[height=0.22\linewidth]
        {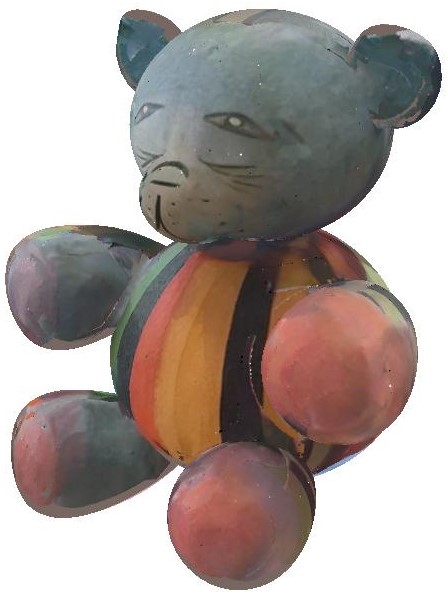} 
        \includegraphics[height=0.22\linewidth]{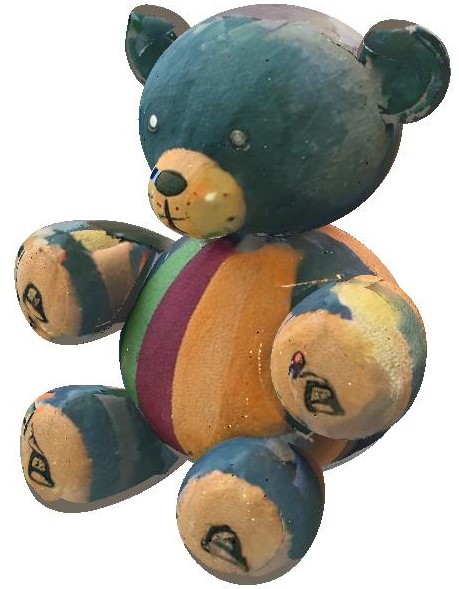} 
         \\
        Images Set & Textured Meshes
    \end{tabular}
    }
    \vspace{-0.3cm}
    \caption{Token-based Texture Transfer from images. New meshes are presented only once. All meshes are textured with the \textbf{exact} prompt. The teddy bear is also presented with the ``..in the style of'' prompt, one can see that this results in a semantic mix between the image set and a teddy bear. }
    \label{fig:images2mesh}
    \vspace{-0.2cm}
\end{figure}

%% file: sections/conclusion.tex
\input{figures/limitations/fig.tex}

\section{Discussion, Limitations and Conclusions}
This paper presents TEXTure, a novel method for text-guided generation, transfer, and editing of textures for 3D shapes. There have been many models for generating and editing high-quality images using diffusion models. However, leveraging these image-based models for generating seamless textures on 3D models is a challenging task, in particular for non-stochastic and non-stationary textures.
Our work addresses this challenge by introducing an iterative painting scheme that leverages a pretrained depth-to-image diffusion model. Instead of using a computationally demanding score distillation approach, we propose a modified image-to-image diffusion process that is applied from a small set of viewpoints. This results in a fast process capable of generating high-quality textures in mere minutes. With all its benefits, there are still some limitations to our proposed scheme, which we discuss next.

While our painting technique is designed to be  spatially coherent, it may sometimes result in inconsistencies on a global scale, caused by occluded information from other views. See~\Cref{fig:limitations}, where different looking eyes are added from different viewpoints. Another caveat is viewpoint selection. We use eight fixed viewpoints around the object, which may not fully cover adversarial geometries.
This issue can possibly be solved by finding a dynamic set of viewpoints that maximize the coverage of the  given mesh.
Furthermore, the depth-guided model sometimes deviates from the input depth, and may generate images that are not consistent with the geometry (See~\Cref{fig:limitations} left). 
This in turn may result in conflicting projections to the mesh, that cannot be fixed in later painting iterations.

With that being said, we believe that TEXTure takes an important step toward revolutionizing the field of graphic design and further opens new possibilities for 3D artists, game developers, and modelers who can use these tools to generate high-quality textures in a fraction of the time of existing techniques. Additionally, our trimap partitioning formulation provides a practical and useful framework that we hope will be utilized and \refined  in future studies.

%% file: figures/limitations/fig.tex
\begin{figure}
    \centering
    \setlength{\belowcaptionskip}{-4pt}
    \setlength{\tabcolsep}{0pt}
    \newcommand{\pl}{0.19}
    
    {\small
    \begin{tabular}{c c}
        \includegraphics[width=\pl\linewidth]{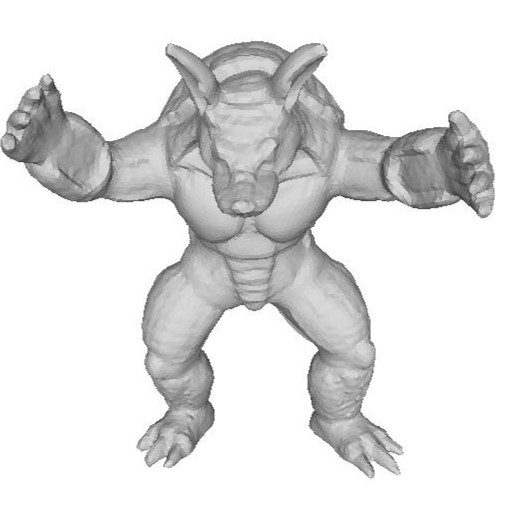} 
        \includegraphics[width=\pl\linewidth]{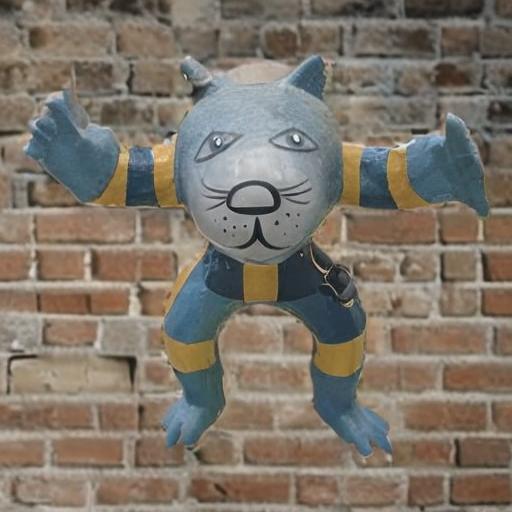} 
        \includegraphics[width=\pl\linewidth]{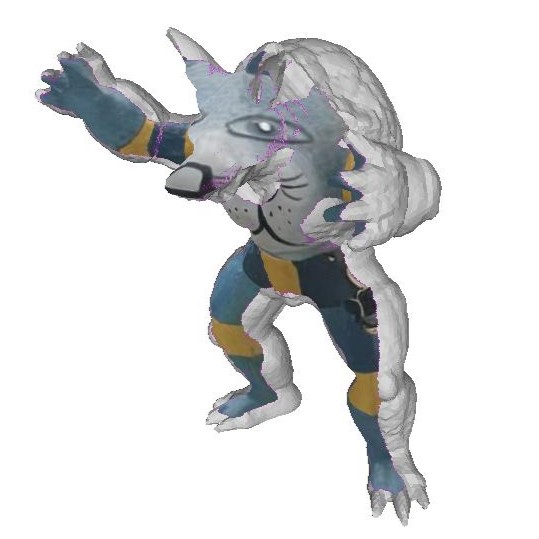} &
        \includegraphics[width=\pl\linewidth]{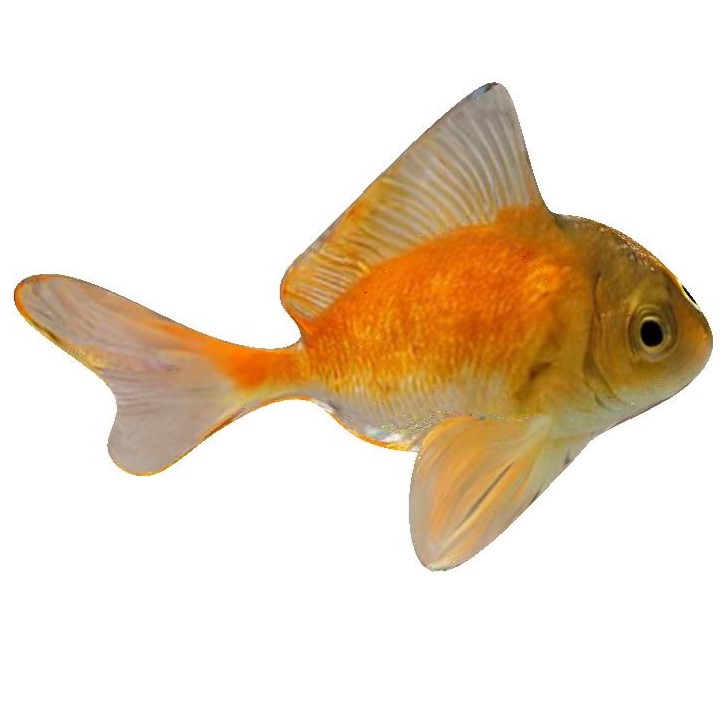} 
        \includegraphics[width=\pl\linewidth]{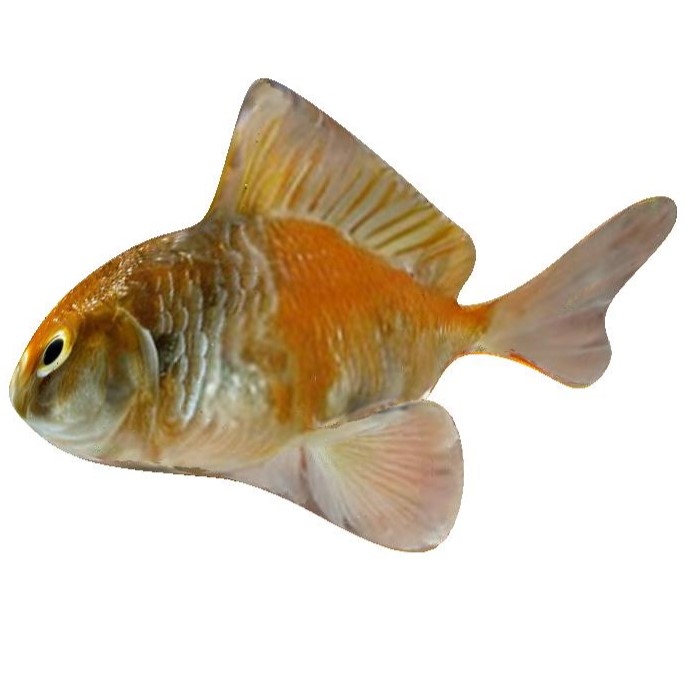} 
         \\
         Depth Inconsistencies & Viewpoint Inconsistencies
    \end{tabular}
    }
    \vspace{-0.3cm}
    \caption{Limitations. Left: input mesh, where the diffusion output was inconsistent with the geometry of the model face. These inconsistencies are clearly visible from other views and affect the painting process. Right: two different viewpoints of ``a goldfish'' where the diffusion process added inconsistent eyes, as the previously painted eyes were not visible.
    }
    \label{fig:limitations}
    \vspace{-0.2cm}
\end{figure}

%% file: figures/paint_additional_results/fig.tex
\begin{figure*}
\centering
    \centering
    \setlength{\tabcolsep}{0pt}
    {\small

    \begin{tabular}{c c c c c c}
        \includegraphics[height=0.13\linewidth,trim={10cm 9cm 6cm 8cm},clip]{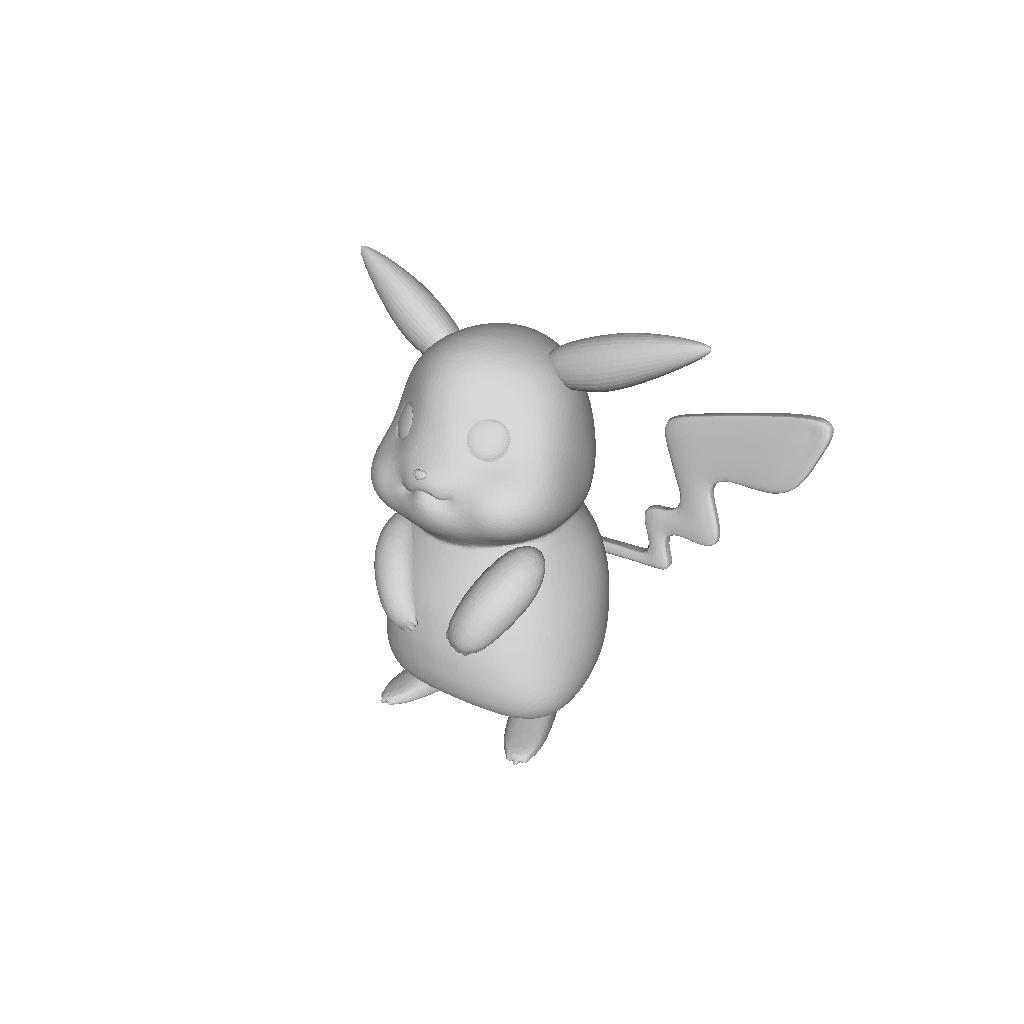} &
        \includegraphics[height=0.13\linewidth,trim={10cm 9cm 6cm 8cm},clip]{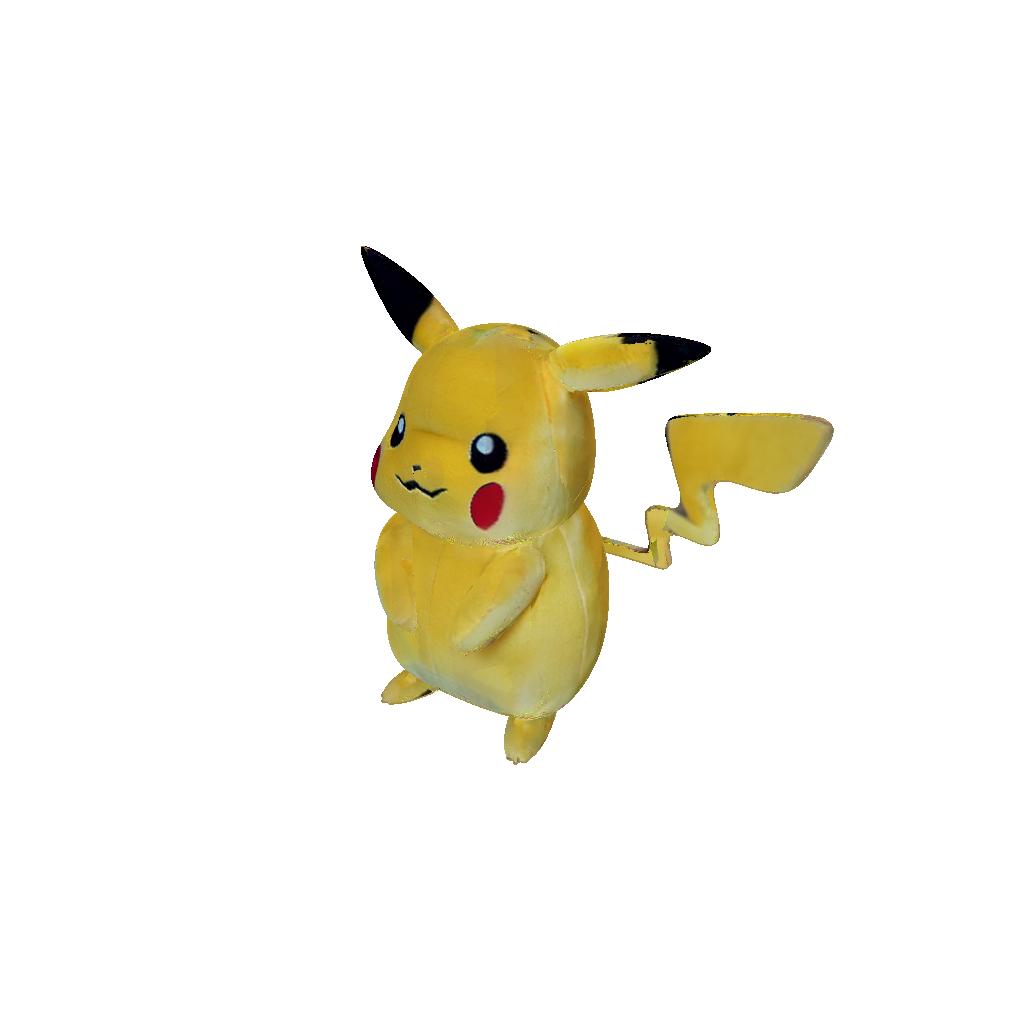} &
        \includegraphics[height=0.13\linewidth,trim={10cm 9cm 6cm 8cm},clip]{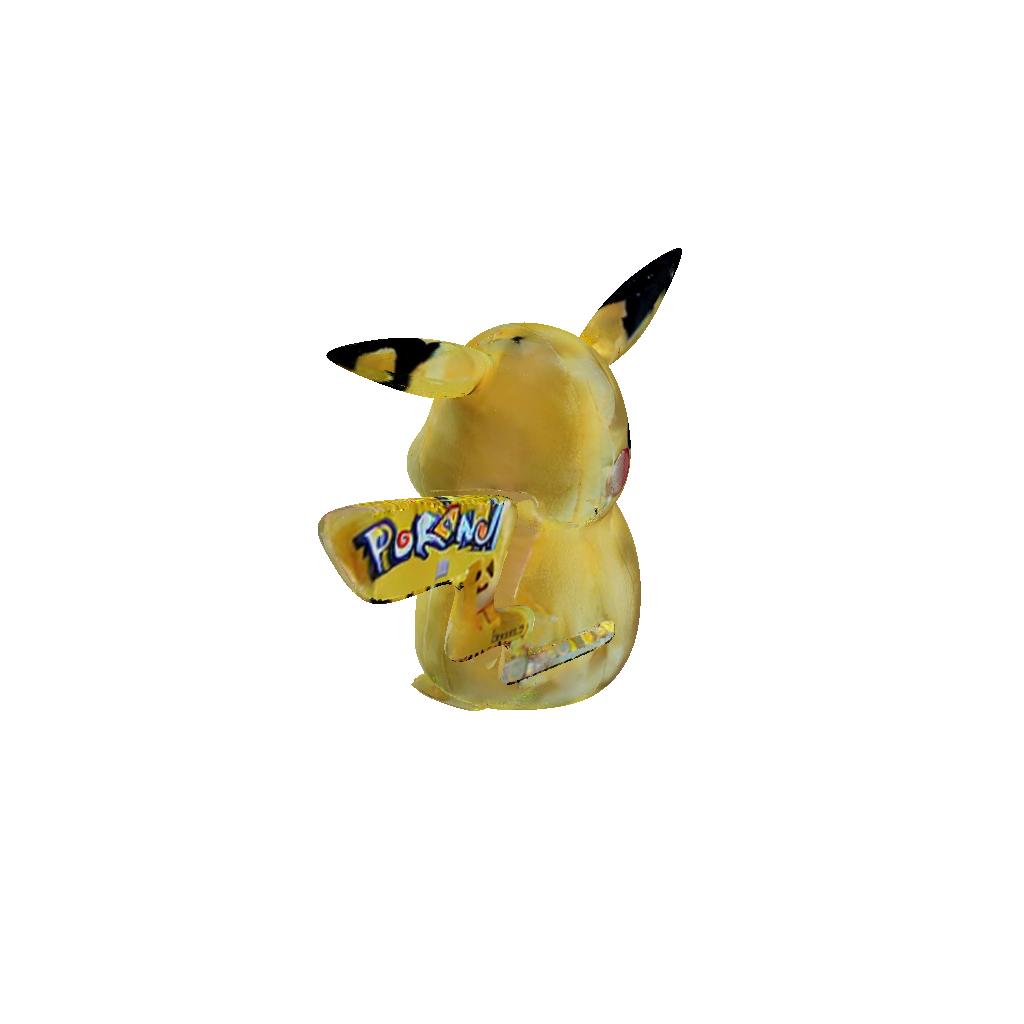} &
        \includegraphics[height=0.13\linewidth,trim={8cm 9.5cm 7cm 8cm},clip]{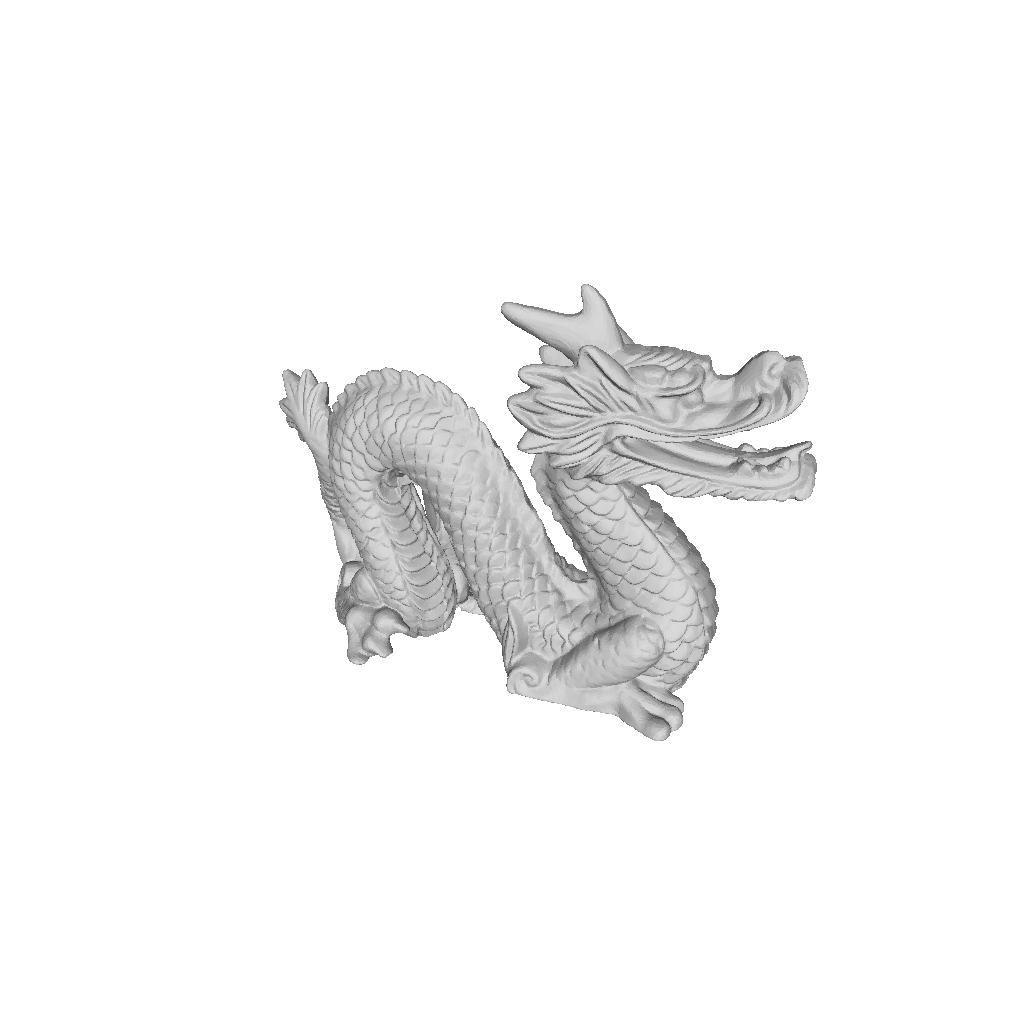} &
        \includegraphics[height=0.13\linewidth,trim={8cm 9.5cm 7cm 8cm},clip]{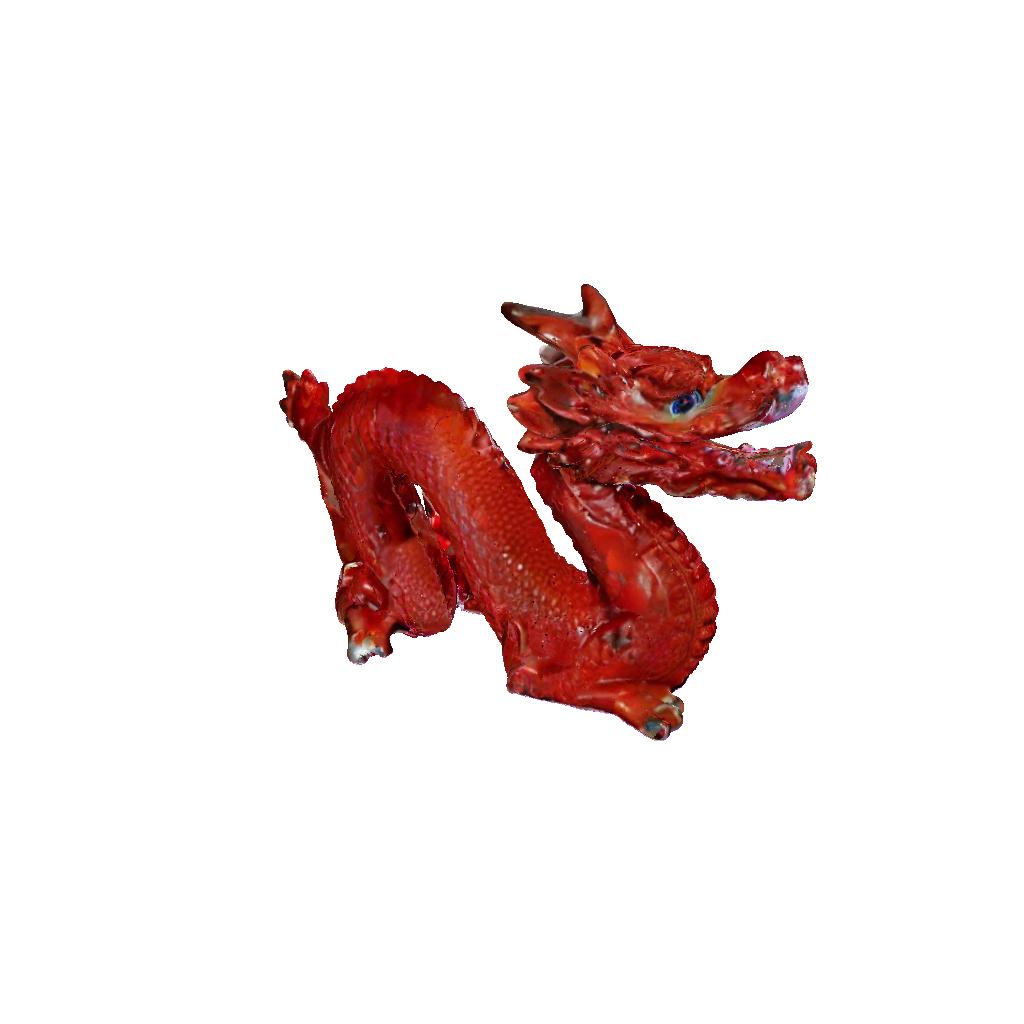} &
        \includegraphics[height=0.13\linewidth,trim={8cm 9.5cm 7cm 8cm},clip]{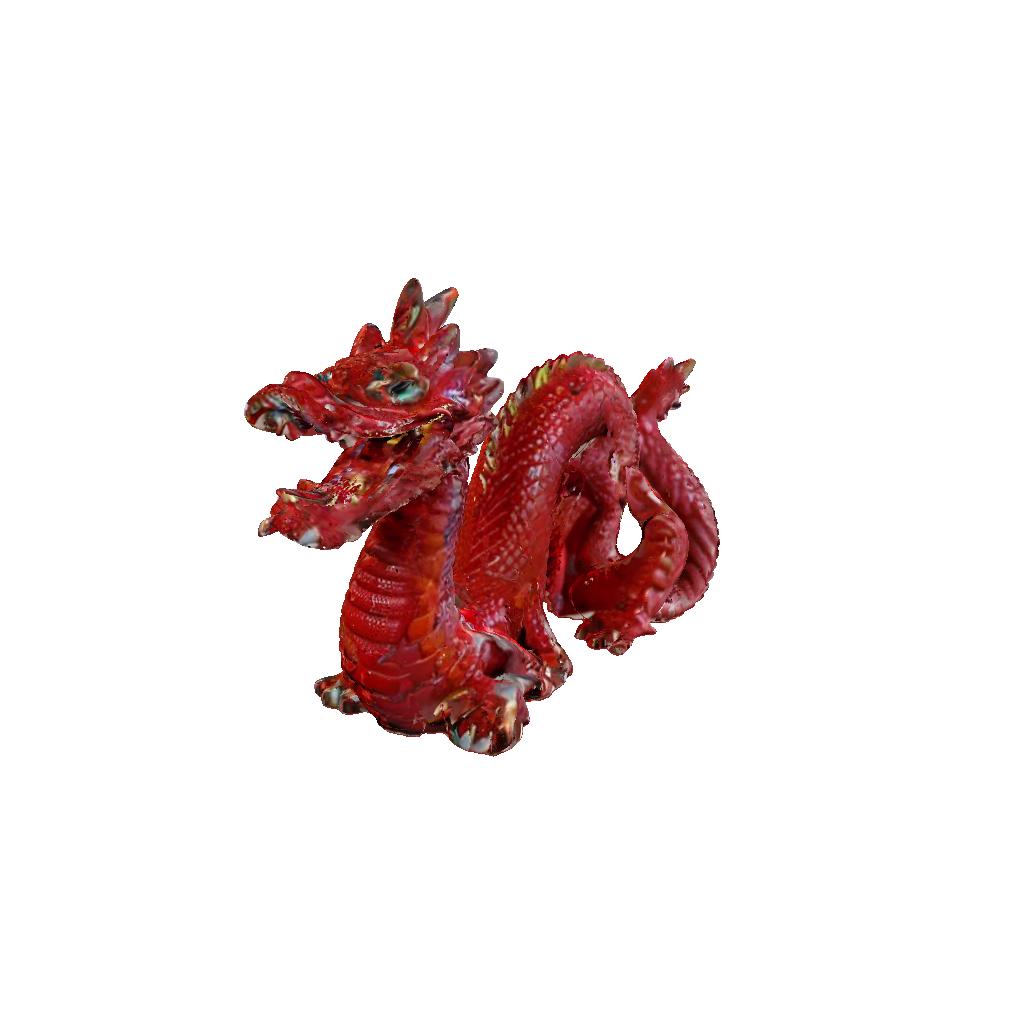} \\
         & \multicolumn{2}{c}{``A plush pikachu''} &  & \multicolumn{2}{c}{``A red porcelain dragon''}
    \end{tabular}

        \begin{tabular}{c c c c c c c c c}
        \includegraphics[height=0.105\linewidth]{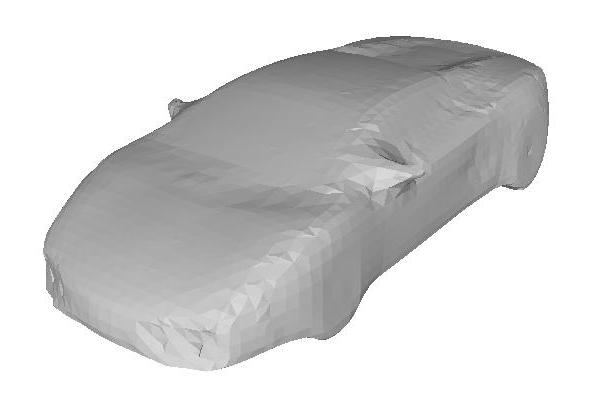} &
        \includegraphics[height=0.105\linewidth]{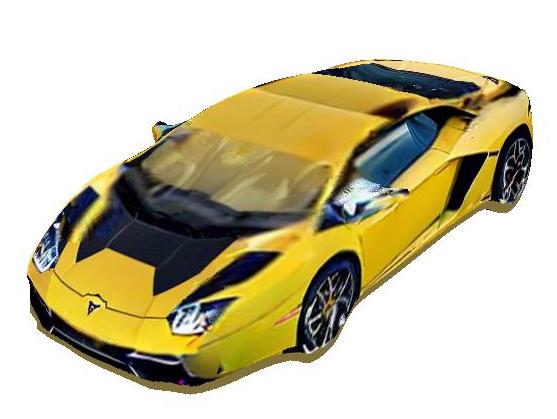} &
        \includegraphics[height=0.105\linewidth]{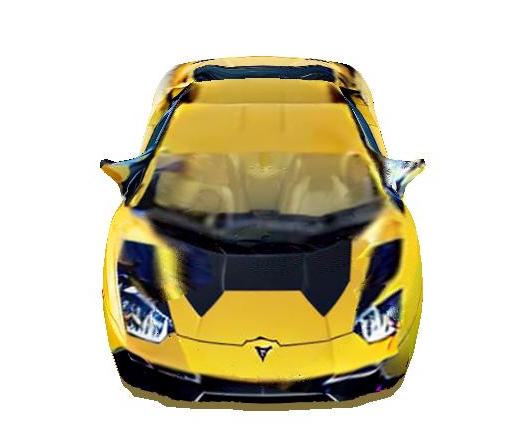} &
        \includegraphics[height=0.105\linewidth]{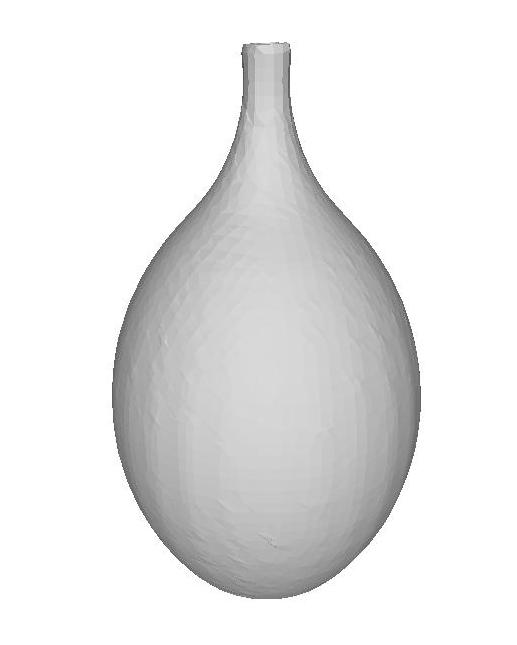} &
        \includegraphics[height=0.105\linewidth]{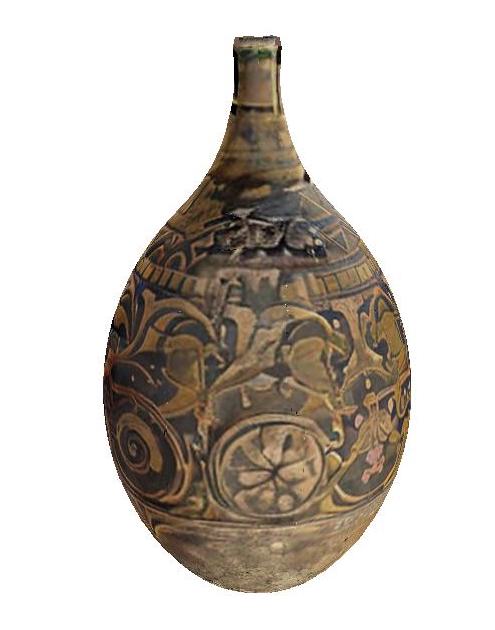} &
        \includegraphics[height=0.105\linewidth]{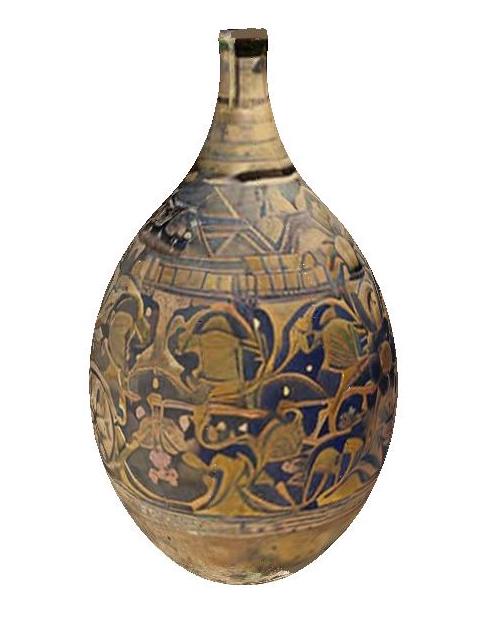} &
        \includegraphics[height=0.105\linewidth]{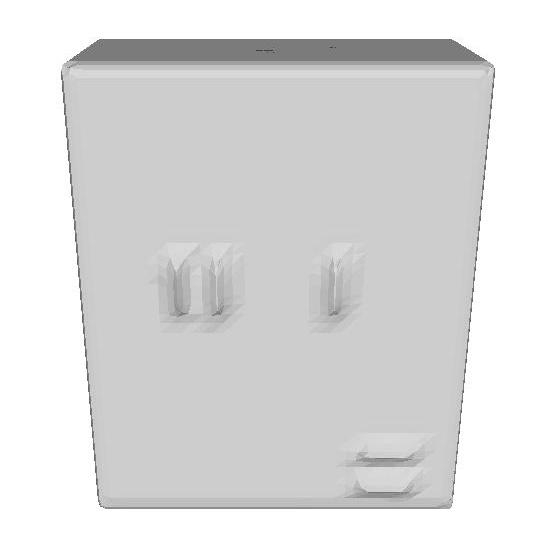} &
        \includegraphics[height=0.105\linewidth]{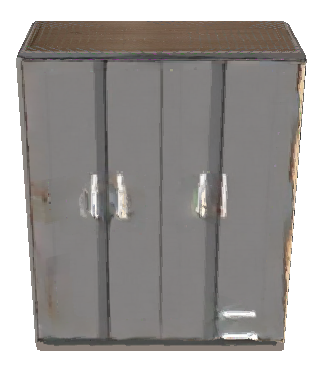} &
        \includegraphics[height=0.105\linewidth]{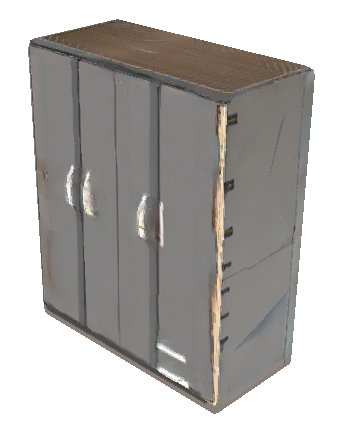} \\
         & \multicolumn{2}{c}{``A yellow Lamborghini''} & & \multicolumn{2}{c}{``An ancient vase''} & & \multicolumn{2}{c}{``An ikea wardrobe''}  \\ 
    \end{tabular}
    
    \begin{tabular}{c c @{\hskip 3pt} c  @{\hskip 3pt} c}
        \includegraphics[height=0.13\linewidth]{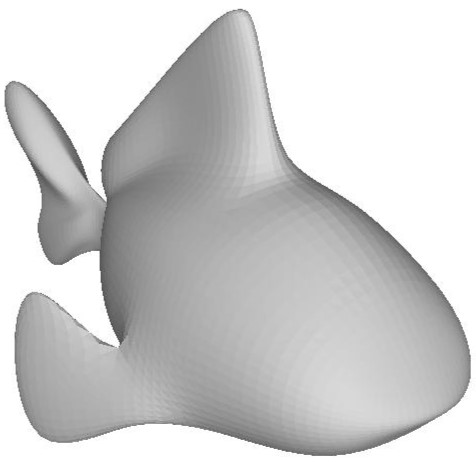} &
        \includegraphics[height=0.13\linewidth]{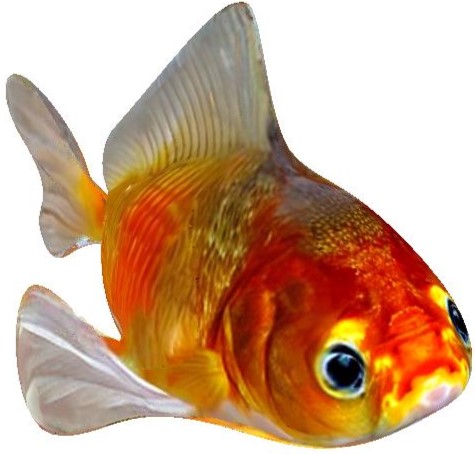} 
        \includegraphics[height=0.13\linewidth]{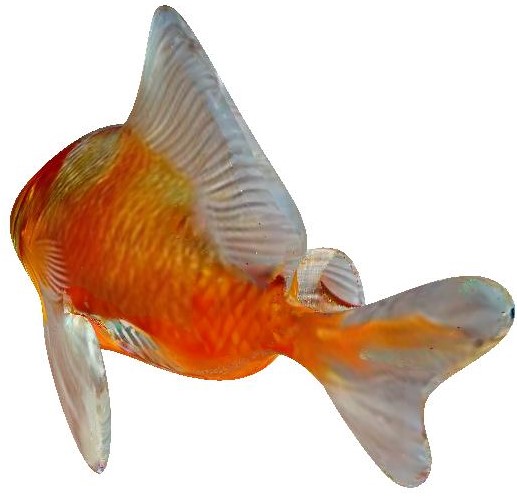} &
        \includegraphics[height=0.13\linewidth]{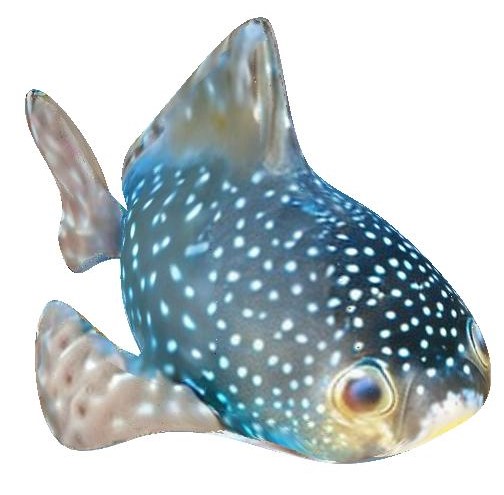} 
        \includegraphics[height=0.13\linewidth]{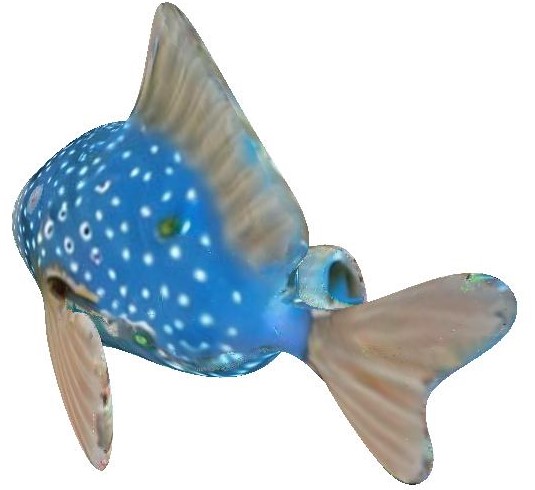} &
        \includegraphics[height=0.13\linewidth]{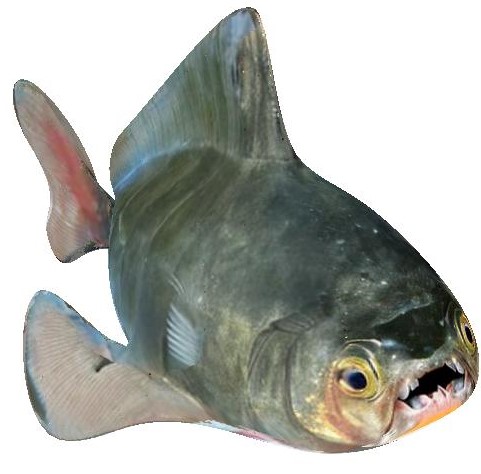} 
        \includegraphics[height=0.13\linewidth]{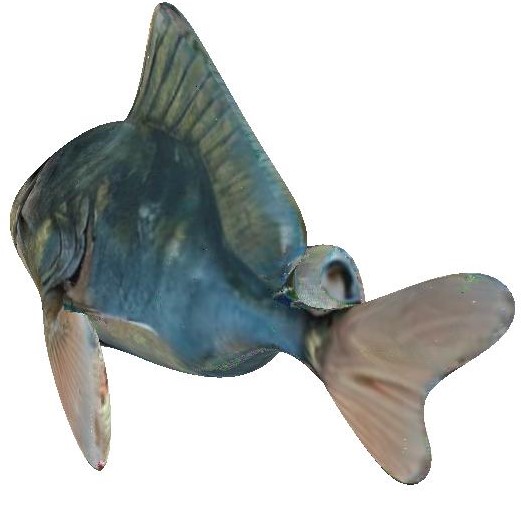} 
        \\
         Input Mesh & ``A goldfish'' & ``A puffer fish''  & ``A piranha fish''
    \\
    \end{tabular}
    
    \begin{tabular}{c  c}
        \includegraphics[height=0.13\linewidth,trim={12cm 11cm 10cm 12cm},clip]{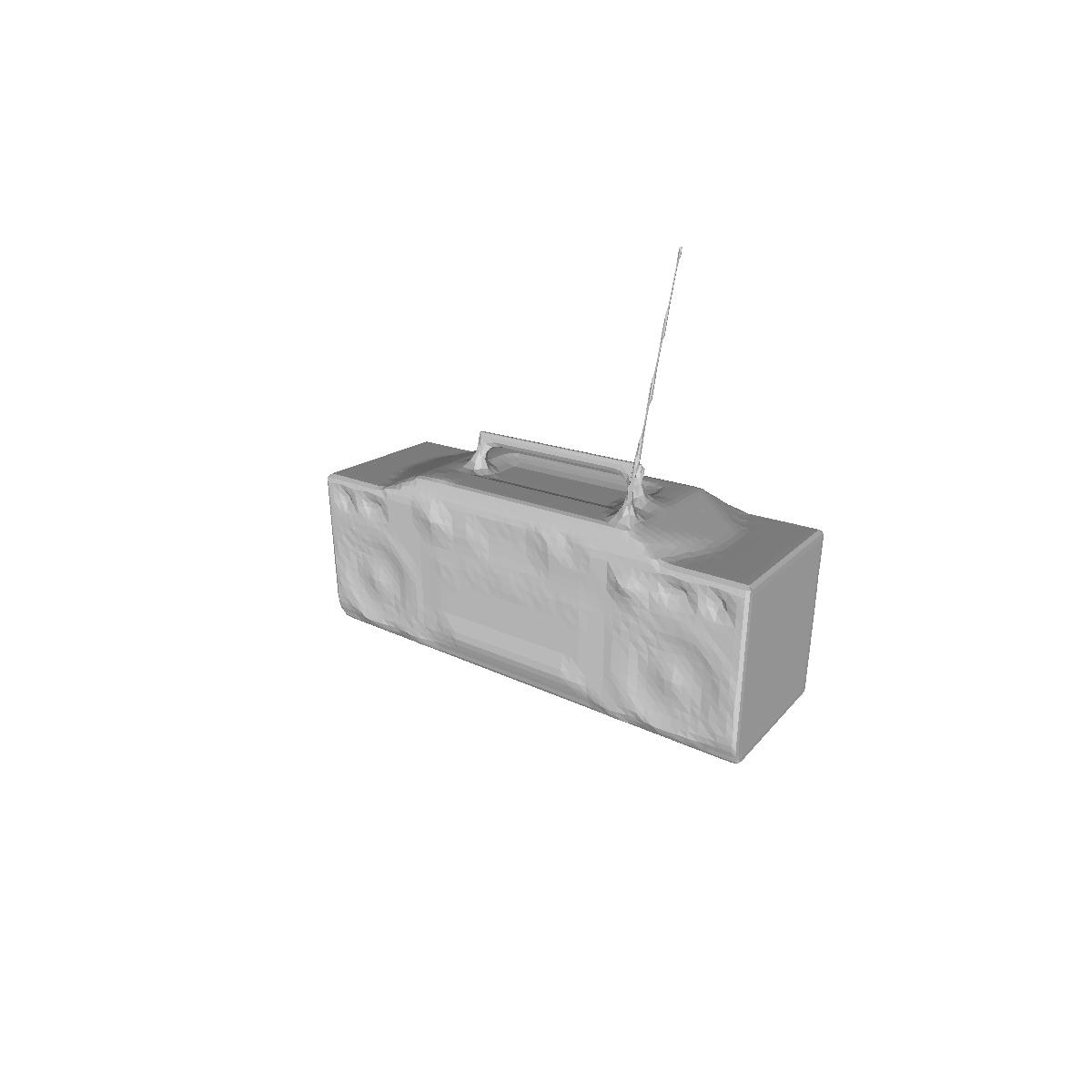} &
        \includegraphics[height=0.13\linewidth,trim={12cm 11cm 10cm 12cm},clip]{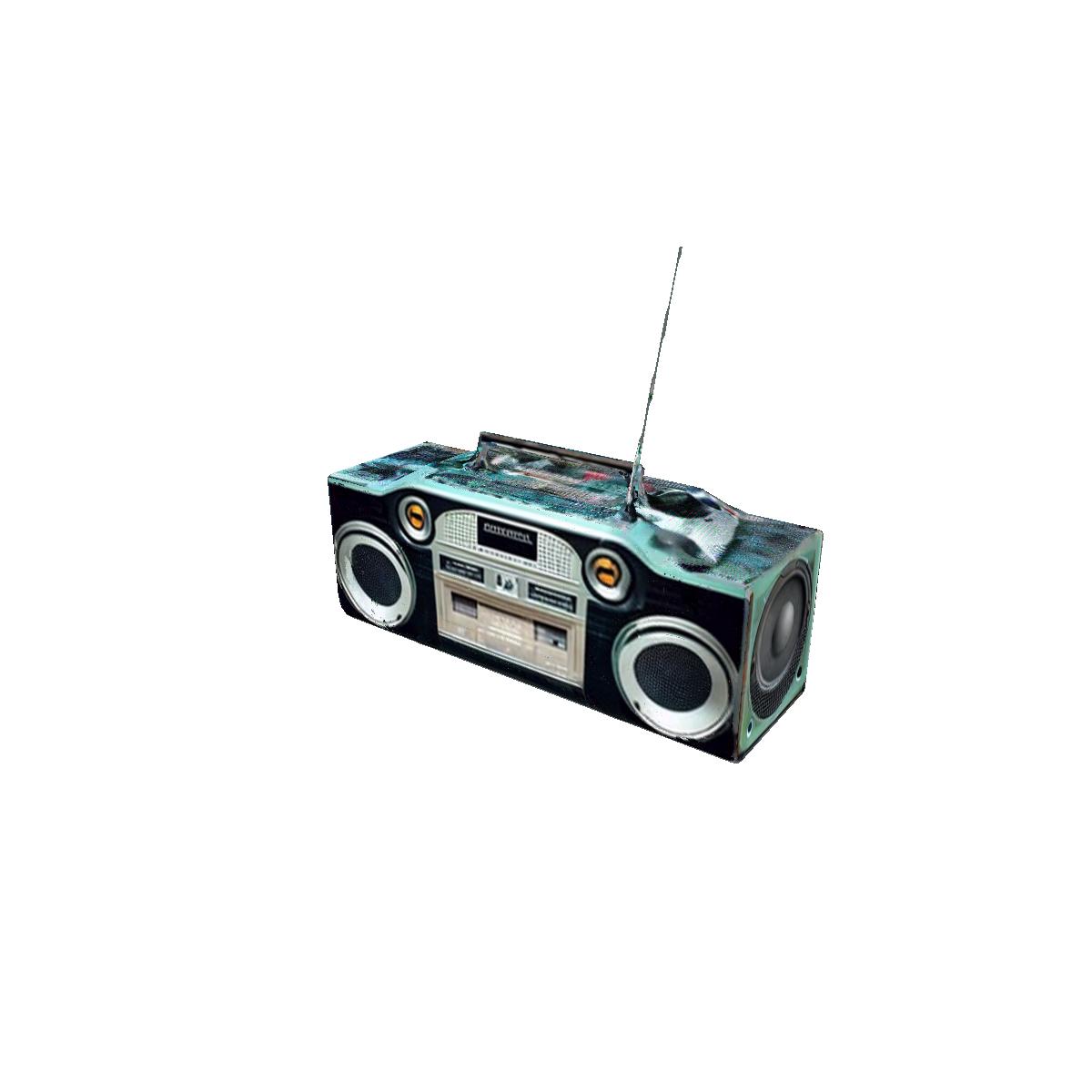}
        \includegraphics[height=0.13\linewidth,trim={12cm 11cm 10cm 12cm},clip]{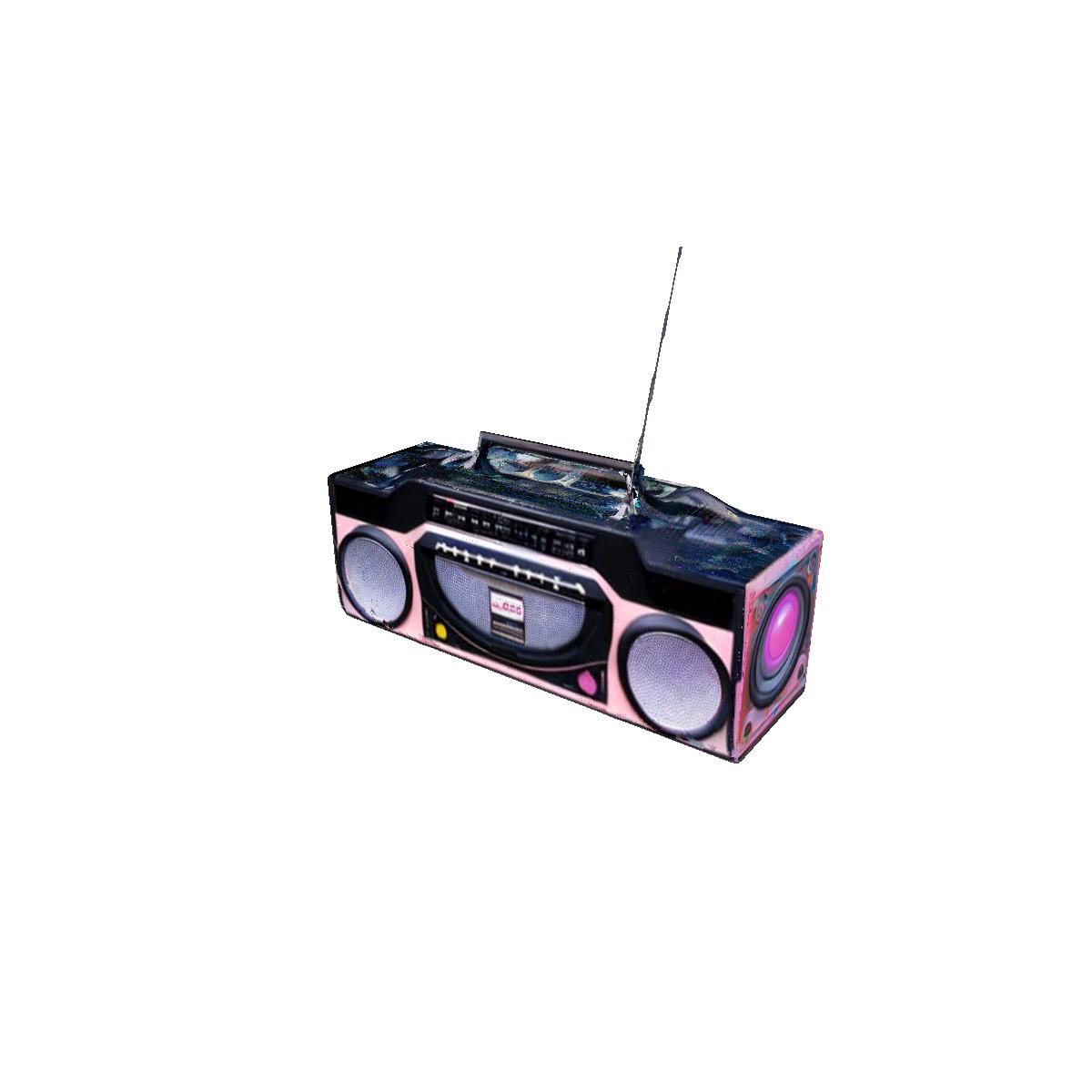}
        \includegraphics[height=0.13\linewidth,trim={12cm 11cm 10cm 12cm},clip]{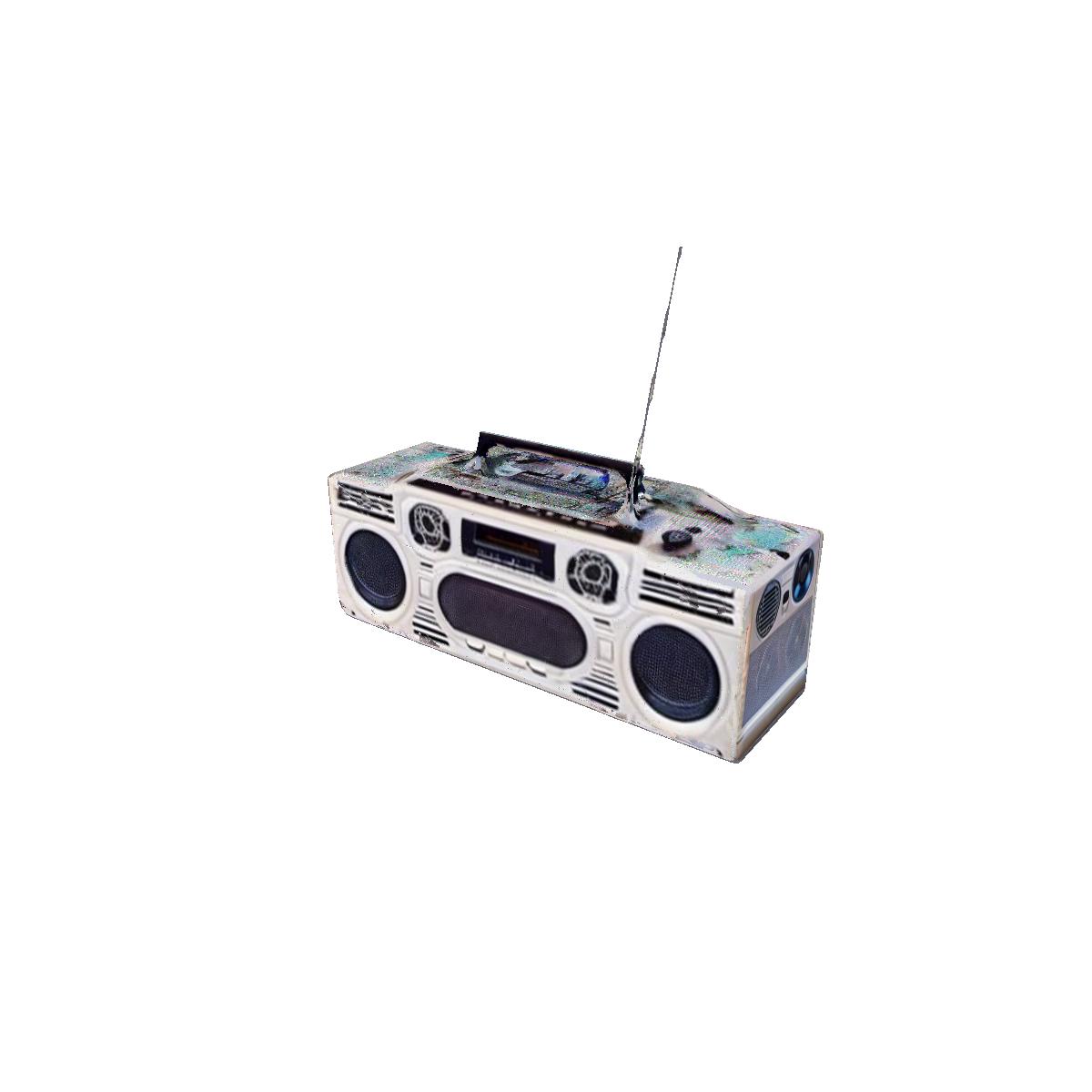}
        \includegraphics[height=0.13\linewidth,trim={12cm 11cm 10cm 12cm},clip]{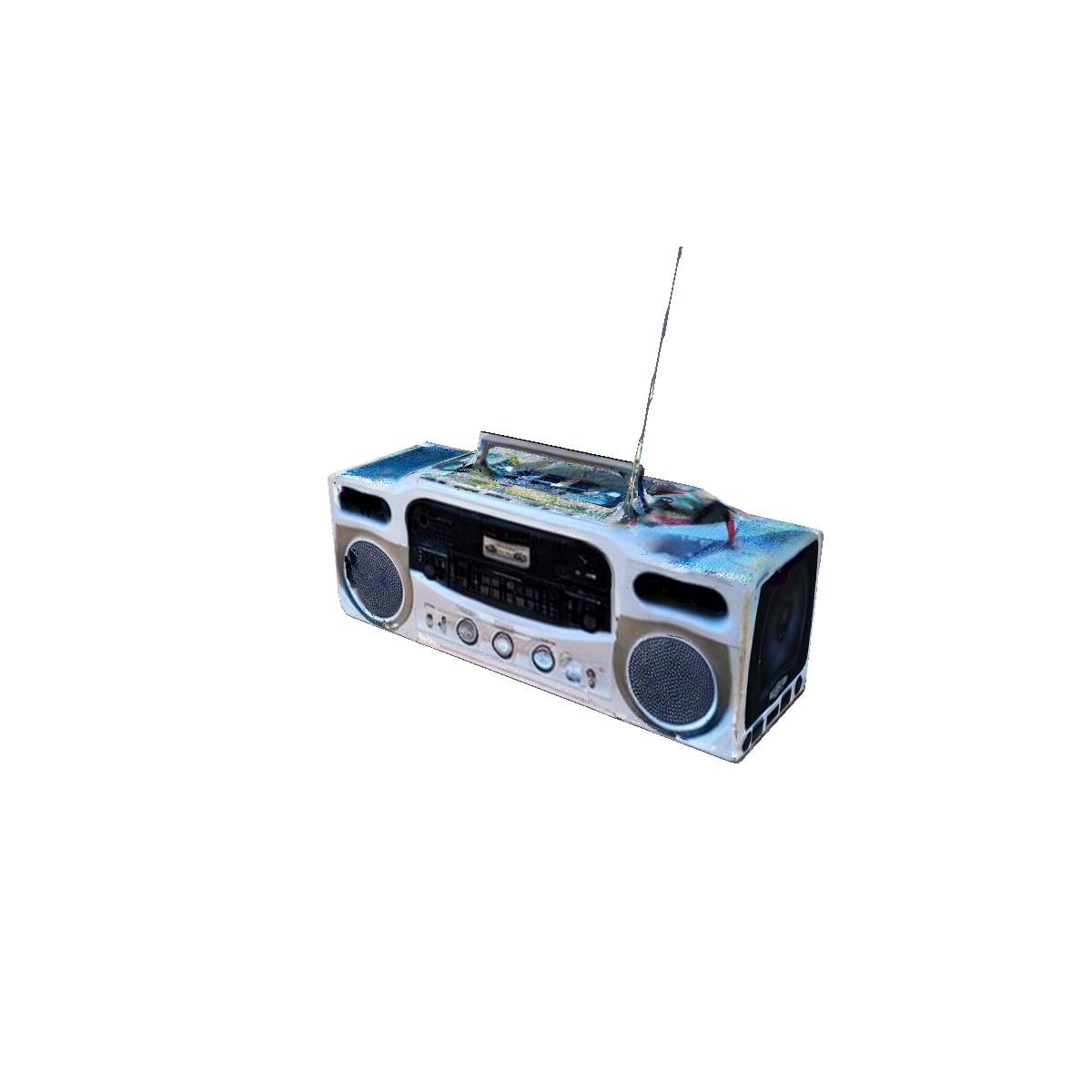}
        \includegraphics[height=0.13\linewidth,trim={12cm 11cm 10cm 12cm},clip]{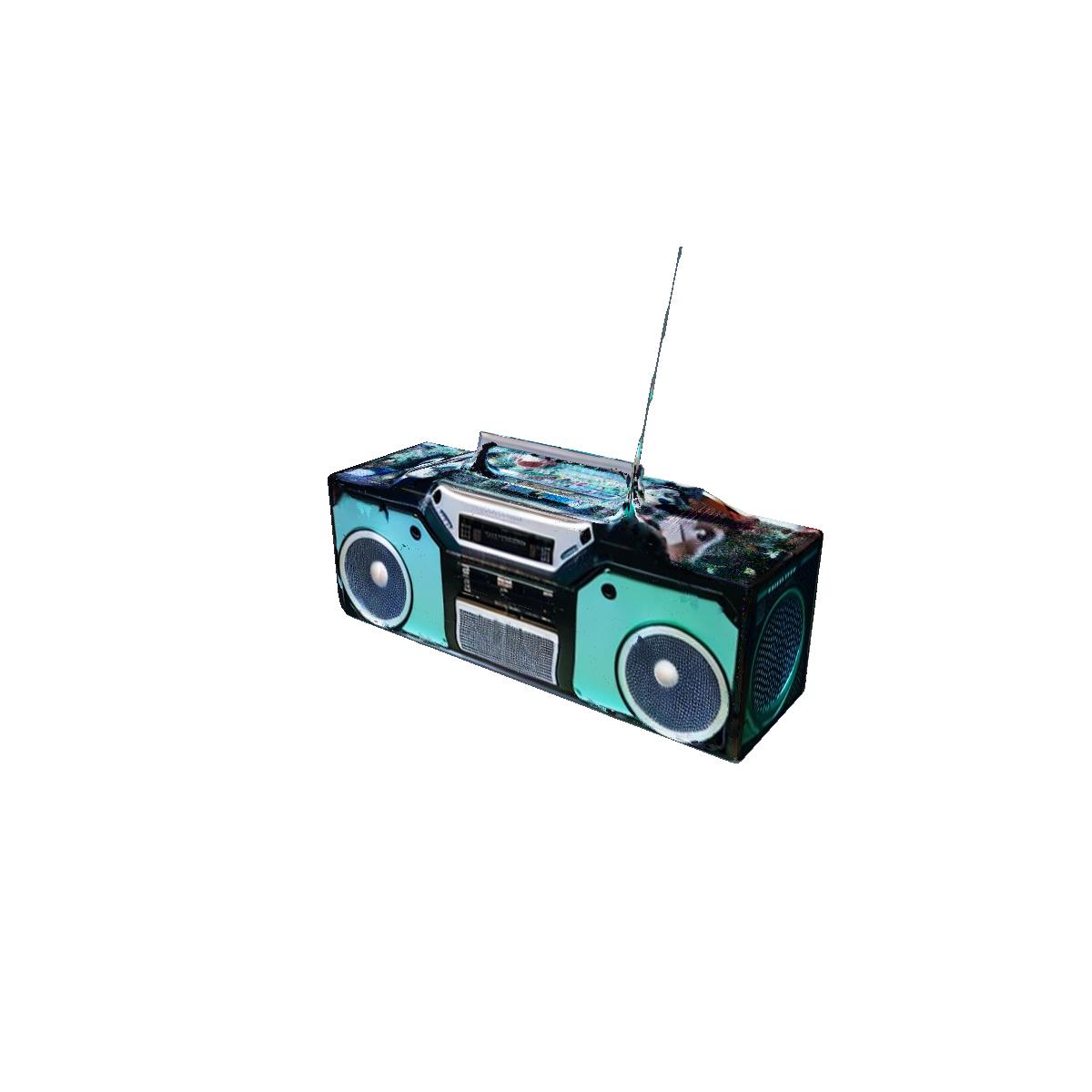}
        \includegraphics[height=0.13\linewidth,trim={12cm 11cm 10cm 12cm},clip]{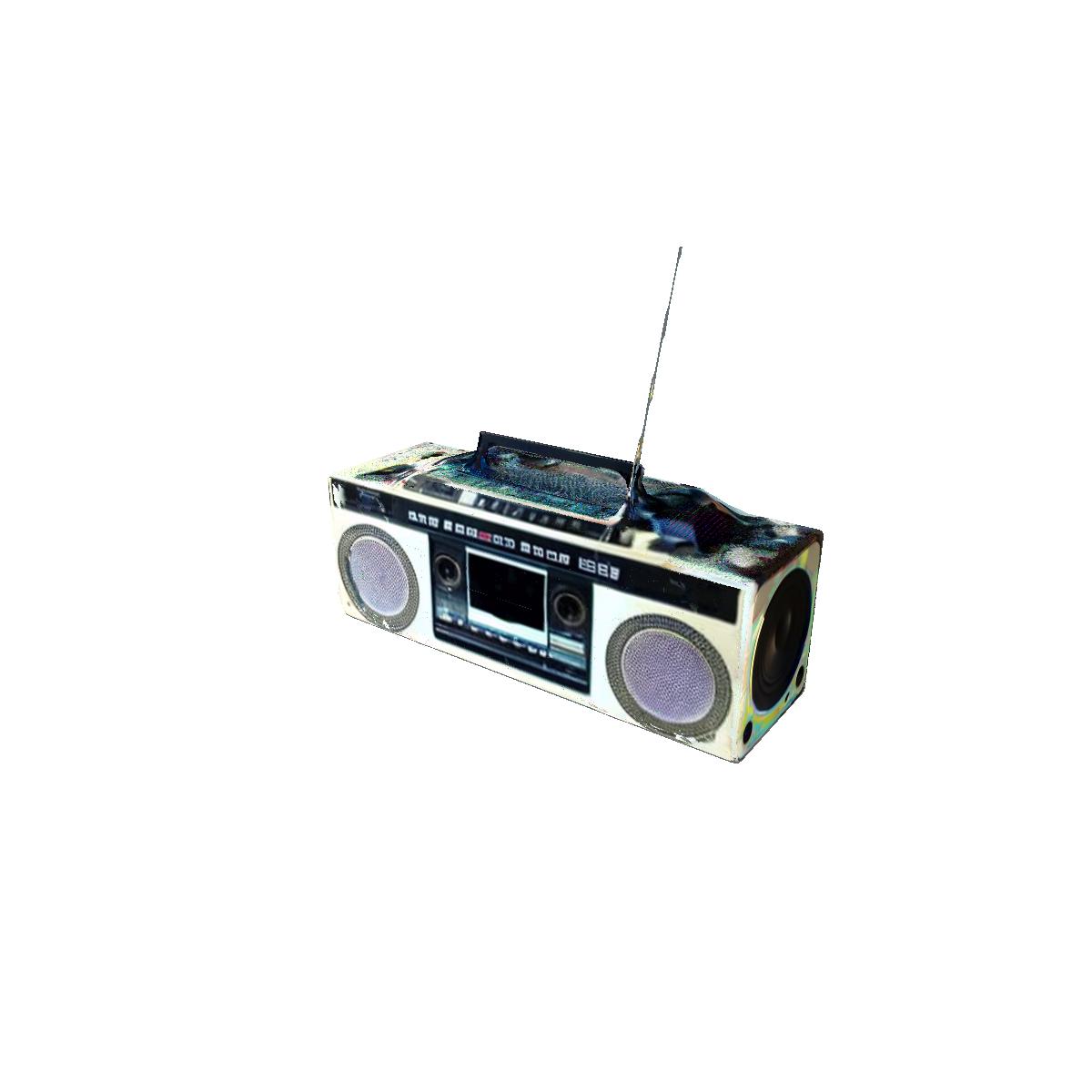}
        \\
         Input Mesh &  ``A 90s boombox'' under various seeds
    \end{tabular}
    }
    \vspace{-0.1cm}
    \caption{Additional texturing results achieved with TEXTure.}
    \label{fig:more_paint_results}
\end{figure*}

%% file: figures/experiments/joint_fig.tex
\begin{figure*}
    \begin{minipage}{.45\linewidth}
    \newcommand{\pl}{0.2}
    \newcommand{\pll}{1pt}
    \newcommand{\rl}{0.050\textwidth}
    \centering
    \setlength{\tabcolsep}{5pt}
    {\small
    \begin{tabular}{c c c c}

        \includegraphics[width=\pl\linewidth]{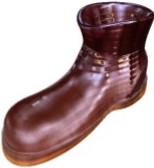} \hspace{\pll} & 
        \includegraphics[width=\pl\linewidth]{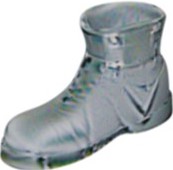} \hspace{\pll} & 
        \includegraphics[width=\pl\linewidth]{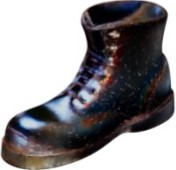} \hspace{\pll} &  
        \includegraphics[width=\pl\linewidth]{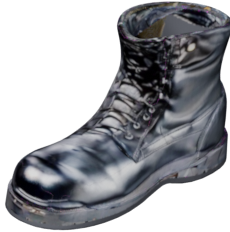}\\
        \multicolumn{4}{c}{``A black boot''} \\
        
        \includegraphics[width=\pl\linewidth]{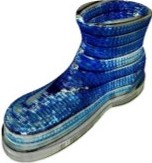} \hspace{\pll} & 
        \includegraphics[width=\pl\linewidth]{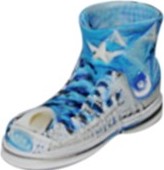} \hspace{\pll} & 
        \includegraphics[width=\pl\linewidth]{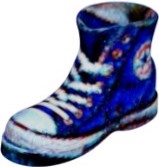} \hspace{\pll} &  
        \includegraphics[width=\pl\linewidth]{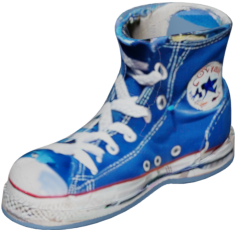} \\
        \multicolumn{4}{c}{``A blue converse allstar shoe''} \\
        
        \includegraphics[width=\pl\linewidth]{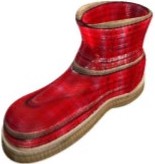} \hspace{\pll} & 
        \includegraphics[width=\pl\linewidth]{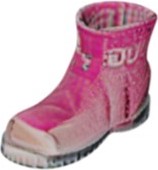} \hspace{\pll} & 
        \includegraphics[width=\pl\linewidth]{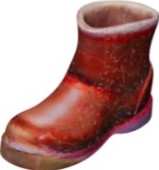} \hspace{\pll} &  
        \includegraphics[width=\pl\linewidth]{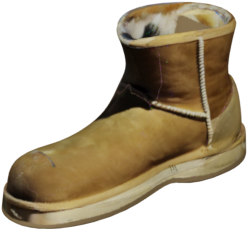}\\
        \multicolumn{4}{c}{``An UGG boot''} \\ \\
        
        Tango \hspace{\pll}& CLIPMesh \hspace{\pll} & Latent-Paint \hspace{\pll} & Ours

    \end{tabular}}
    \end{minipage} 
    \begin{minipage}{.55\linewidth}
    \centering
    {\small
    \begin{tabular}{c c c c c}

        {\raisebox{0.1in}{\rotatebox{90}{Text2Mesh}}} &
        \includegraphics[width=0.2\linewidth]{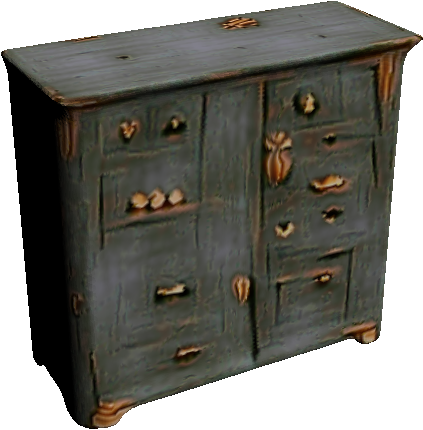} & 
        \includegraphics[width=0.2\linewidth]{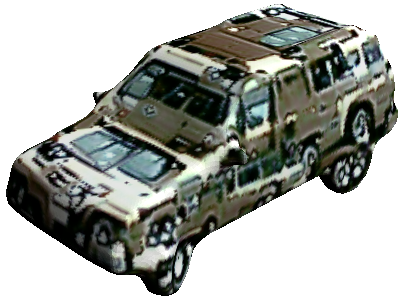} & 
        \includegraphics[width=0.2\linewidth]{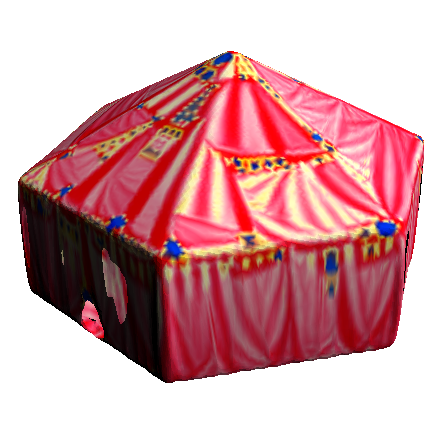} & 
        \includegraphics[width=0.2\linewidth]{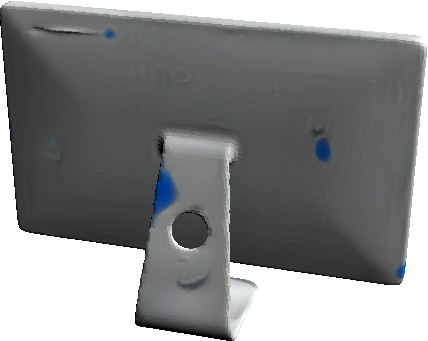} \\ 

        {\raisebox{0.1in}{\rotatebox{90}{Latent-Paint}}} &
        \includegraphics[width=0.2\linewidth]{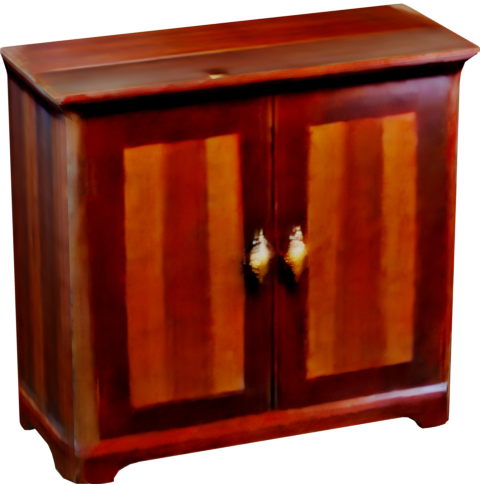} & 
        \includegraphics[width=0.2\linewidth]{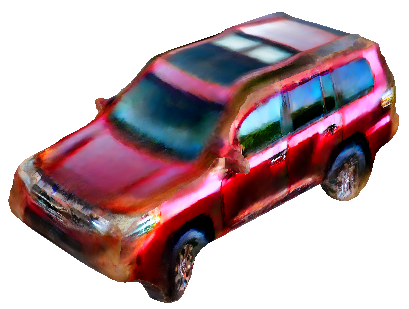} & 
        \includegraphics[width=0.2\linewidth]{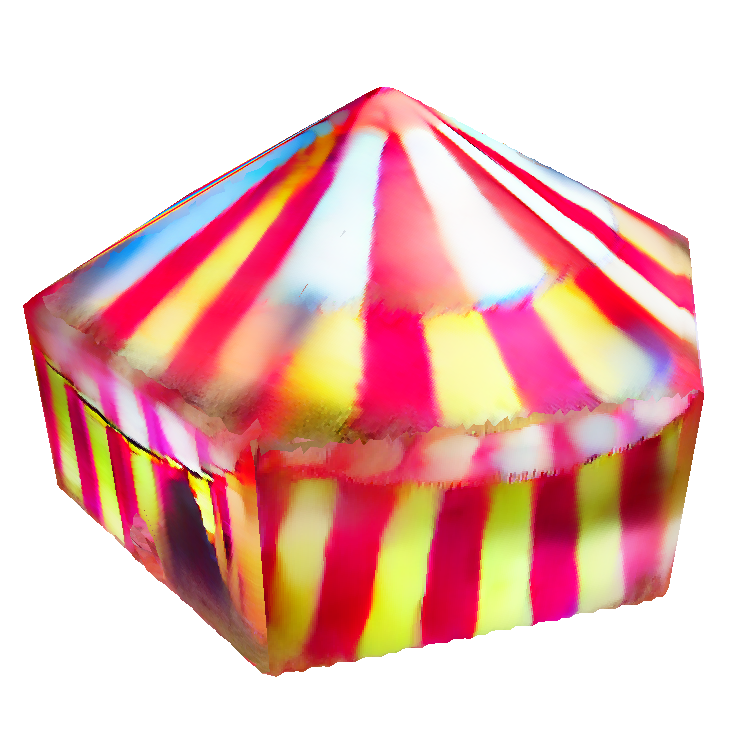} & 
        \includegraphics[width=0.2\linewidth]{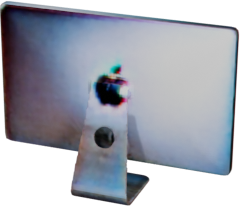} \\

        {\raisebox{0.1in}{\rotatebox{90}{TEXTure}}} &
        \includegraphics[width=0.2\linewidth]{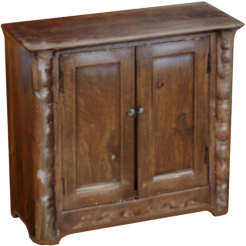} &
        \includegraphics[width=0.2\linewidth]{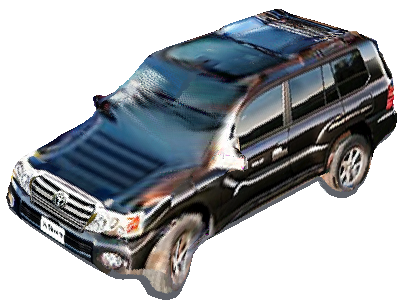} &
        \includegraphics[width=0.2\linewidth]{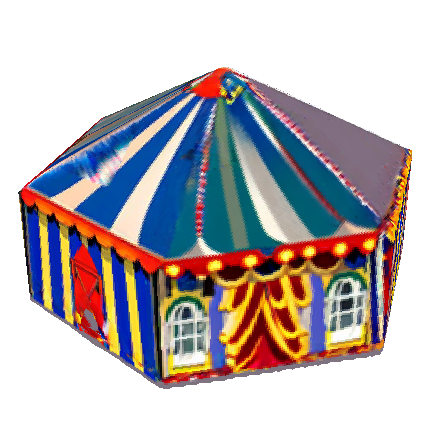} &
        \includegraphics[width=0.2\linewidth]{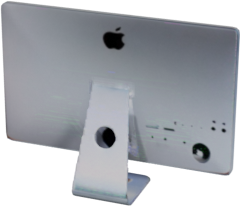} \\

        &
        \begin{tabular}{c} ``A wooden \\ cabinet'' \end{tabular}&
        \begin{tabular}{c} ``A toyota \\ land cruiser'' \end{tabular}&
        \begin{tabular}{c} ``A circus \\ tent'' \end{tabular}&
        \begin{tabular}{c} ``A desktop \\ iMac'' \end{tabular}\\ 
        
    \end{tabular}}
    \end{minipage}%
    \caption{Additional qualitative comparison for text-guided texture generation. Best viewed zoomed in.}
\end{figure*}

%% file: figures/paint_more_results/fig.tex
\begin{figure*}
\centering
    \centering
    \setlength{\tabcolsep}{0pt}
    {\small

    \begin{tabular}{c c @{\hskip 2pt} c  @{\hskip 2pt} c}
        \includegraphics[height=0.28\linewidth,trim={14cm 9cm 14cm 7cm},clip]{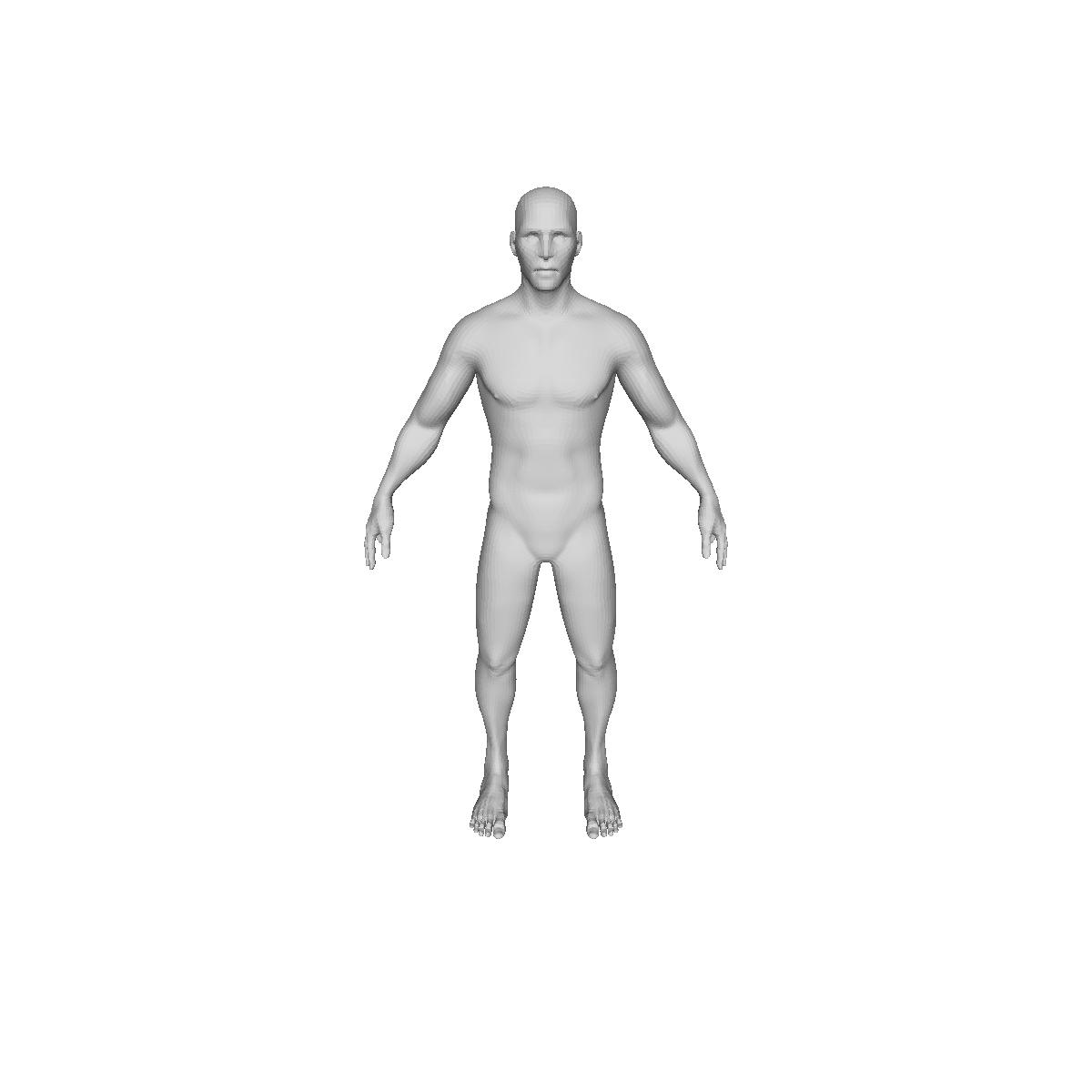} &
        \includegraphics[height=0.28\linewidth,trim={14cm 9cm 14cm 7cm},clip]{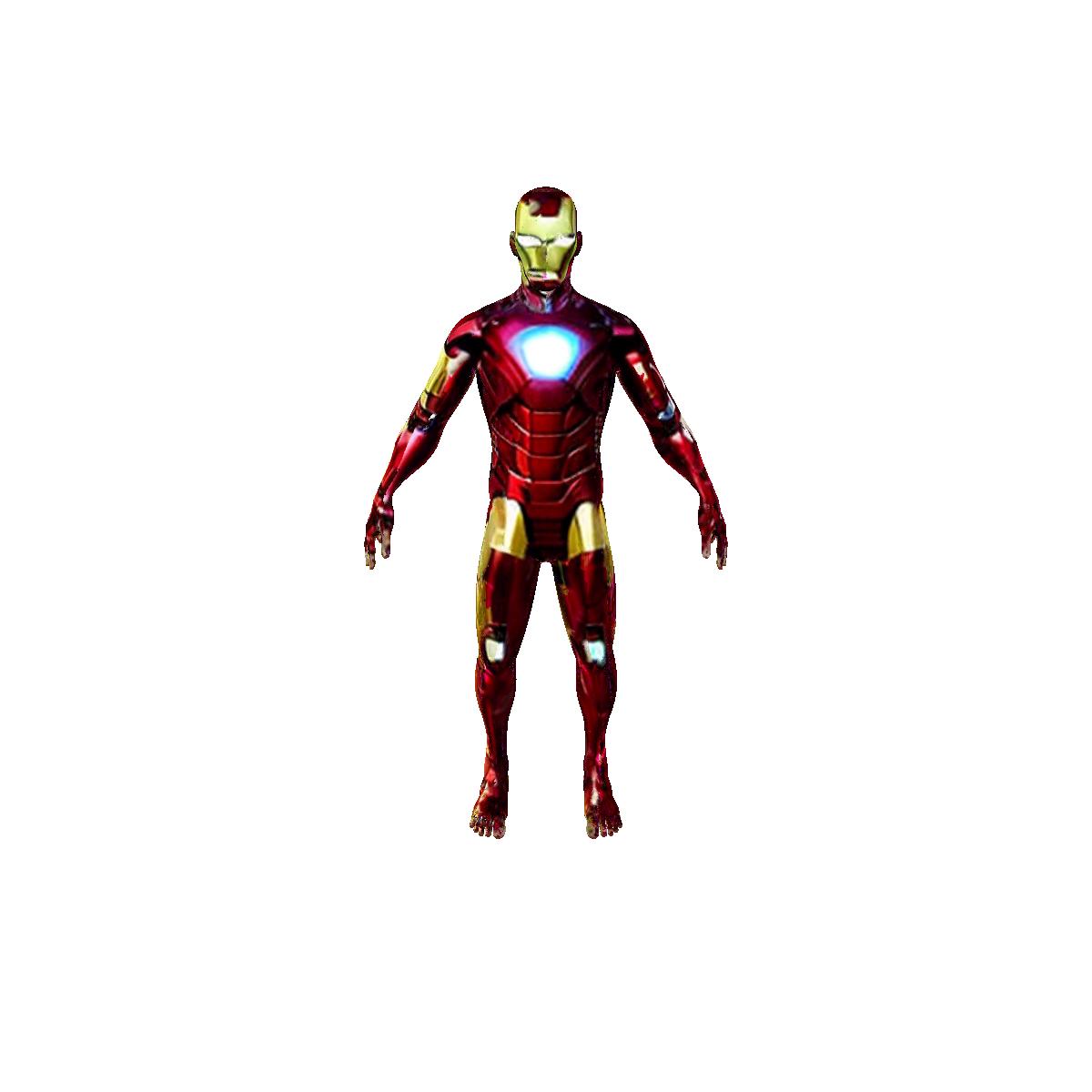} 
        \includegraphics[height=0.28\linewidth,trim={16cm 9cm 14cm 7cm},clip]{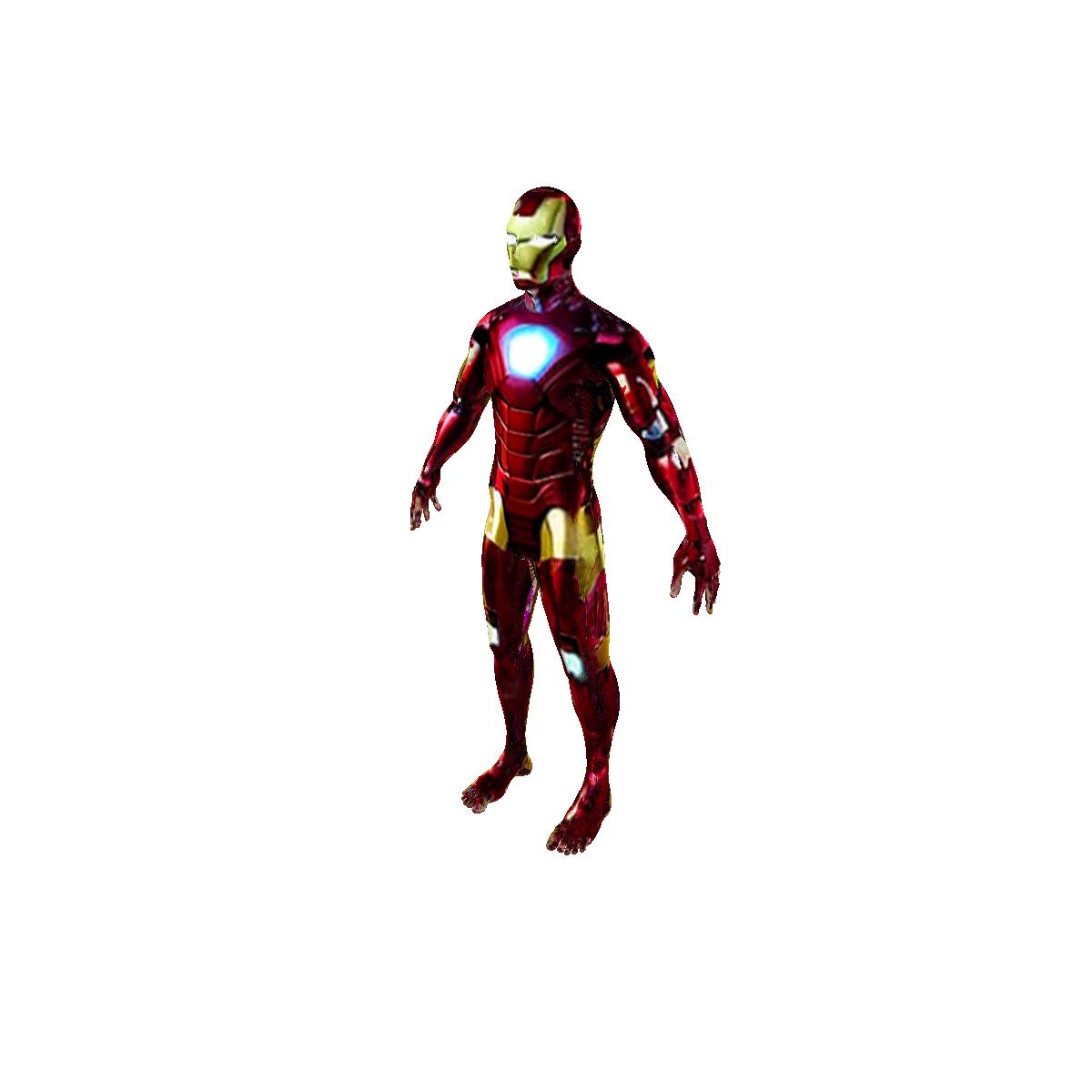}  &
        \includegraphics[height=0.28\linewidth,trim={14cm 9cm 14cm 7cm},clip]{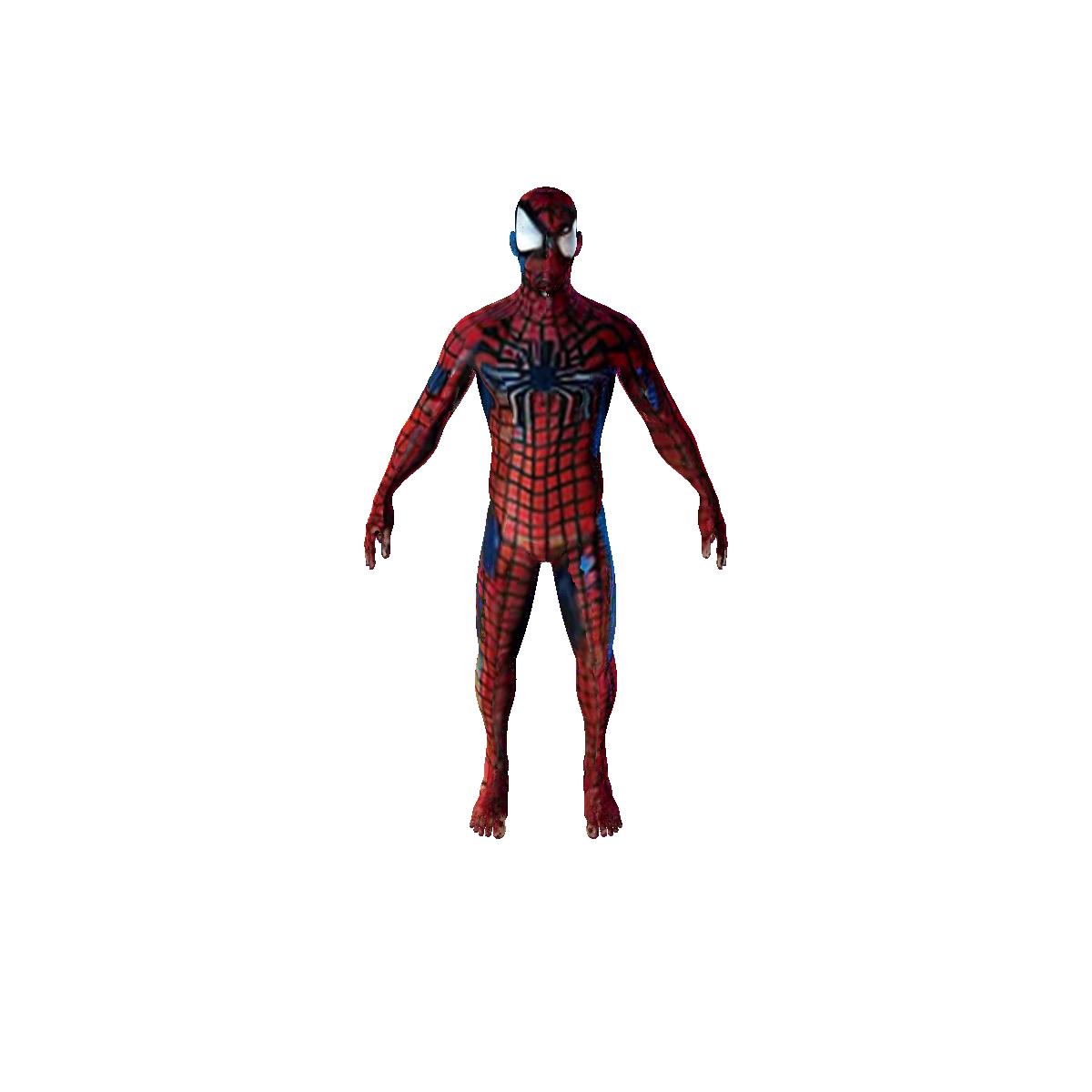} 
        \includegraphics[height=0.28\linewidth,trim={16cm 9cm 14cm 7cm},clip]{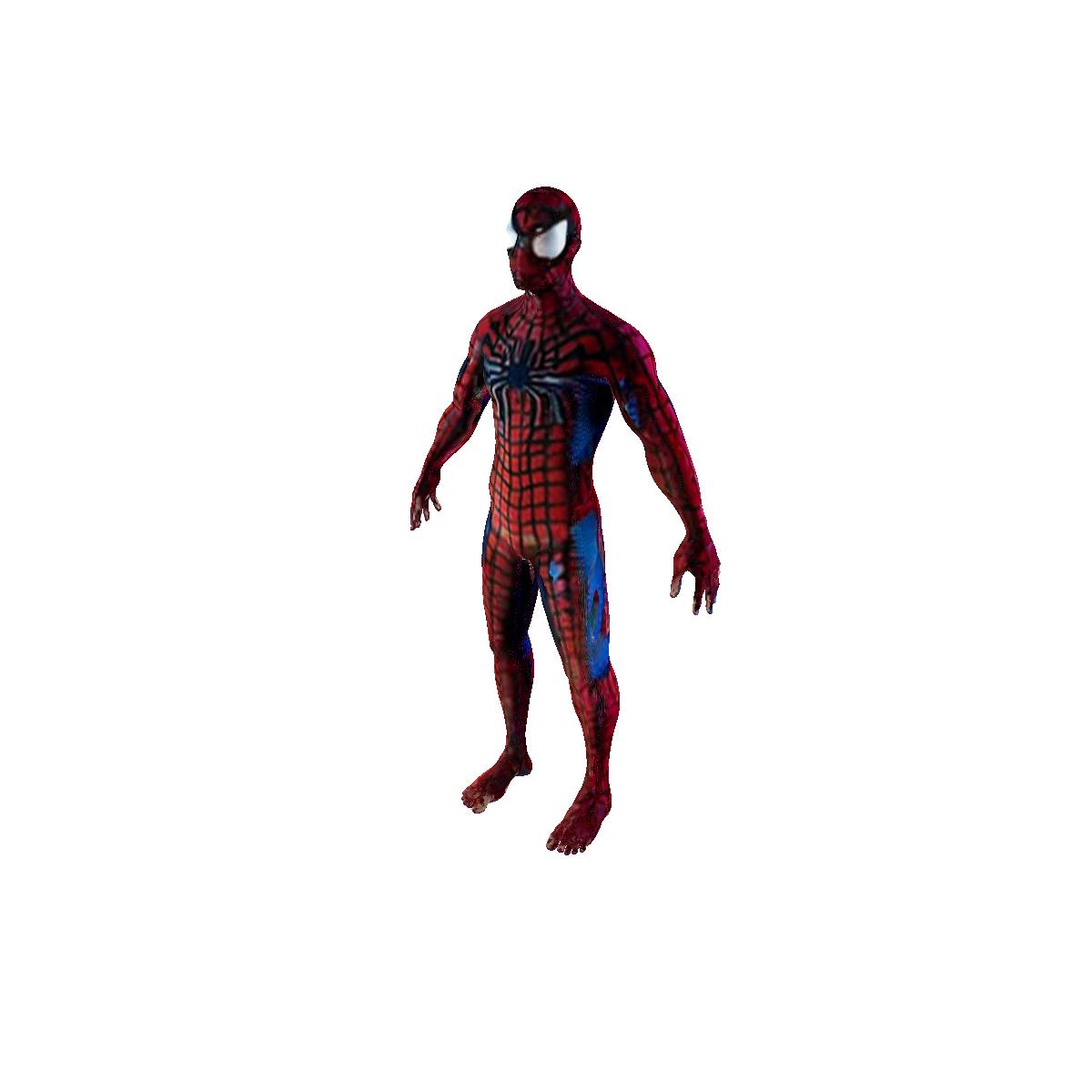} &
        \includegraphics[height=0.28\linewidth,trim={14cm 9cm 14cm 7cm},clip]{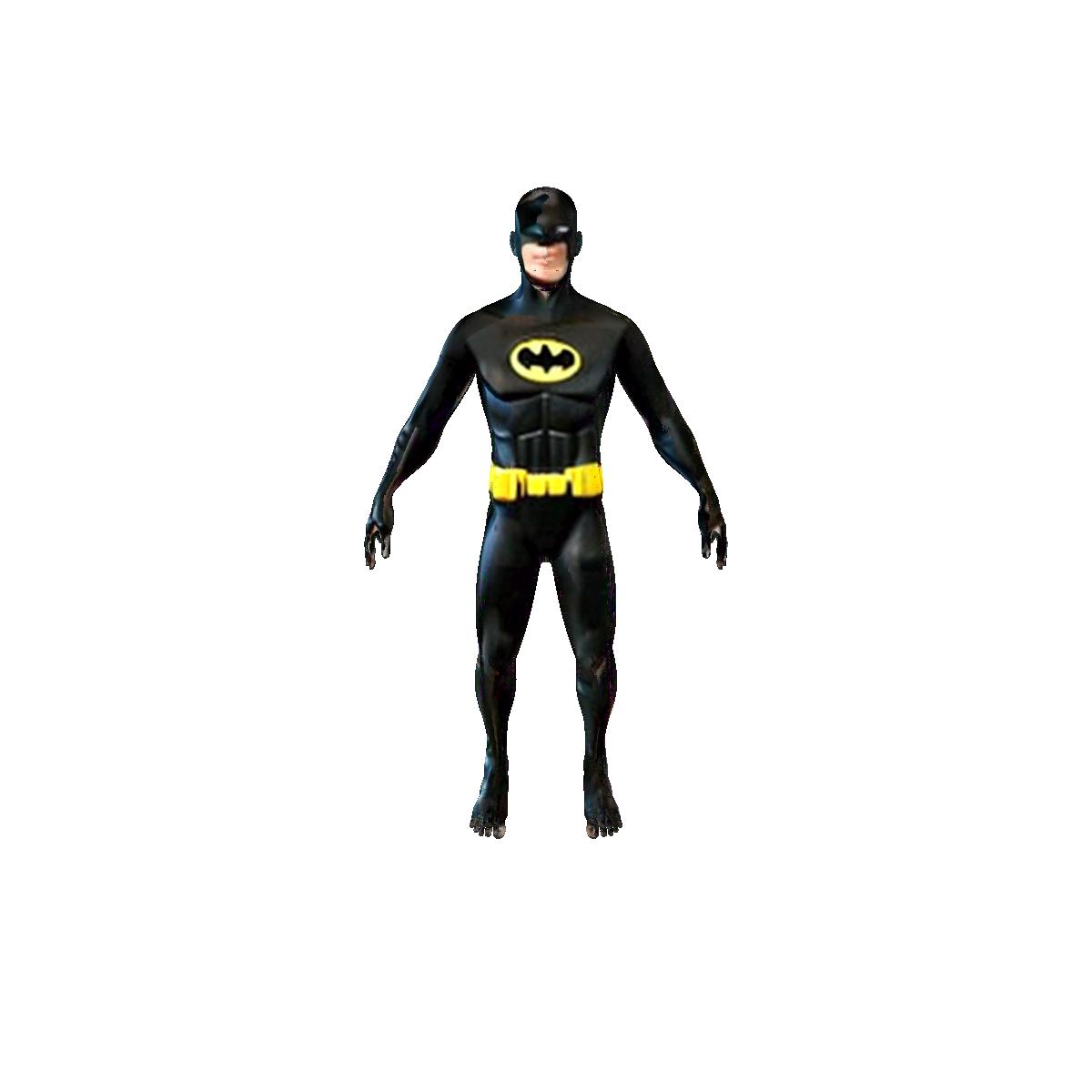} 
        \includegraphics[height=0.28\linewidth,trim={16cm 9cm 14cm 7cm},clip]{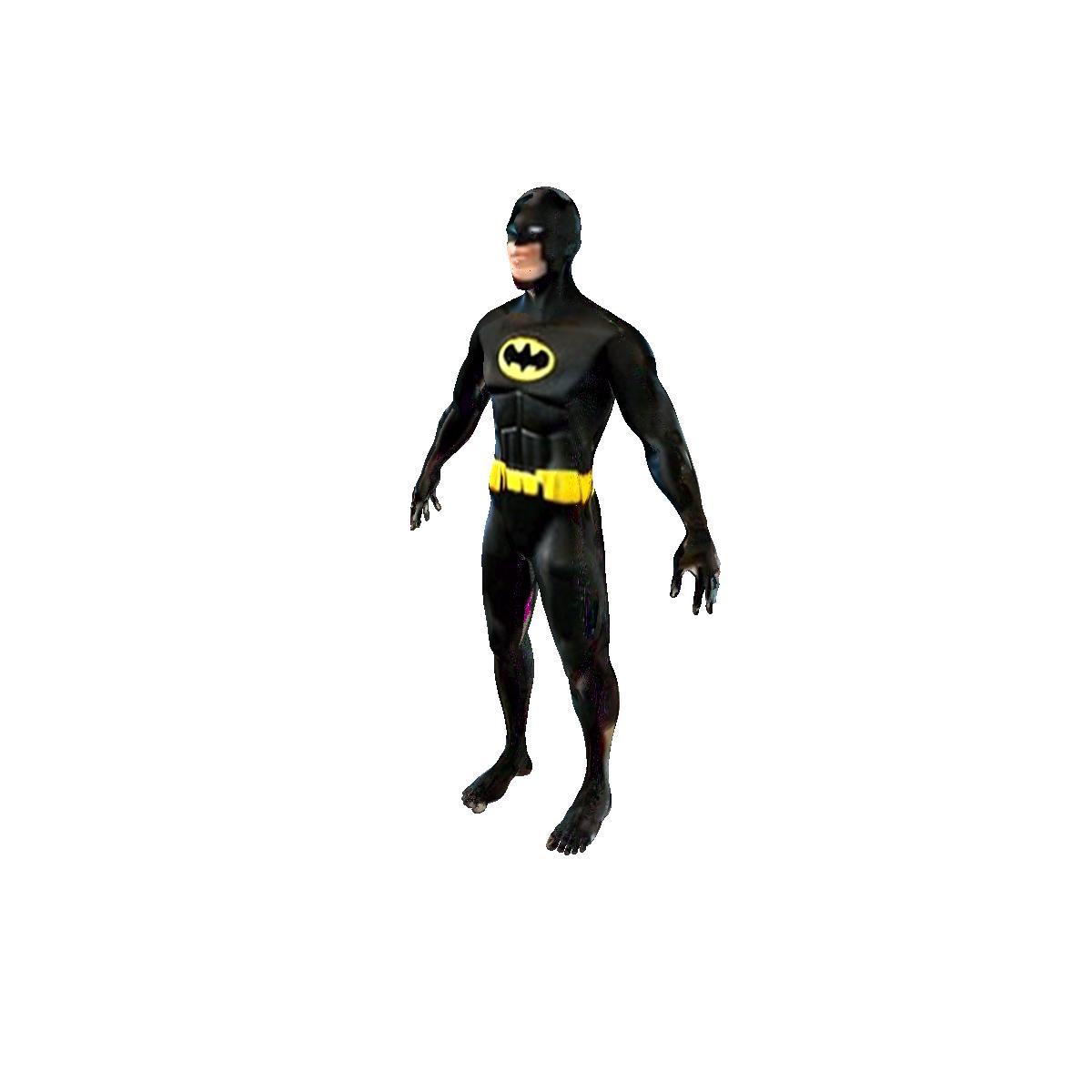}
        \\
         Input Mesh & ``A photo of ironman'' & ``A photo of spiderman''  & ``A photo of batman''
    \\
    \end{tabular}

    }
    \vspace{-0.1cm}
    \caption{Additional texturing results achieved with TEXTure. Our method generates a high-quality texture for a collection of prompts and geometries.}
    \label{fig:even_more_paint_results}
\end{figure*}

%% file: figures/images2mesh/fig_extra.tex
\begin{figure*}
    \centering
    \setlength{\tabcolsep}{0pt}
    {\small
    \begin{tabular}{c c c c c c c}
        \\ \\ 
        \includegraphics[height=0.10\linewidth]{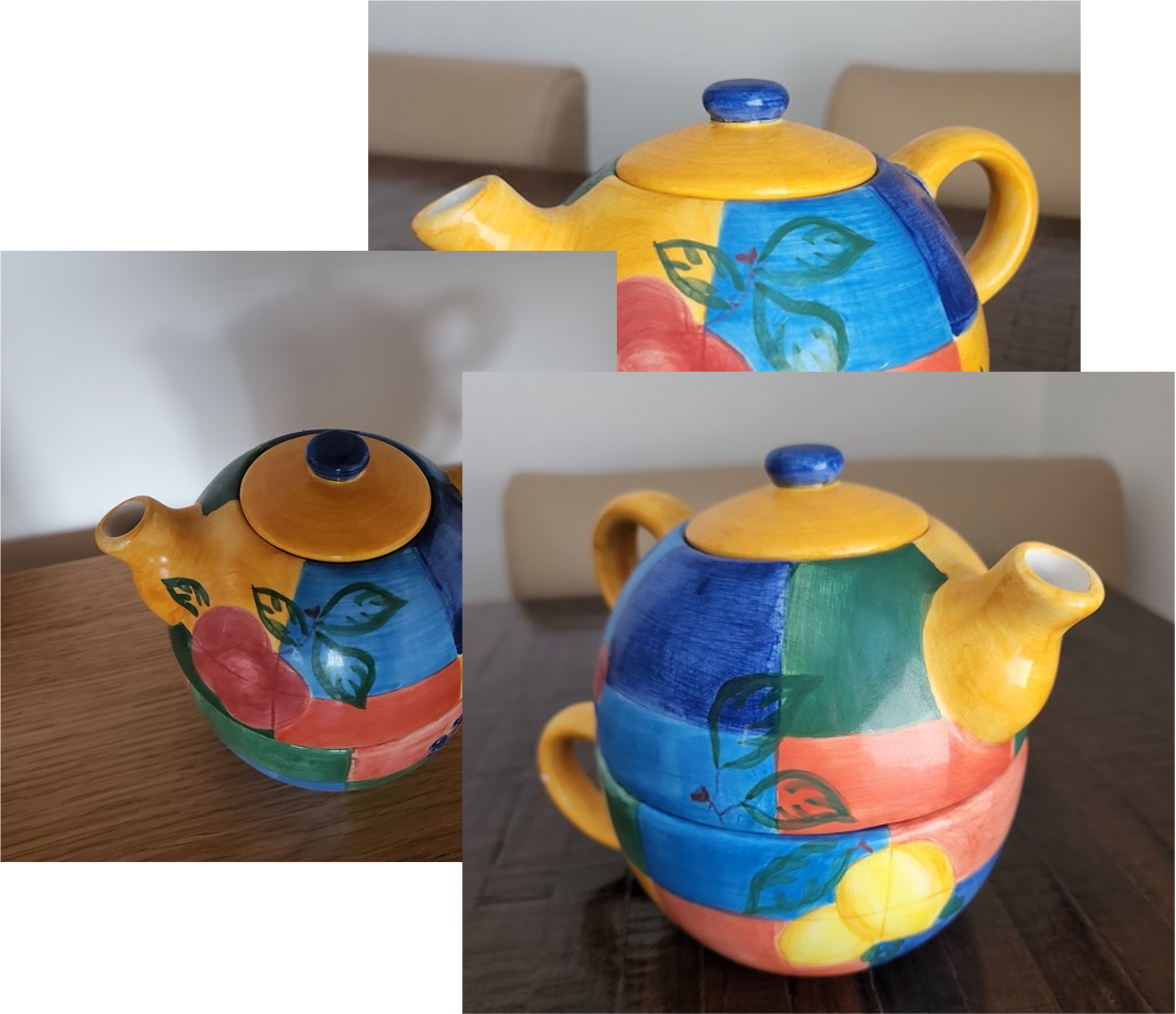} &
        \includegraphics[height=0.10\linewidth]{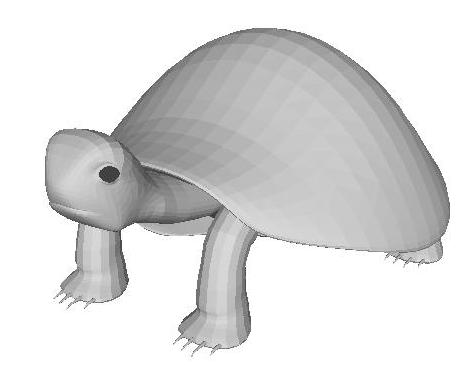} &
        \includegraphics[height=0.10\linewidth]{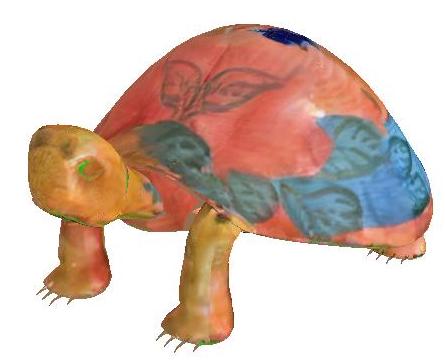} &
        \includegraphics[height=0.10\linewidth]{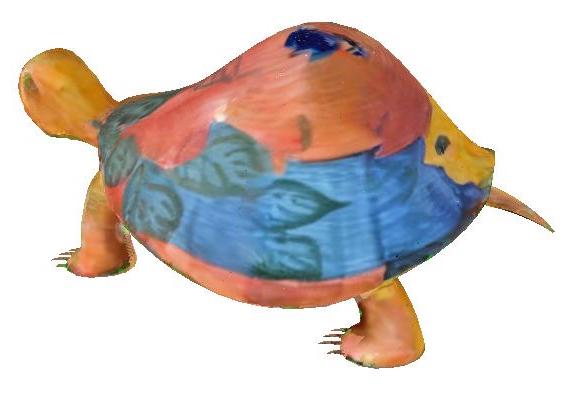} &
        \includegraphics[height=0.10\linewidth]{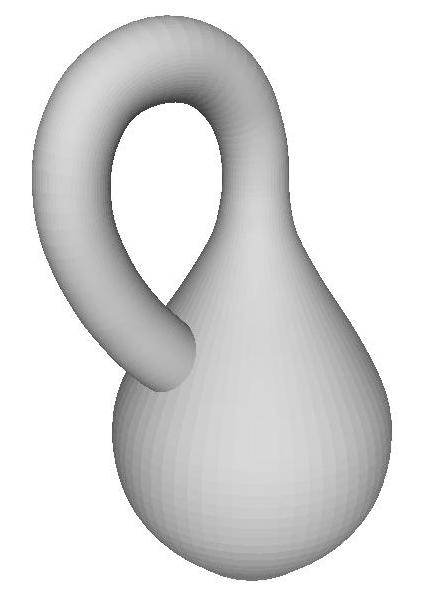} &
        \includegraphics[height=0.10\linewidth]{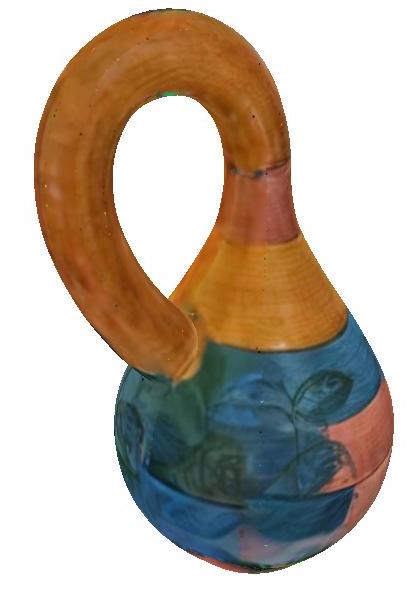} &
        \includegraphics[height=0.10\linewidth]{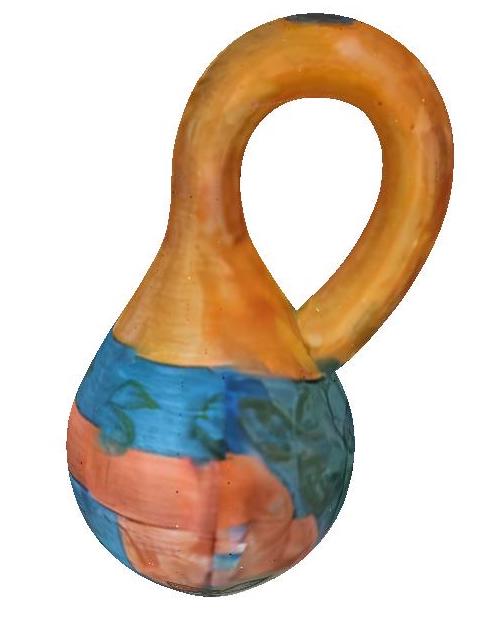} \\

        \includegraphics[height=0.10\linewidth]{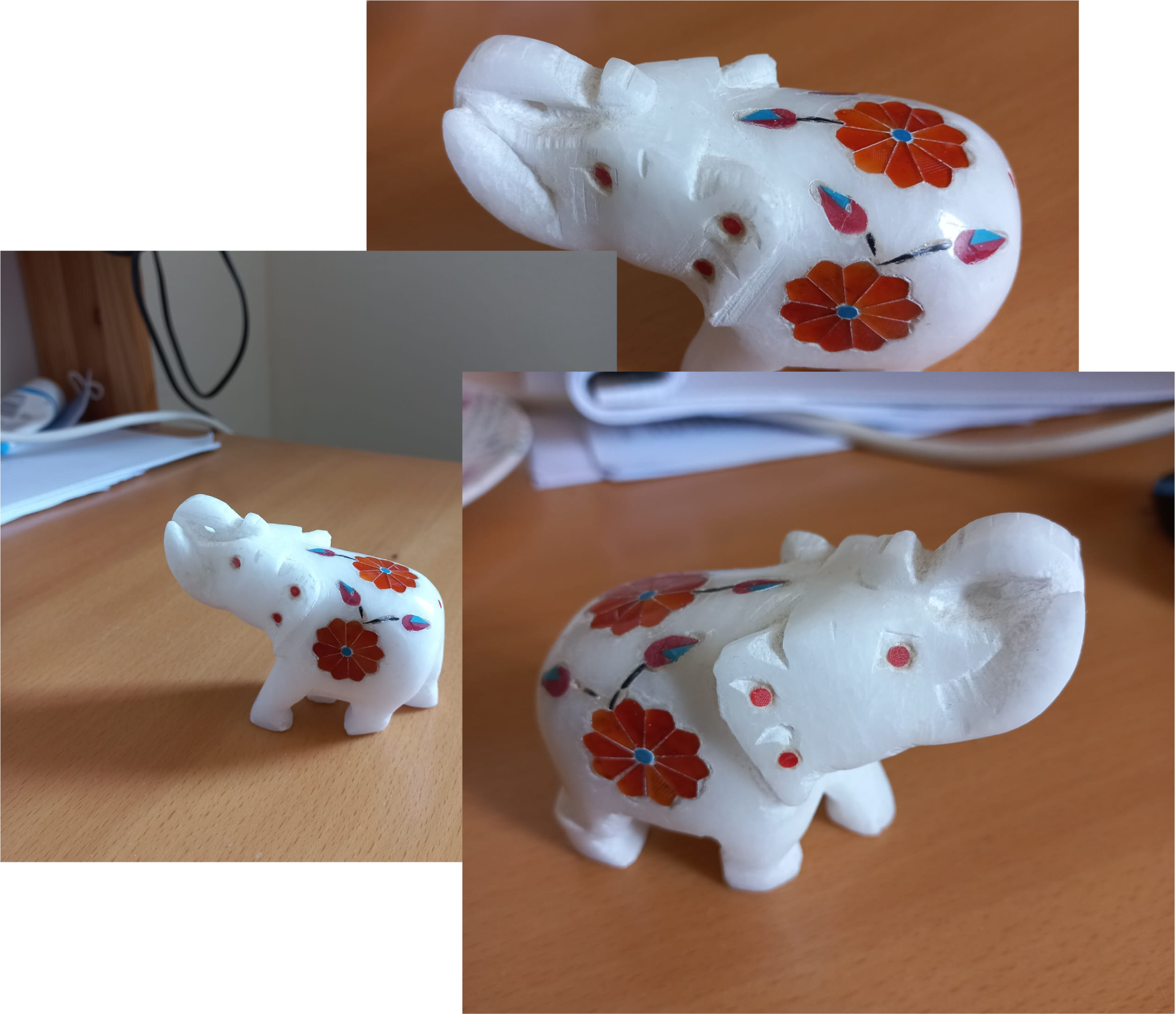} &
        \includegraphics[height=0.10\linewidth]{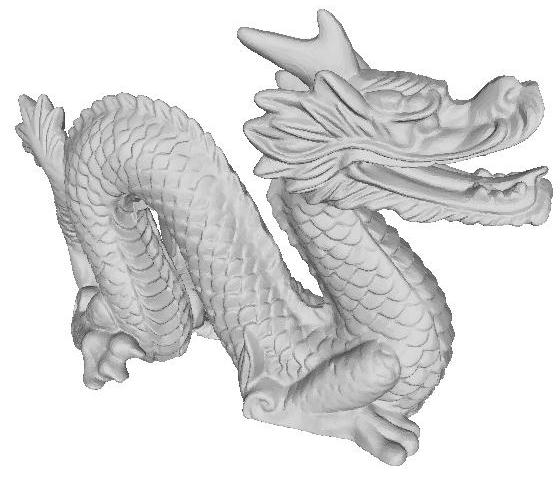} &
        \includegraphics[height=0.10\linewidth]{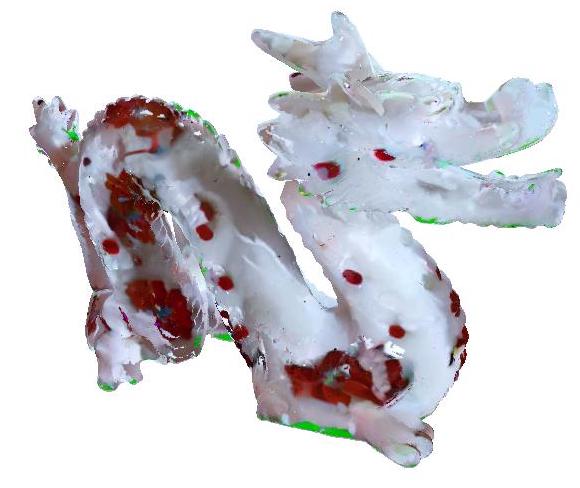} &
        \includegraphics[height=0.10\linewidth]{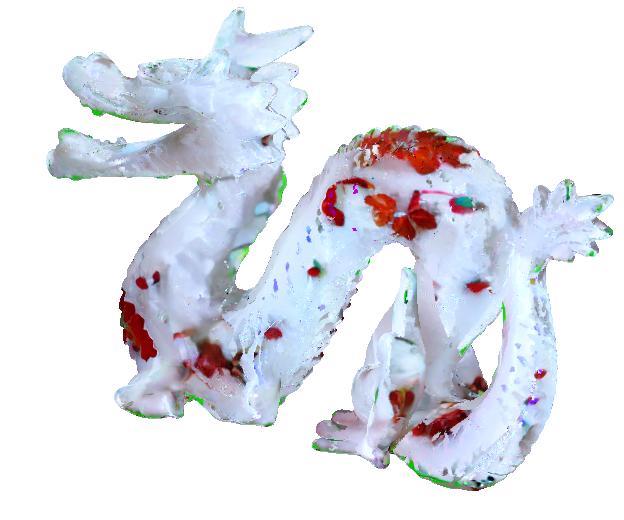} &
        \includegraphics[height=0.10\linewidth]{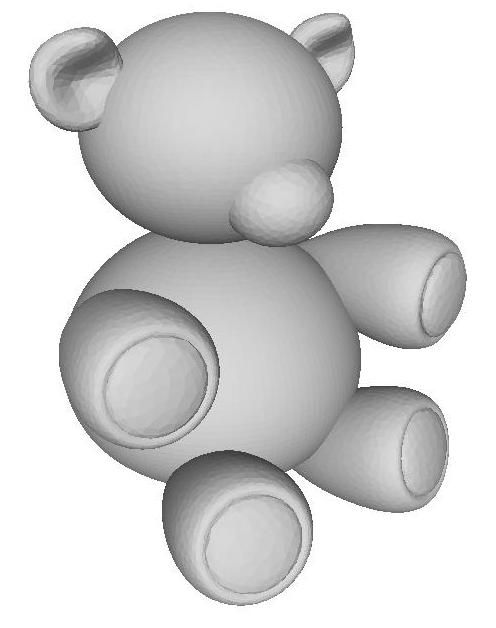} &
        \includegraphics[height=0.10\linewidth]{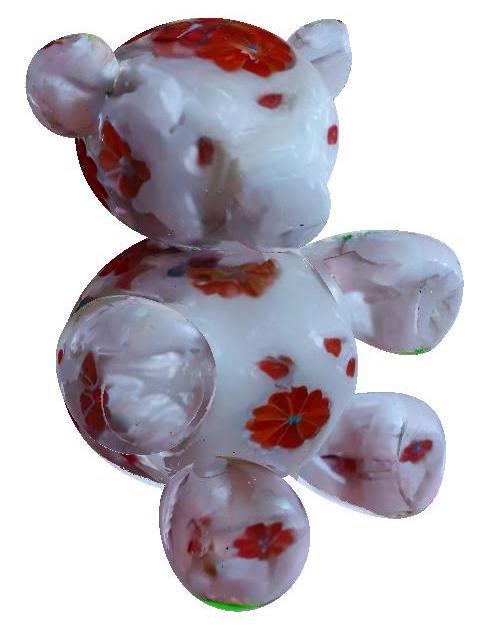} &
        \includegraphics[height=0.10\linewidth]{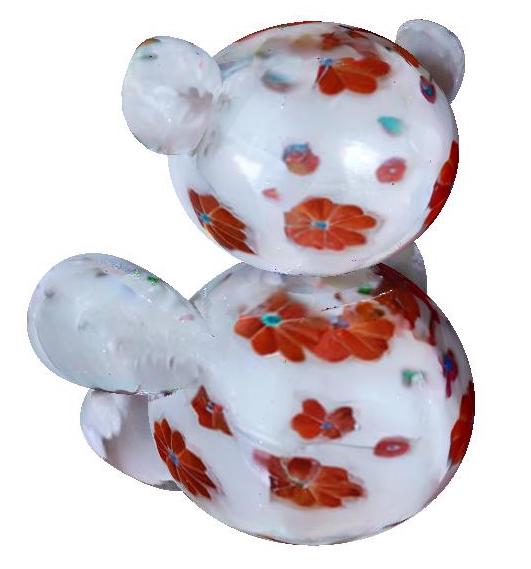} \\

        \includegraphics[height=0.10\linewidth]{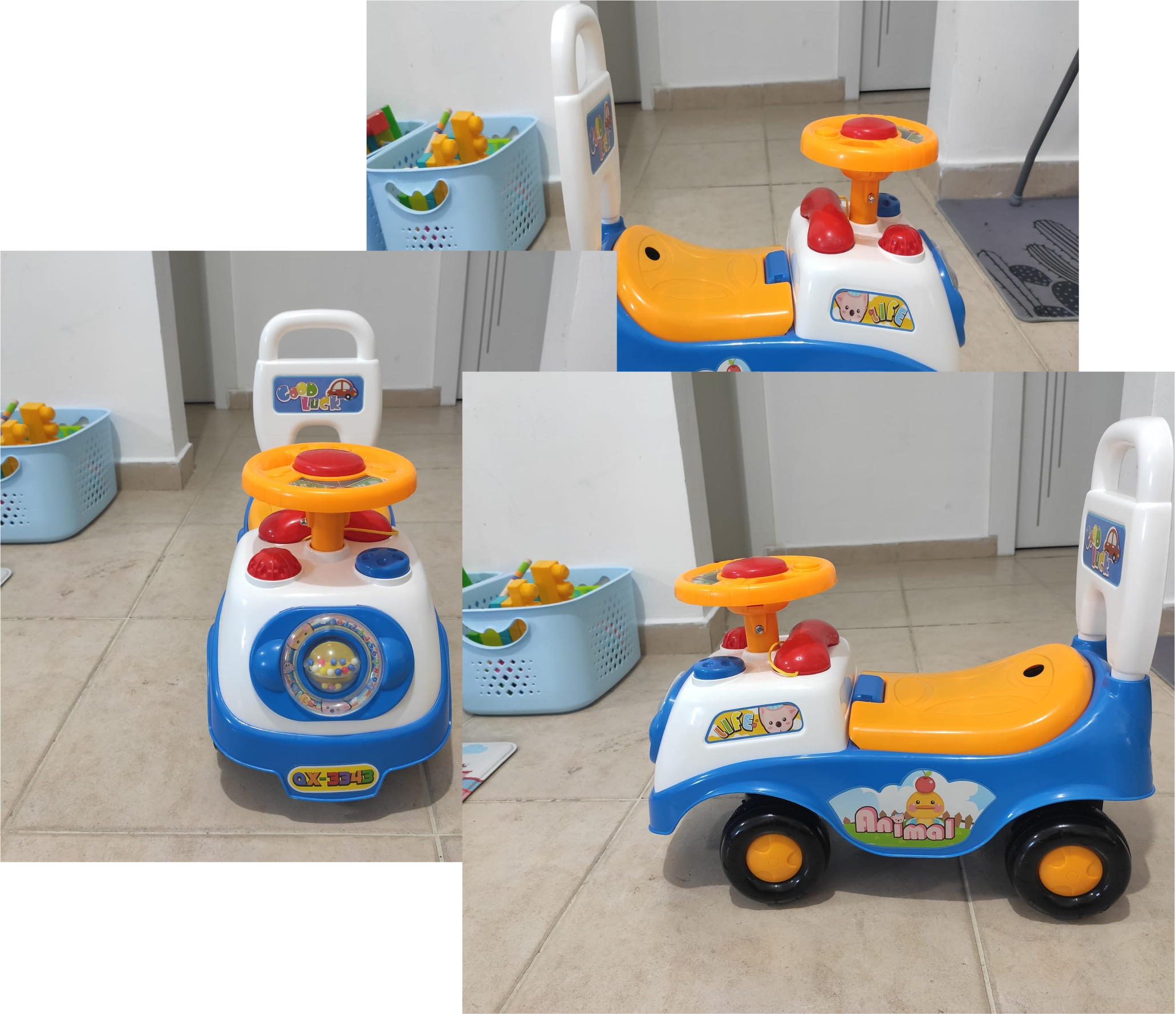} &
        \includegraphics[height=0.10\linewidth]{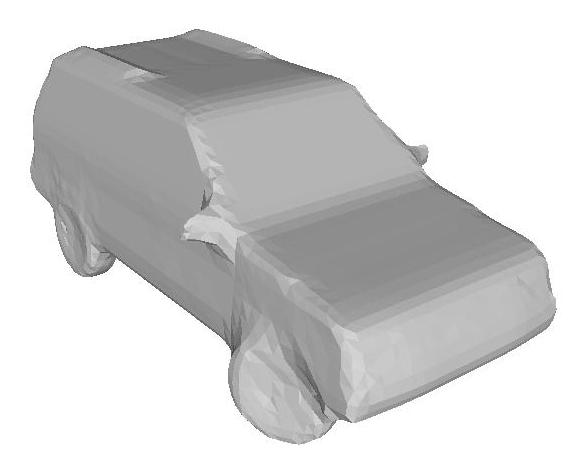} &
        \includegraphics[height=0.10\linewidth]{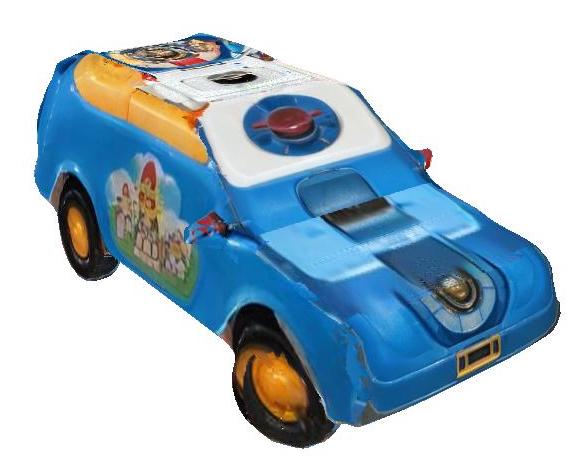} &
        \includegraphics[height=0.10\linewidth]{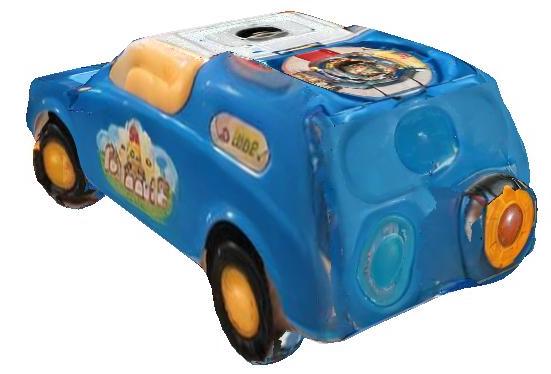} &
        \includegraphics[height=0.10\linewidth]{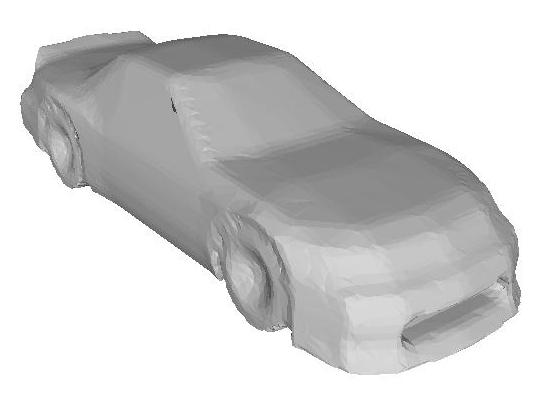} &
        \includegraphics[height=0.10\linewidth]{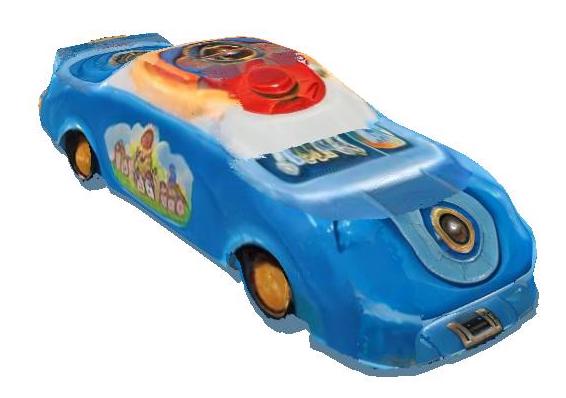} &
        \includegraphics[height=0.10\linewidth]{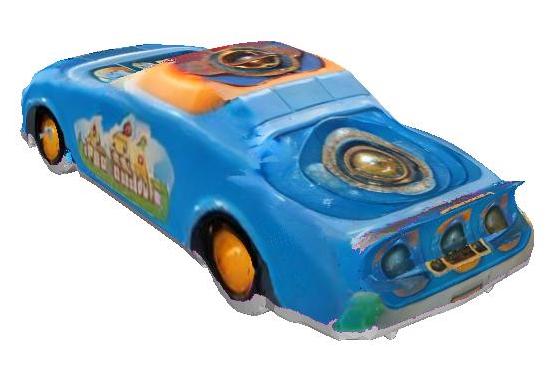} \\

        Images Set & Input Mesh & \multicolumn{2}{c}{Textured Meshes} & Input Mesh & \multicolumn{2}{c}{Textured Meshes}
        
    \end{tabular}
    }
    \vspace{-0.3cm}
    \caption{Token-based Texture Transfer from images. All meshes are textured with the \textbf{exact} prompt ``A photo of a $\langle S_* \rangle$ using the fine-tuned diffusion model.}
    \label{fig:images2mesh_extra}
    \vspace{-0.2cm}
\end{figure*}

%% file: figures/edits/fig.tex
\begin{figure*}
    \centering
    {\small
    \begin{tabular}{c c c c c c c}
    \\ \\
        \includegraphics[width=0.105\linewidth,trim={11cm 11cm 10cm 8.5cm},clip]{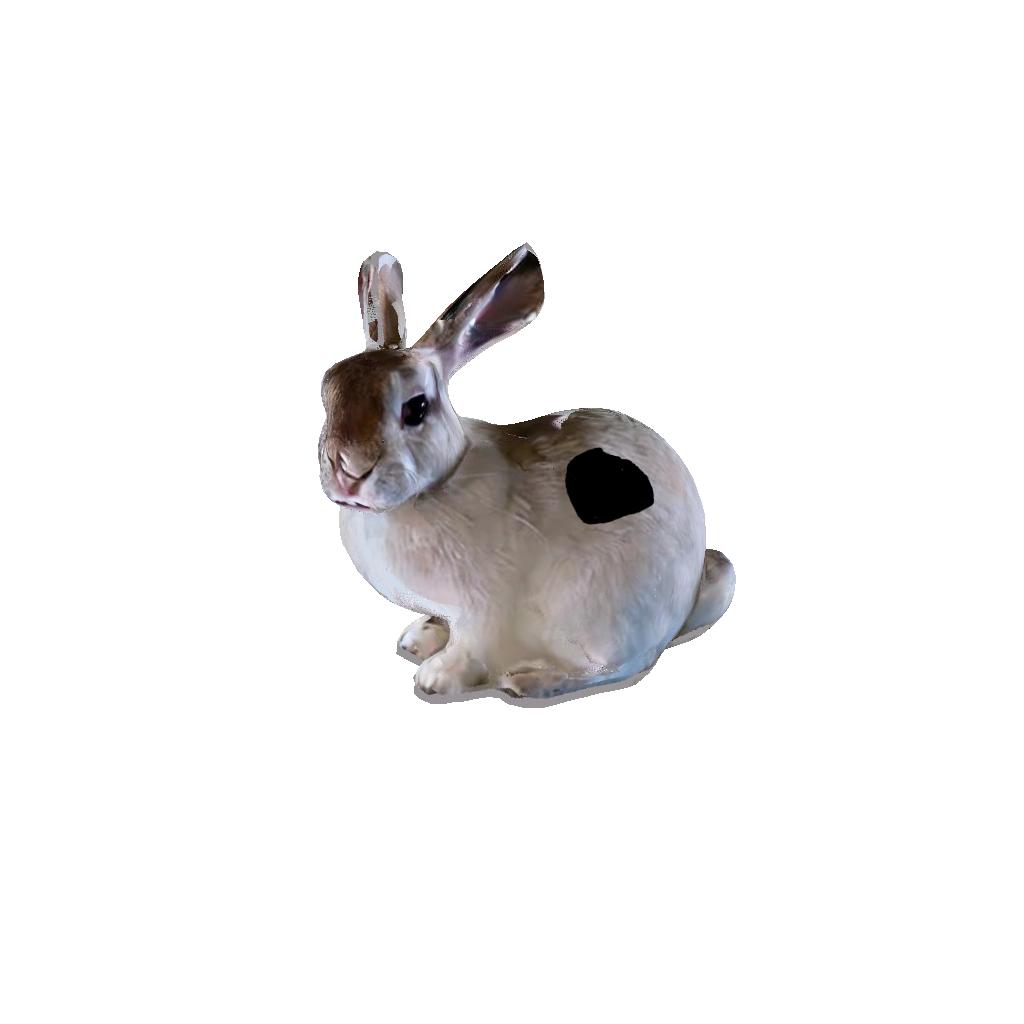} &
        \includegraphics[width=0.105\linewidth,trim={11cm 11cm 10cm 8.5cm},clip]{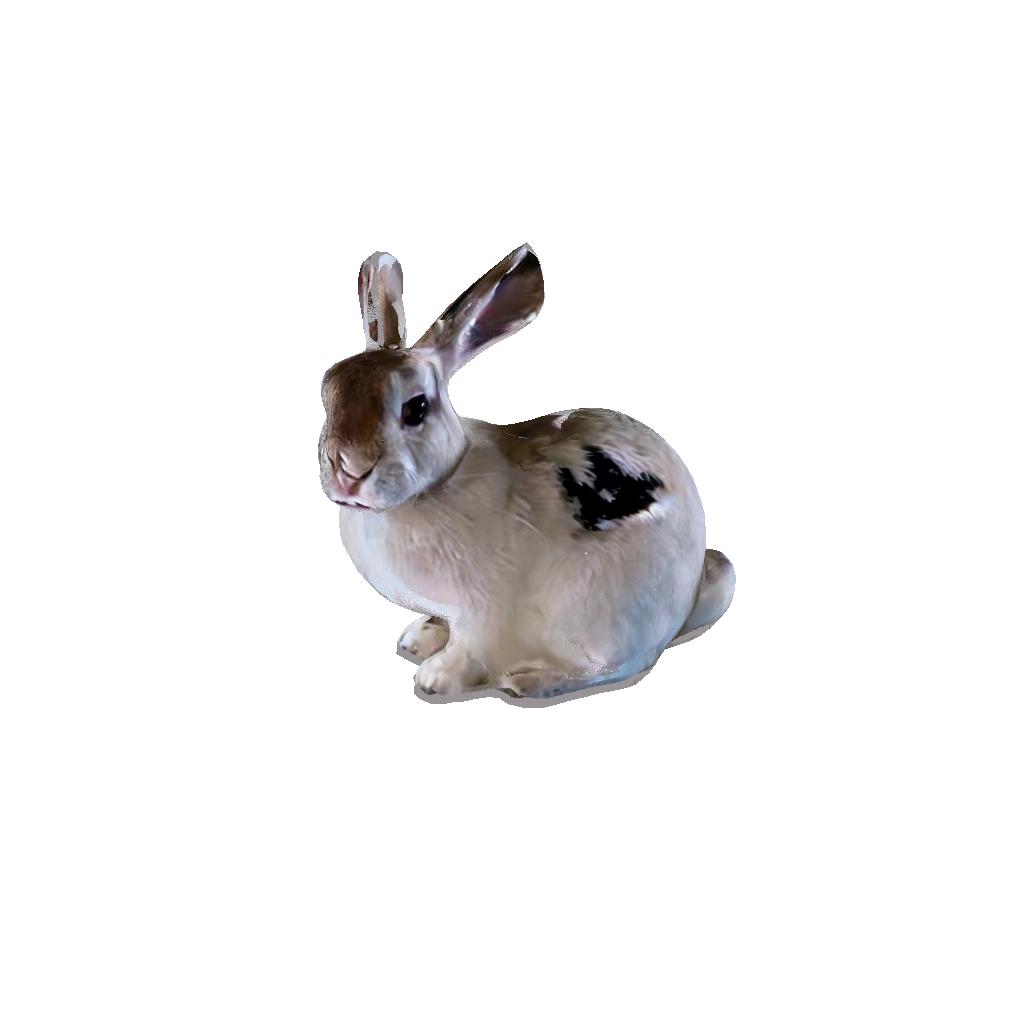} &
        \includegraphics[width=0.105\linewidth,trim={11cm 11cm 10cm 8.5cm},clip]{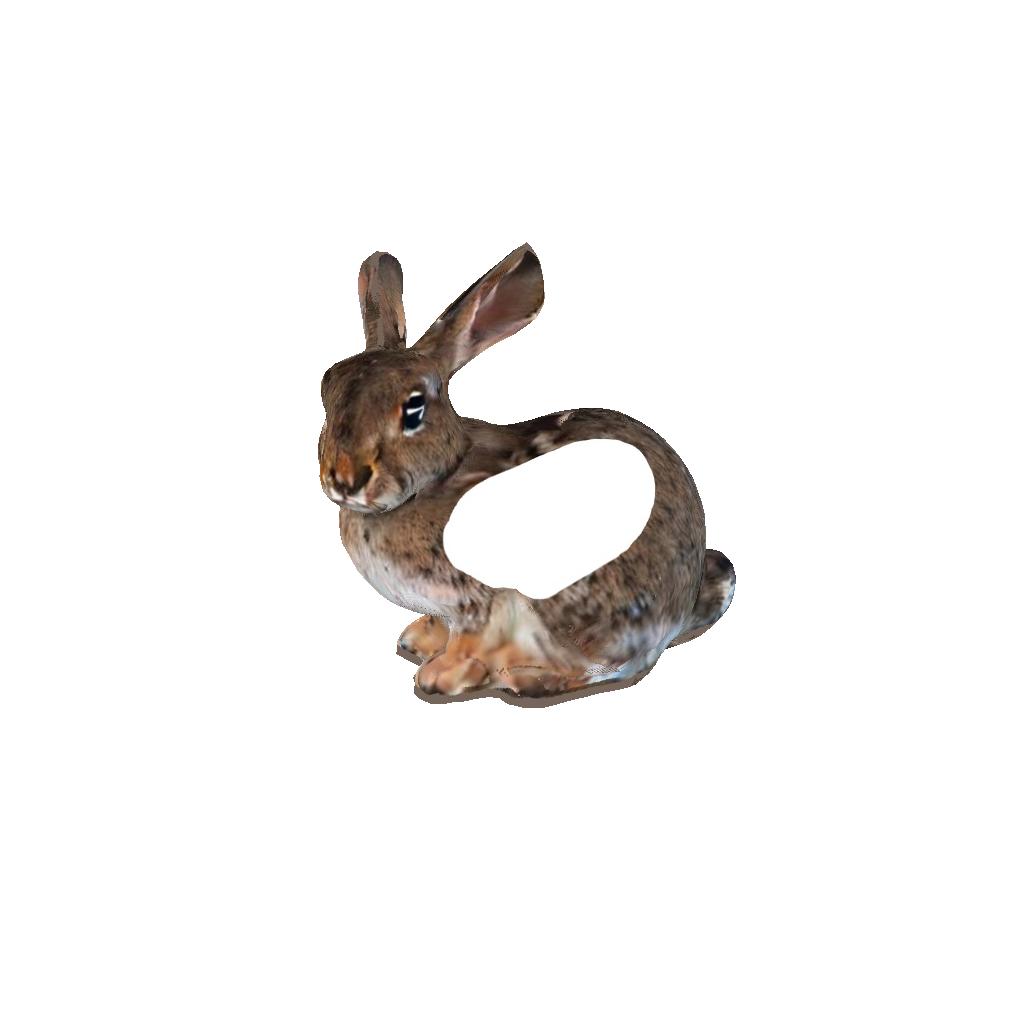} &
        \includegraphics[width=0.105\linewidth,trim={11cm 11cm 10cm 8.5cm},clip]{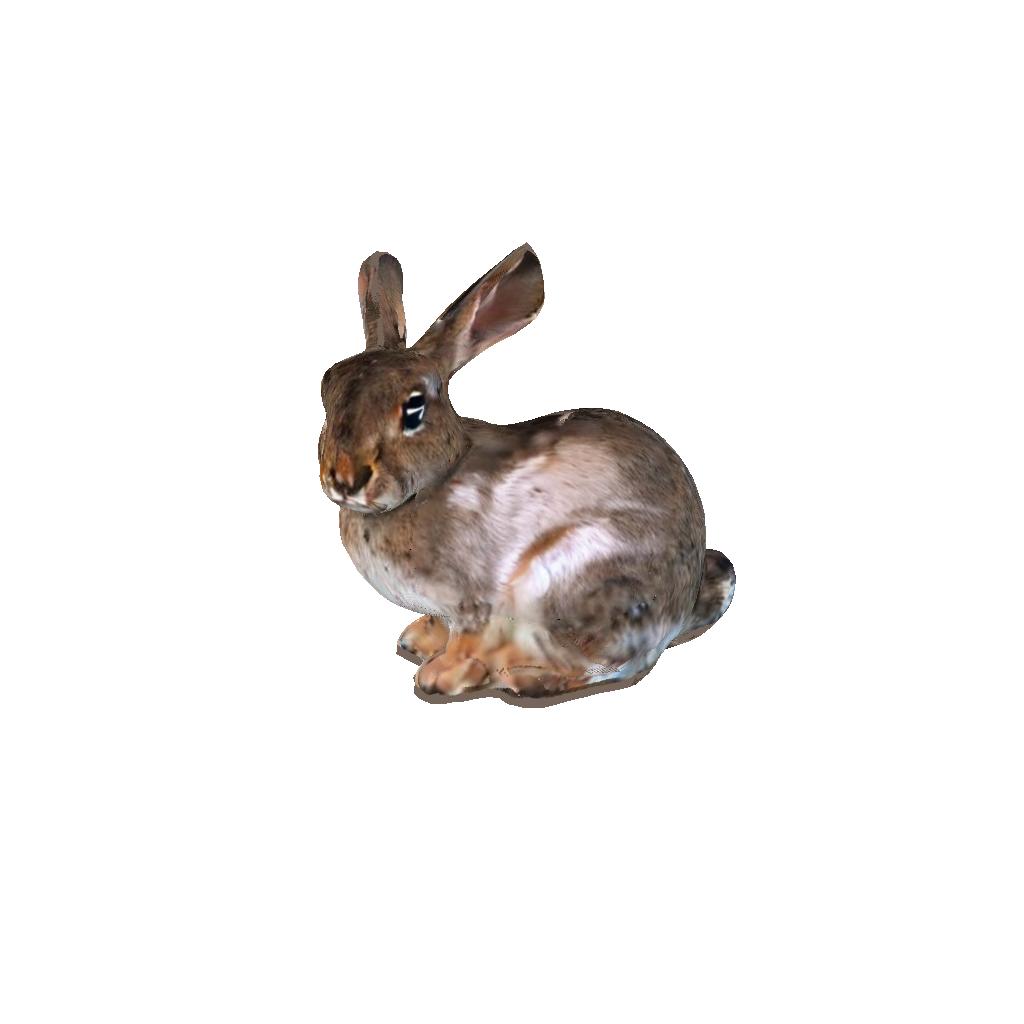} &
        \includegraphics[width=0.105\linewidth]{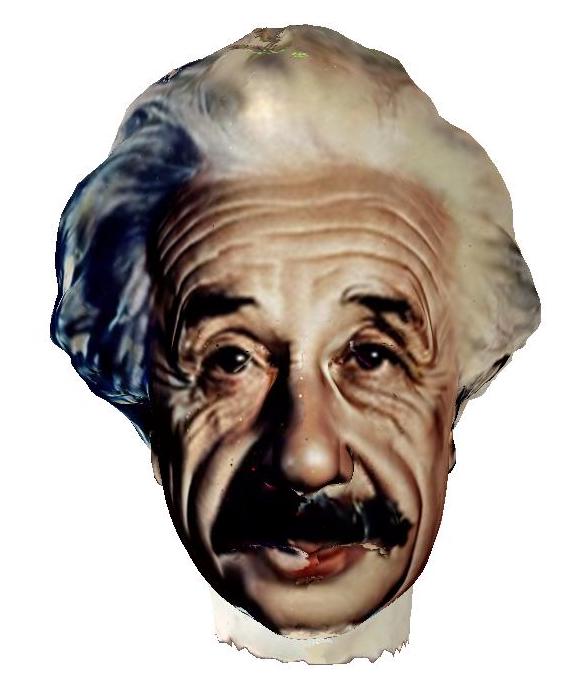} &
        \includegraphics[width=0.105\linewidth]{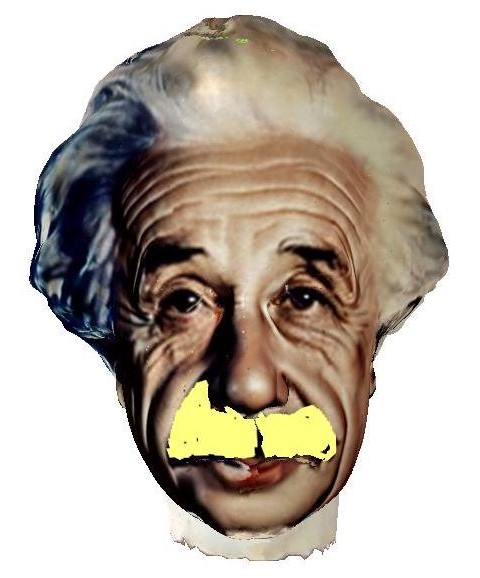} &
        \includegraphics[width=0.105\linewidth]{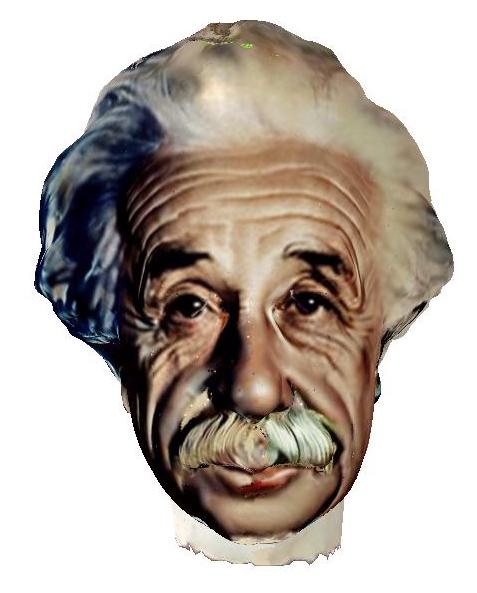} \\
        Input Edit & Result & Input Edit & Result & Input Texture & Input Edit & Result \\
    \end{tabular}
    \begin{tabular}{c c c c@{\hspace{0.2cm}} c c c c}
        \\
        \includegraphics[height=0.0925\linewidth,trim={9.5cm 12cm 10cm 10cm},clip]{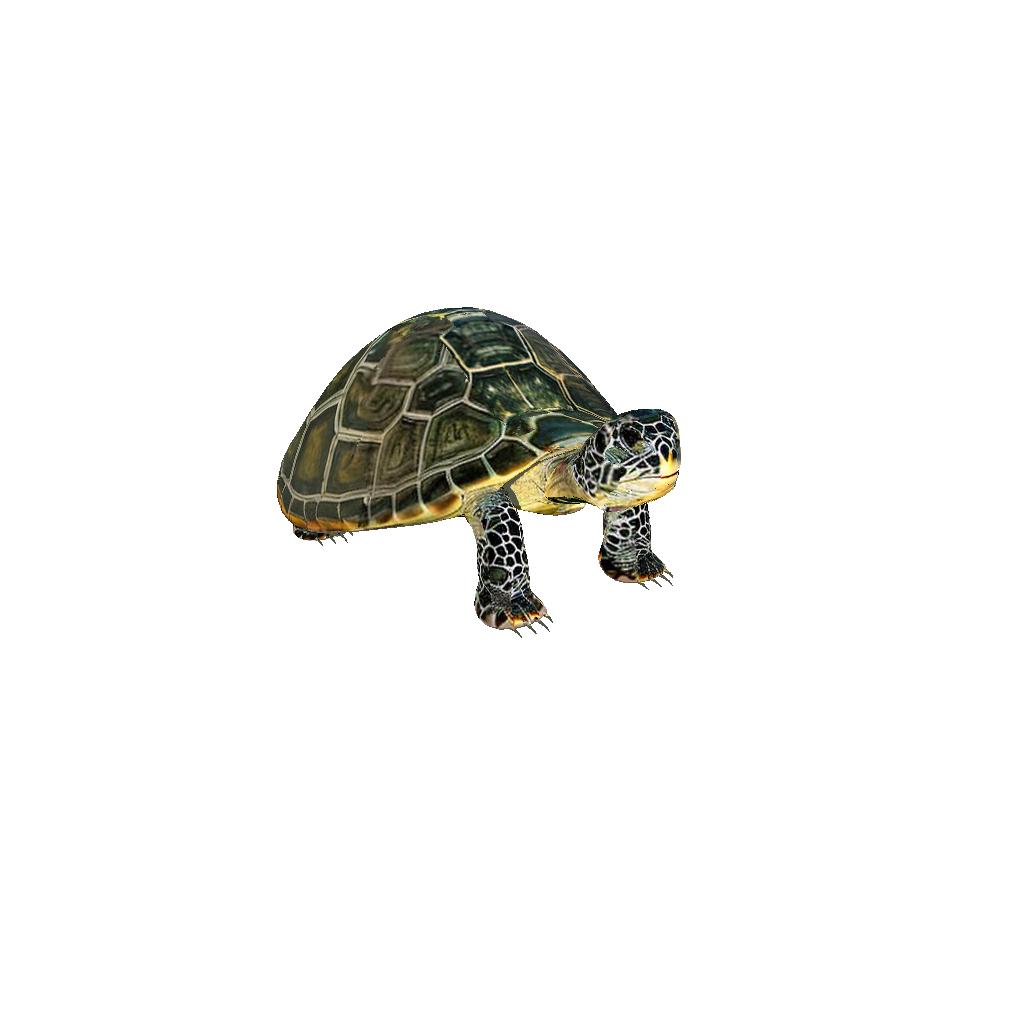} &
        \includegraphics[height=0.0925\linewidth,trim={9.5cm 12cm 10cm 10cm},clip]{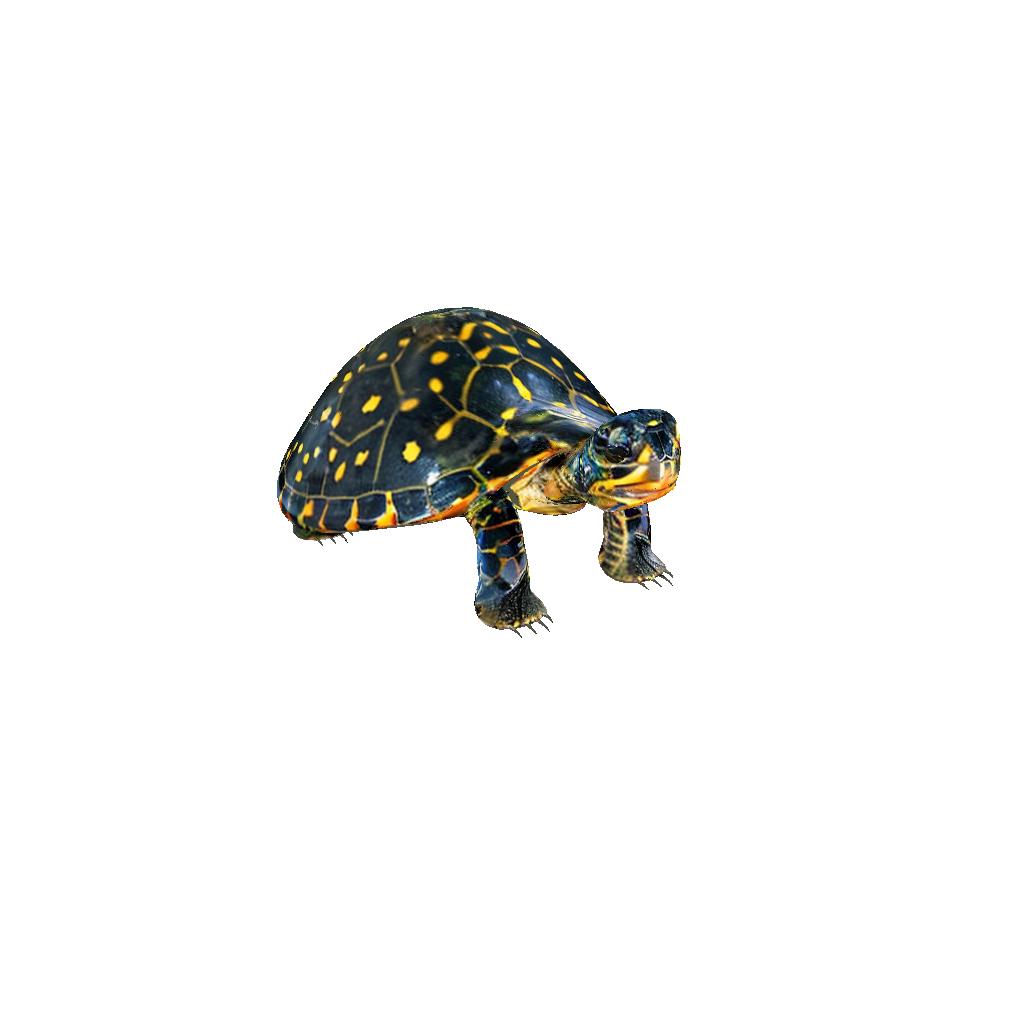} &
        \includegraphics[height=0.0925\linewidth,trim={9.5cm 12cm 10cm 10cm},clip]{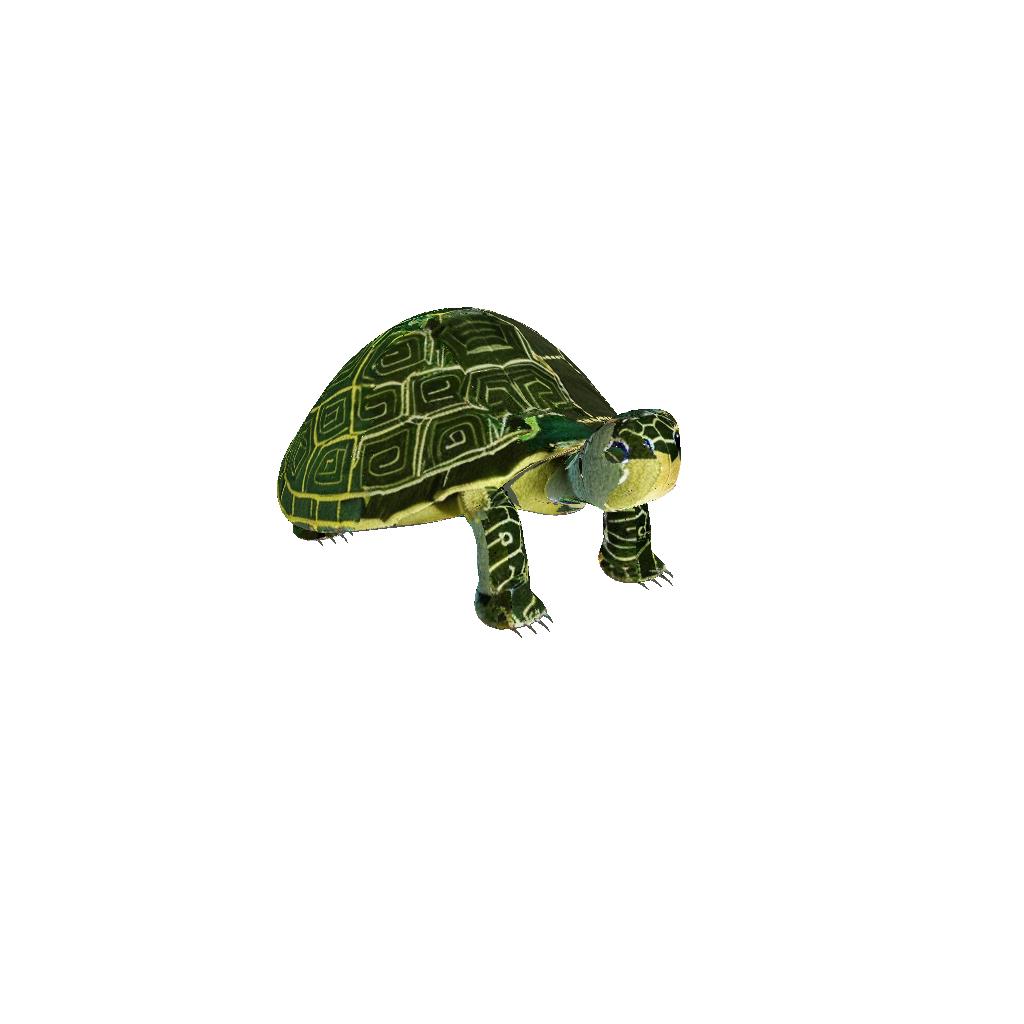} &
        \includegraphics[height=0.0925\linewidth,trim={9.5cm 12cm 10cm 10cm},clip]{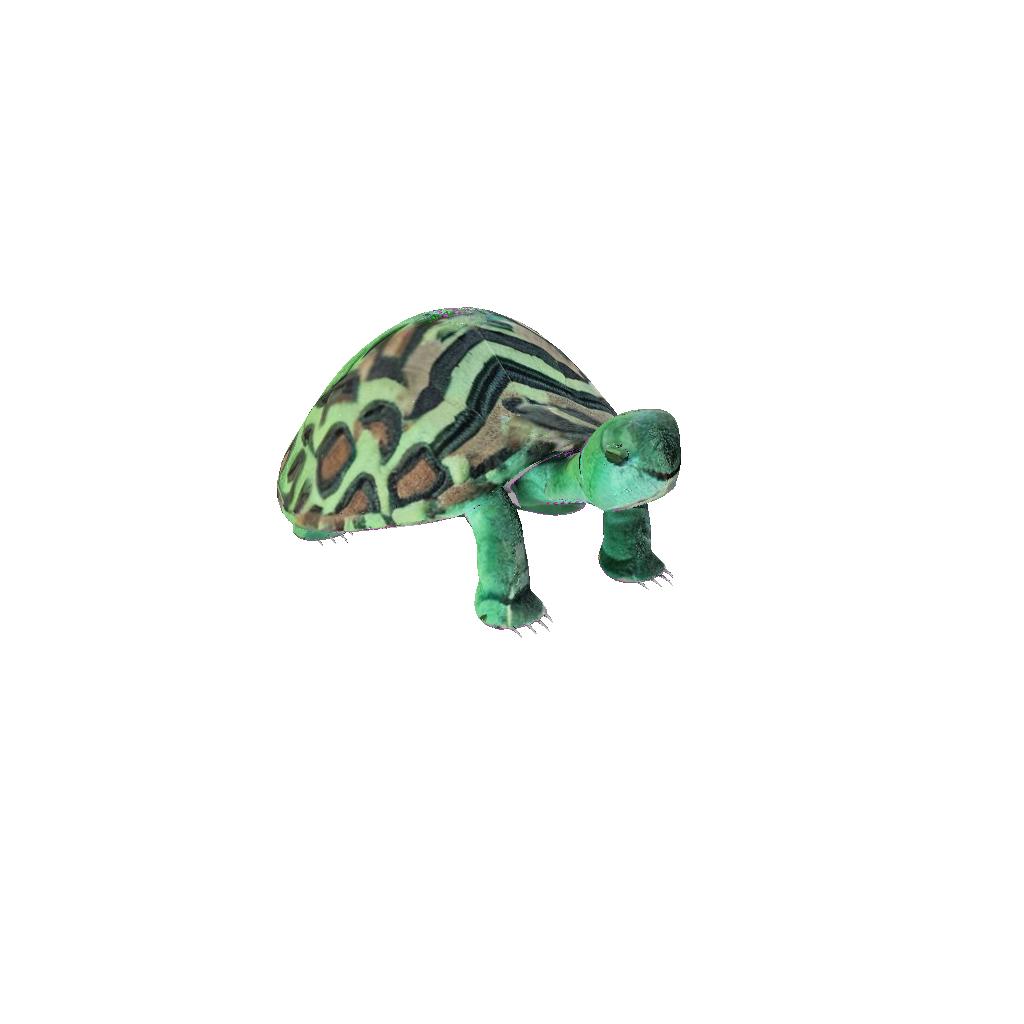} &
        \hspace{0.1cm}
        \includegraphics[height=0.0925\linewidth]{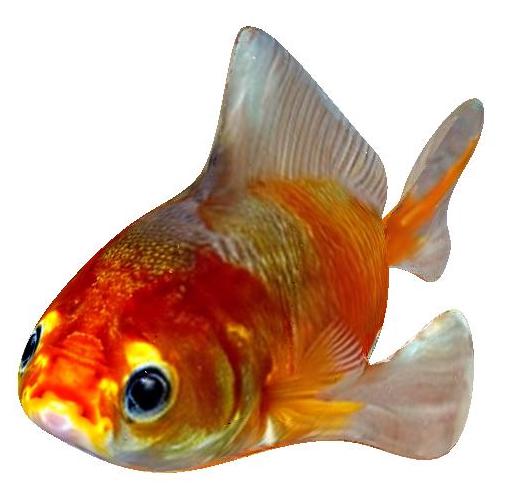} &
        \includegraphics[height=0.0925\linewidth]{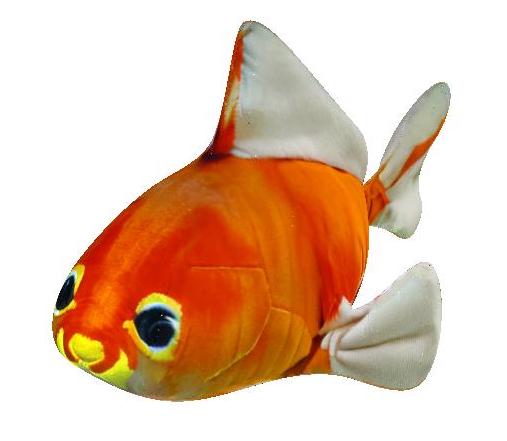} &
        \includegraphics[height=0.0925\linewidth]{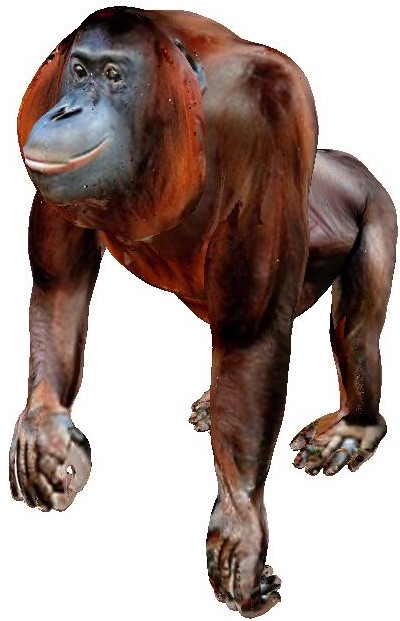} &
        \includegraphics[height=0.0925\linewidth]{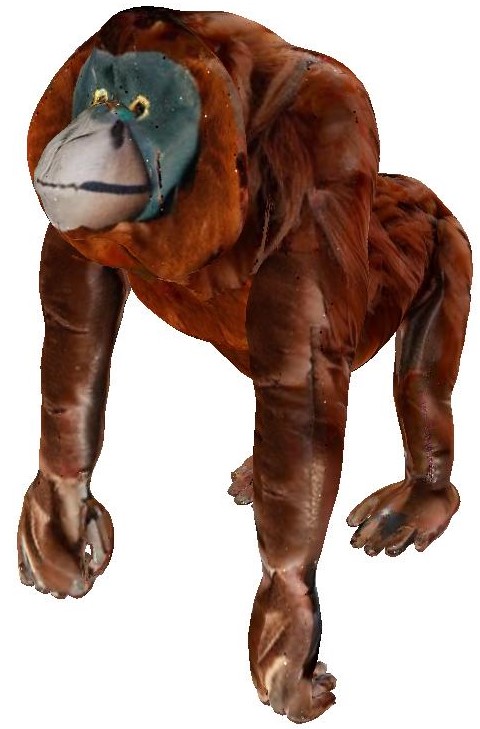} \\
        Input Texture & ``A spotted turtle`` &``A plush turtle''   & ``A plush turtle'' & Input Texture & ``A plush goldfish`` & Input Texture   & ``A plush  \\
        & Refinement & Refinement  & & &  &  & orangutan''
    \end{tabular}
    }
    \vspace{-0.3cm}
    \caption{Texture Editing. The first row presents results for localized scribble-based editing, the original prompt was also used for the refinement step. The second row shows global text-based edits, with the last result showing a texture generated using the same prompt without conditioning on the input texture which clearly results in a completely new texture.}
    \label{fig:edits}
\end{figure*}